\documentclass[11pt]{article}
\usepackage[final]{acl}

\usepackage{times}
\usepackage{latexsym}
\usepackage{svg}
\usepackage[T1]{fontenc}
\usepackage[utf8]{inputenc}

\usepackage{microtype}
\usepackage{hyperref}
\usepackage{url}
\usepackage{booktabs}
\usepackage{xspace}
\usepackage{graphicx}
\usepackage{textcomp}
\usepackage{multicol}
\usepackage{multirow}
\usepackage{enumitem}
\usepackage{pifont}
\usepackage{amsmath}
\usepackage{float}
\usepackage{tikz}
\usepackage{subcaption}
\usepackage{lineno}
\usepackage{amsfonts}

\usepackage[bottom]{footmisc}

\usepackage{stfloats}
\usepackage{algorithm}
\usepackage[noend]{algpseudocode}

\usepackage{inconsolata}
\usepackage{xcolor}
\definecolor{tokred}{HTML}{BD1D1E}
\definecolor{tokblue}{HTML}{0730B8}
\usepackage[version=4]{mhchem}
\usepackage{relsize}
\usepackage{amssymb}
\usepackage{tcolorbox}
\usepackage{caption}
\usepackage{xurl}

\newtcbox{\btokbox}{on line, arc=2pt, colback=gray!15, colframe=gray!15,
  boxrule=0pt, boxsep=1pt, left=1pt, right=1pt, top=1pt, bottom=1pt}
\newcommand{\btok}[1]{\btokbox{\texttt{\scalebox{0.89}{#1}}}}

\definecolor{darkblue}{rgb}{0, 0, 0.5}
\hypersetup{colorlinks=true, citecolor=darkblue, linkcolor=darkblue, urlcolor=darkblue}

\def \idcd{$\mathcal{I}_{\text{dcd}}$\xspace}

\title{Embracing Anisotropy: Turning Massive Activations \\ into Interpretable Control Knobs for Large Language Models}

\author{
  Youngji Roh \quad
  Hyunjin Cho \quad
  Jaehyung Kim \\
  Yonsei University \\
  \texttt{\{youngjiroh, cyberhyunjin, jaehyungk\}@yonsei.ac.kr}
}

\begin{document}
\maketitle

\begin{abstract}
Large Language Models (LLMs) exhibit highly anisotropic internal representations, often characterized by \textit{massive activations}, a phenomenon where a small subset of feature dimensions possesses magnitudes significantly larger than the rest.
While prior works view these extreme dimensions primarily as artifacts to be managed, we propose a distinct perspective: these dimensions serve as intrinsic interpretable functional units arising from domain specialization.
Specifically, we propose a simple magnitude-based criterion to identify \textit{Domain-Critical Dimensions} in a training-free manner.
Our analyses reveal that such dimensions behave as interpretable semantic detectors for symbolic/quantitative patterns or domain-specific terms.
In addition, we introduce \textit{Critical Dimension Steering}, which applies activation steering exclusively to the identified dimensions.
Empirical results show that this approach outperforms conventional whole-dimension steering in domain adaptation and jailbreaking scenarios.\footnote{Code is available at \url{https://github.com/nyancat0222/dimension-analyzer}}
\end{abstract}
\section{Introduction}

Despite the versatile capabilities of Large Language Models (LLMs) across diverse tasks, 
prior research indicates that their internal activations exhibit a distinct lack of \textit{isotropy} \citep{ethayarajh2019representationanisotropic, rudman2023stable}.
In particular, analyses on Transformer-based models reveal that representations do not uniformly utilize the embedding space dimensions  \citep{ding2022isotropycalibration, razzhigaev2024anisotropytransformer}. 
This anisotropy, characterized by a highly non-uniform distribution of activations, is closely linked to the presence of a few outlier dimensions exhibiting variances orders of magnitude larger than others \citep{hammerl2023anisotropyoutlier, dettmers2022int8}.
\citet{sun2024massive} characterizes this phenomenon as \textit{massive activations}, where a small set of features attains values tens of thousands of times larger (\textit{e.g.}, $100,000\times$) than the remainder.
\begin{figure}[t]{
\centering
\vspace{-0.03in}
\includegraphics[width=1\columnwidth]{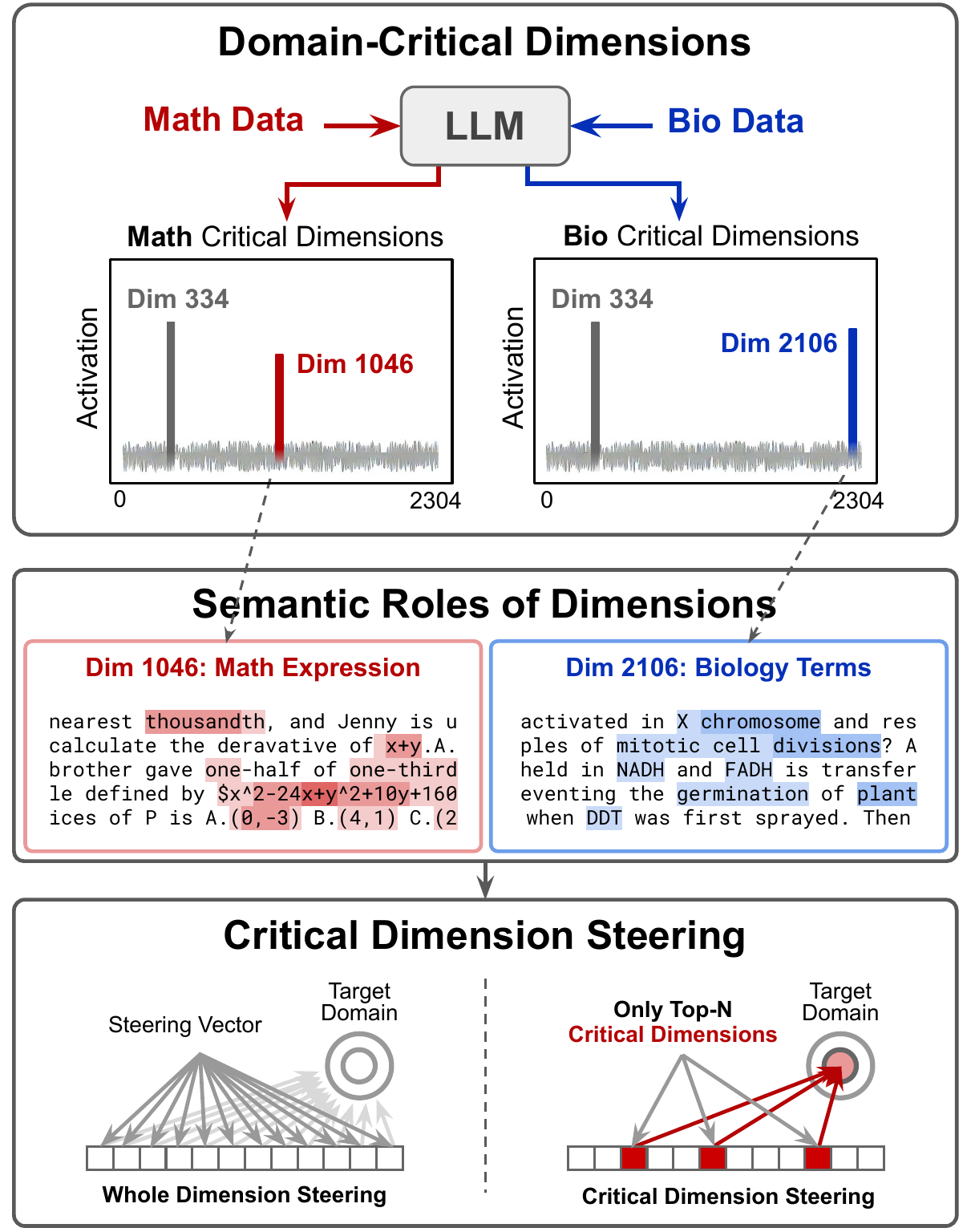}
\vspace{-6mm}
\caption{
{Overview of the study.}
\textit{Domain-Critical Dimensions} identified via activation magnitude act as interpretable functional units. 
We also introduce \textit{Critical Dimension Steering}, which intervenes only on the identified dimensions for precise model control.
}
\label{fig:figure_1}
}
\vspace{1mm}
\end{figure}

A substantial body of literature argues that this representational imbalance prevents models from fully exploiting their capacity, potentially obscuring meaningful linguistic distinctions \citep{gao2019repdeg, timkey2021rogue}. 
Numerous studies have proposed to algorithmically suppress the influence of extreme dimensions in order to promote representational isotropy \citep{wei2022outliersupp, he2024minimisingoutlier}.
Conversely, some perspectives suggest that anisotropy is a \textit{natural} and potentially \textit{beneficial} byproduct of stochastic gradient descent \citep{zhu2018anisotropicsgd}, arguing that preserving these characteristics is crucial for maintaining downstream performance \citep{rudman2023stable}. 
However, both perspectives largely treat these extreme dimensions as \textit{artifacts} to be managed, either suppressed or preserved, rather than as features whose intrinsic functional role  should be examined.

In this work, we propose a distinct perspective: \textit{we hypothesize the emergence of extreme dimensions as a direct consequence of functional specialization}.
Consequently, these dimensions likely to encode domain-specific knowledge,  serving as intrinsic {interpretable functional units} that can be leveraged for {better model control}.
Unlike interpretation methods such as probe-based classifiers \citep{burns2023discoveringlatent, li2023inference} or sparse autoencoders (SAEs) \citep{cunningham2023sae, bricken2023monosemanticity}, which require learning additional parameters to map internal states to concepts or disentangle polysemantic features, our perspective allows us to uncover specialized features directly from the pre-trained model by exploiting simple statistics without additional training.

Based on this hypothesis, we present a method for identifying specialized, domain-specific critical dimensions in \S\ref{sec:2}. 
Our approach is grounded in two key observations regarding feature dimensions: (1) \textit{sparsity}, where a small subset of dimensions plays a crucial role in model performance, and (2) \textit{extremity}, where domain-discriminative dimensions naturally emerge with high-magnitude activations.
Building on these insights, we propose a simple yet effective approach to identify \textbf{Domain-Critical Dimensions} by selecting the top-$k$ dimensions based on activation magnitude.
These dimensions substantially overlap with the ground-truth critical dimensions discovered through computationally expensive iterative masking and evaluation, indicating that functional importance can be inferred from activation statistics without supervision.

Given the functional importance of these dimensions, in \S\ref{sec:3}, we conduct qualitative analyses to investigate their semantic roles by examining token-level activation patterns.
These analysis reveals that individual domain-critical dimensions behave as highly interpretable feature detectors, as they are selectively activated on coherent concepts.
For example, with experiments on Gemma-2-2b-it \citep{team2024gemma}, we reveal that dimension 1046 strongly activates on mathematical terms (\btok{\textcolor{tokred}{+}}, \btok{\textcolor{tokred}{x}}, \btok{\textcolor{tokred}{$\infty$}}), dimension 2106 on specific biological terms (\btok{\textcolor{tokblue}{\_ATP}}, \btok{\textcolor{tokblue}{\_NAD}}, \btok{\textcolor{tokblue}{\_phosphorylation}}), and dimension 334 on the topic keywords (\btok{{\_mathematics}}, \btok{{\_biology}}).

Finally, we demonstrate that domain-critical dimensions can be utilized for better model control (\S\ref{sec:4}).
We introduce \textbf{Critical Dimension Steering (CDS)}, a targeted inference-time intervention that applies activation steering—\textit{i.e.}, modifying internal activations during the forward pass \citep{turner2023steering, rimsky2024steering}—only to the top-$k$ dimensions identified by our method.
As a sparse set of dimensions dominates behavior within a target domain, manipulating activations along these specific axes serves as more effective control knobs than influencing the entire latent space. 
We validate the efficacy of CDS in two distinct applications.
In a domain adaptation setup using MMLU benchmark \citep{hendrycks2021measuring}, CDS improves domain-specific accuracy over whole-dimension steering baseline in 34 out of 57 subjects. 
In a jailbreaking setup with AdvBench \citep{zou2023universal}, applying CDS increases attack success rate to 92\%, surpassing 84\% achieved by whole-dimension steering baseline. 

\section{Existence and Identification of Domain-Critical Dimensions}\label{sec:2}
In this section, we present a method for identifying the top-$k$ dimensions critical to a target domain.
In \S\ref{sec:2.1}, we first observe the existence of two types of feature dimensions: \textit{functionally critical} dimensions and \textit{domain-discriminative} dimensions. 
In \S\ref{sec:2.2}, we then leverage these properties to identify the final \textit{{domain-critical dimensions}} using simple statistical indicators.
 
\paragraph{Problem setup.}
We consider pre-trained LLM composed of $L$ Transformer layers and hidden dimension $D$.
Given an input token sequence $\mathbf{x}=[x_1,\dots,x_T]$, the model first maps tokens to the input embedding ${h}_{0}\in \mathbb{R}^{T \times D}$.
Each layer $l$ takes the hidden state ${h}_{l-1}$ as input and produces ${h}_{l}$ via residual connection, where the sub-layer output is added to the input. 
We refer to \textit{activations} as scalar entries of ${h}_{l}$, and dimensions that encode significant semantic information or functional roles as \textit{feature dimensions}.
Our goal is to identify domain-critical dimensions $\mathcal{I}_\text{dcd} \subset \{1, \dots, D\}$, feature dimensions that are both specialized to a target domain and critical for preserving performance within that domain.

Throughout this work, we operationalize a \textit{domain} using the MMLU benchmark \citep{hendrycks2021measuring}, which comprises 57 subjects spanning STEM, humanities, and social sciences.
For each subject, we sample 100 prompts from the test split and partition them into two disjoint subsets: 50 prompts as an \textit{identification set} and 50 prompts as an \textit{evaluation set}.
The identification set is used to extract $\mathcal{I}_\text{dcd}$, while the evaluation set is reserved for validating the identified dimensions and for downstream evaluations.

\subsection{Intriguing Feature Dimensions of LLMs}\label{sec:2.1}

\paragraph{Existence of critical dimensions.}
We first observe sparse dimensions that are \textit{functionally critical}.
Intuitively, one way to estimate the importance of a specific dimension is measuring the performance degradation when target dimension is removed from the inference process.
If all dimensions contribute equally to the model's performance, masking a single dimension among thousands should result in a small performance loss.
To verify this, we iteratively evaluate the performance after masking individual dimension on Gemma-2-2b-it \citep{team2024gemma} {and Qwen-3-8b \citep{yang2025qwen3}.}
For each dimension, we set its activation to zero at every layer and evaluate the accuracy using the evaluation set. 
Input embedding layer is excluded from masking, to isolate the effect of internal processing.

\begin{table}[t]
    \begin{center}
    \caption{Masking effect of individual dimensions. Rank $k$ denotes the average accuracy after masking the single dimension that rank $k$ in terms of accuracy drop for each subject. Baseline accuracy for Gemma-2-2b-it is 56.53\% and Qwen-3-8b is 73.30\%.}
    \vspace{-0.1cm}
    \resizebox{1.0\linewidth}{!}{
        \begin{tabular}{l| ccccc}
            \toprule
            {Model} & \textbf{Rank 1} & \textbf{Rank 2} & \textbf{Rank 5} & \textbf{Rank 10} & \textbf{Rank 100} \\ 
            \midrule
             \textbf{Gemma-2-2B-IT} & 41.97 & 46.81 & 50.71 & 52.39 & 54.25 \\ 
             \hfill Acc. Drop (\%) & \textbf{(-14.56)} & (-9.72) & (-5.82) & (-4.14) & (-2.28) \\
             \midrule
             {\textbf{Qwen-3-8B}} & 21.97 & 38.67 & 63.44 & 68.98 & 71.97 \\ 
             \hfill Acc. Drop (\%) & \textbf{(-51.33)} & (-34.63) & (-9.86) & (-4.32) & (-1.33) \\
            \bottomrule
        \end{tabular}
    }
    \par 
    \label{tab:table_1}
    \end{center}
    \vspace{2mm}
\end{table}

\begin{figure}[t]{
\begin{center}
    \includegraphics[width=\columnwidth]{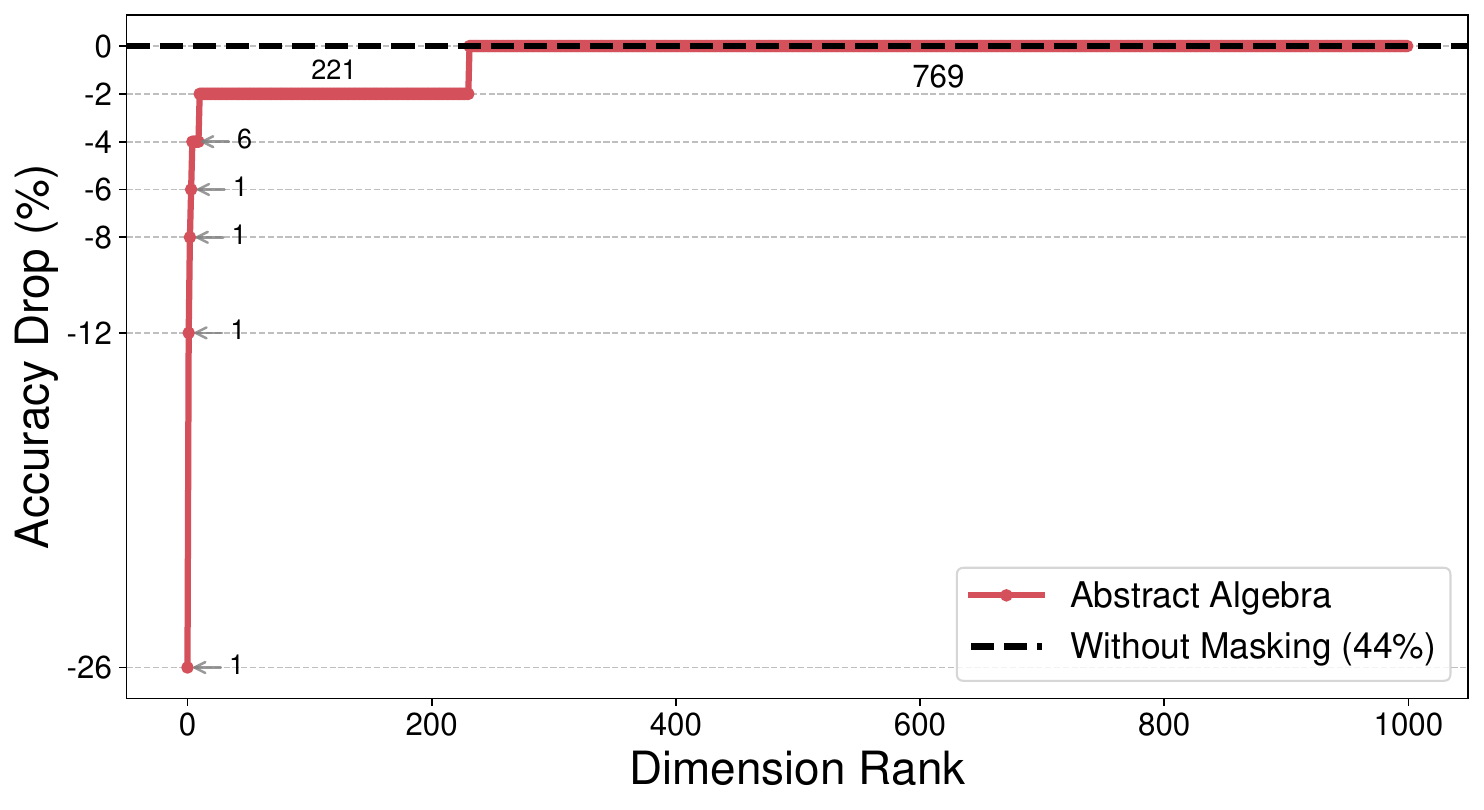}
    \vspace{-6mm}
    \caption{Impact of masking individual dimensions on \textit{Abstract Algebra}.
    A sharp accuracy drop is observed in few dimensions.
    Annotations indicate the number of dimensions to each level of accuracy drop.
    }
    \label{fig:figure_2}
\end{center}
}
\end{figure}

From this experiment, we observe that \textit{disabling an even single dimension can yield a catastrophic collapse in model capabilities}.
Table \ref{tab:table_1} presents the rank-wise average masking effect of individual dimensions across subjects of evaluation set, showing that masking the most critical individual dimension alone yields an average performance drop of 51.3\% in Qwen-3-8b.
Figure \ref{fig:figure_2} shows the distribution of performance drop on \textit{abstract algebra} subject when masking individual dimensions {of Gemma-2-2b-it}. While most of dimensions show negligible impact on performance, very small number of dimensions exhibit significant performance drops, showing a maximum accuracy drop of 26\%.
The same trend appears across all subjects, and these results are presented in \S\ref{supp:A1}. 
Overall, these results indicate that hidden state dimensions do not contribute equally to the model's task performance, but rather that there exist few \textit{functionally critical dimensions} that crucially affect performance.
\begin{figure}[b]{
\vspace{3mm}
\begin{center}
    \includegraphics[width=\columnwidth]{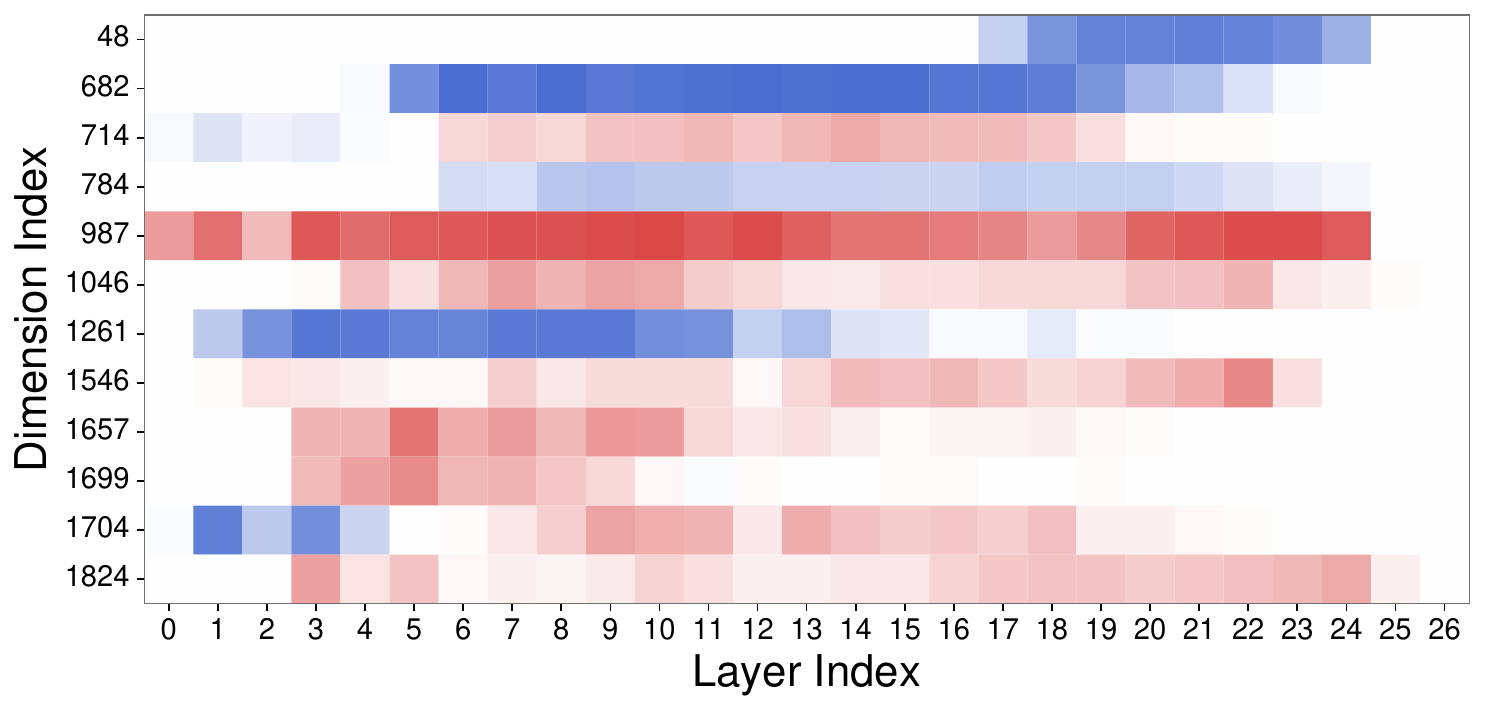}
    \vspace{-6mm}
    \caption{Dimensions with discriminated activation patterns between \textit{high school mathematics} and \textit{high school biology}.
    \textcolor{tokred}{Red} indicates higher activation frequency for mathematics, while \textcolor{tokblue}{Blue} indicates biology.
    }
    \label{fig:figure_3}
\end{center}
}
\end{figure}
\paragraph{Existence of discriminative dimensions.}
Next, we observe extreme dimensions that are activated in a \textit{domain-discriminative} manner.
As illustrated in \S\ref{supp:A2}, hidden states corresponding to queries from 14 high school related subjects naturally form distinct clusters across layers, confirming domain-wise separation in the representational space. 
To identify specific dimensions driving this distinction, we conducted a comparative analysis between \textit{high school mathematics} and \textit{high school biology} subjects.
Specifically, we consider a dimension as \textit{active} for a given query, if its activation value deviates by more than $3\sigma$ from the mean, following the outlier dimension criterion of \citet{zhao2025analysis}.

For each subject, we compute activation frequency for every dimension, defined as the percentage of queries within the dataset where the dimension is classified as active.
In Figure \ref{fig:figure_3}, we observe discriminative dimensions that exhibit a distinct disparity in activation frequency (\textit{i.e.}, $>30\%$) between the two subjects.
Dimensions such as 682 and 1261 are persistently active in biology, whereas dimensions like 987 and 1046 exhibit the opposite pattern.
This suggests that the model's ability to differentiate domains is encoded within specific, highly active \textit{domain-discriminative dimensions}.

\subsection{Identifying Domain-Critical Dimensions}\label{sec:2.2}

Our preliminary observations highlight two key properties of feature dimensions in LLMs:
\begin{enumerate}[label=\arabic*), topsep=-0.1ex, itemsep=-0.1ex]
    \item {\textit{Sparsity}}: a small set of {critical dimensions} critically affects performance.
    \item {\textit{Extremity}}: {discriminative dimensions} are characterized by extreme activation values.
\end{enumerate}

\begin{figure*}[t]{
    \centering
    \scriptsize
    \begin{minipage}[t]{0.66\textwidth}
        \vspace{0pt}
        \begin{tabular}{@{} p{\linewidth} @{}}
        \toprule 
        \textbf{The following are multiple choice questions (with answers) about high school biology.}\\
        You must respond with a single alphabet character.\\[0.5em]
        Energy is harvested during cellular respiration in stages. Which of the following correctly states which phase of cellular respiration harvests the most energy and the correct explanation why? \\[0.5em]
        A. The most energy is released during the Krebs cycle because it is here that pyruvate is completely ...\\
        B. ... both \ce{FADH2} and NADH are produced. Each of those molecules will release 2 ATPs ...\\
        C. ... phosphorylation because in addition to the phosphorylation of ADP into ATP ...\\
        D. ... during oxidative phosphorylation because \ce{H2O} is completely broken down into \ce{H+} and \ce{O2}. \\
        Answer: \\
        \bottomrule
        \end{tabular}
    \end{minipage}
    \hfill
    \begin{minipage}[t]{0.33\textwidth}
        \vspace{0pt}
        \centering
        \includegraphics[width=\textwidth]{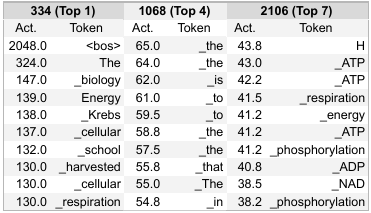}
    \end{minipage}
    \vspace{-0.1cm}
    \caption{Prompt-level analysis of \textit{Domain-Critical Dimensions}. 
        \textbf{(a)} A representative query regarding cellular respiration used for the token analysis.
        \textbf{(b)} Top activating tokens for dimensions 334, 1068, and 2106. Dimension 334 highlights topic-relevant tokens, 1068 targets functional tokens, and 2106 focuses on biological terminology related to cellular respiration.
        }
    \label{fig:figure_4}
    }
\end{figure*}
\vspace{0.2cm}
While we aim to identify and analyze these structurally significant dimensions, locating them via exhaustive iterative search (\S\ref{sec:2.1}) is limited for real-world applications, as this approach is computationally expensive.
On the other hand, we observed that domain-discriminative dimensions could be found by identifying the dimensions with extreme activations.
From this, we hypothesize that functionally critical dimensions also exhibit the statistical property of \textit{extremity}, such as high magnitude.

Therefore, we propose efficient approach to identify these domain-specific critical dimensions, which we term \textbf{Domain-Critical Dimensions}, by leveraging statistical properties of activations instead of explicitly measuring the impact of dimensions to the performance.
Specifically, we extract hidden states across $L$ layers of $N$ queries from a target domain dataset using the identification set. 
Let $h_{l, t, j}^{n}$ denote the activation of the $j$-th dimension at the $t$-th token in layer $l$ for the $n$-th query.
Then, we define \textit{importance score} $s_j$ for $j$-th dimension as the average magnitude of its token-averaged activations across all layers and queries:
\begin{equation*}
    s_j = \frac{1}{L} \sum_{l=1}^{L} \left( \frac{1}{N} \sum_{n=1}^{N} \left| \frac{1}{T} \sum_{t=1}^{T} h_{l, t, j}^{n} \right| \right)
\end{equation*}
Overall, {importance score} reflects \textit{how strongly the dimension is activated in the target domain}.

We rank all dimensions in descending order of importance score and select the top-100 dimensions as our candidate set of domain-critical dimensions $\mathcal{I}_\text{dcd}$.
To validate the selected dimensions, we compare them against the ground-truth functionally critical dimensions found in \S\ref{sec:2.1}.
Remarkably, the proposed approach exhibits an average recall of 89.39\% for the functionally critical dimensions ranked within the top-10.
Furthermore, the masking impact of $\mathcal{I}_\text{dcd}$ closely mirrors the results of Table \ref{tab:table_1}, with masking the rank-1 domain-critical dimension yielding an average performance drop of 11.40\%.
Detailed validation results are in \S \ref{supp:A3}.

\begin{figure*}[t]
    \centering
    \begin{subfigure}[b]{1.0\textwidth}
        \centering
        \hspace{-0.3cm}
        \includegraphics[width=\textwidth]{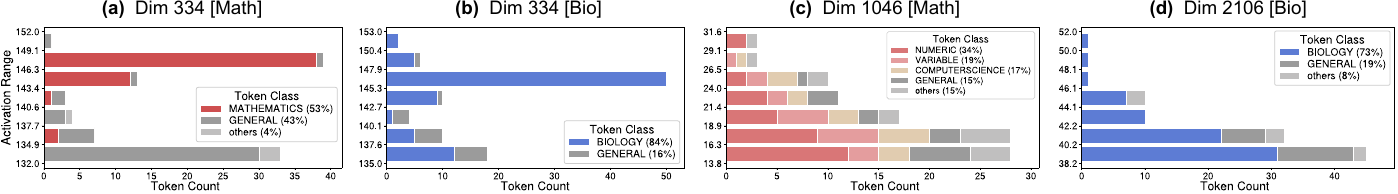} 
    \end{subfigure}
    \par
    \begin{subfigure}[b]{1.0\textwidth}
        \centering
        \includegraphics[width=\textwidth]{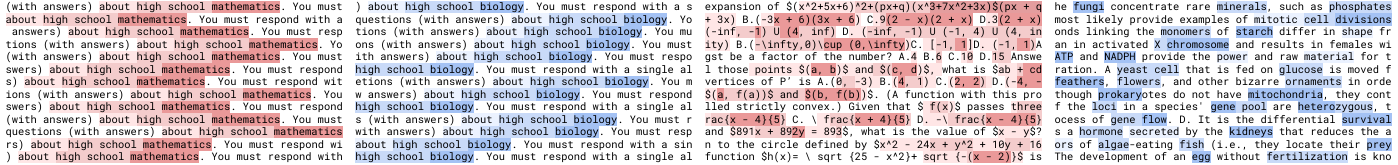}
    \end{subfigure}
    \caption{Dataset-level analysis of \textit{Domain-Critical Dimensions}.
        The bar charts display the distribution of domain token classes across the activation ranges for the most activated 100 tokens.
        The text sequences visualize the activation heatmaps for the corresponding dimensions.
        }
    \label{fig:figure_5}
\end{figure*}
\section{Semantic Roles of Domain-Critical Dimensions}\label{sec:3}
In this section, we investigate the semantic roles of domain-critical dimensions by examining highly activated tokens using Gemma-2-2b-it. 
The results with Gemma-2-9b-it and Llama-3.1-8b-it, which are consistent with the results on Gemma-2-2b-it, are presented in \S\ref{supp:A7} and \S\ref{supp:A8}, respectively.

\paragraph{Prompt-level analysis.}
We first analyze the roles of domain-critical dimensions at the prompt level.
For a given input prompt $x$ and a selected dimension $j$, ranked by $s_j$ in \S\ref{sec:2.2}, we define an activation for $t$-th token as
$
a_{t,j}= \frac{1}{L}\sum_{l=1}^{L} h_{l,t,j}
$
to identify the highest \textit{activated tokens} within the token sequence.
We investigate whether these tokens have semantic interpretability. 
For intuitive illustration, we examine a representative input prompt about \textit{cellular respiration} in Figure \ref{fig:figure_4}a.

Here, we observe the existence of semantic commonalities among tokens.
Figure \ref{fig:figure_4}b presents the highest activated tokens corresponding to the top-1, top-4, top-7 ranked \textit{high school biology} critical dimensions within a representative prompt.\footnotemark
\textit{334} shows extreme values at the initial positional tokens (\btok{<BOS>}, \btok{The}), but subsequently activates strongly on tokens relevant to prompt topic such as \btok{\_biology}, \btok{\_Krebs}, \btok{\_cellular}, \btok{\_respiration}.
\textit{1068} consistently activates on functional words like \btok{\_the}, \btok{\_is}, \btok{\_to}.
\textit{2106} triggers on terms such as \btok{\_ATP}, \btok{\_NAD} and \btok{\_phosphorylation}, sharing a semantic related to the specific biological process of cellular respiration.

\paragraph{Dataset-level analysis.} 
\footnotetext{These are selected to illustrate diverse semantic roles. See \S\ref{supp:A5} for other dimensions and additional prompt examples.}
We extend our analysis to biology and math datasets to determine whether domain-critical dimensions exhibit global semantic roles.
The biology dataset is sourced from the \textit{high school biology} subject, and the math dataset is from the \textit{high school mathematics} subject.
For each dataset, we perform token classification via Named Entity Recognition, and analyze 100 tokens which exhibit the highest activation for each dimension (see \S\ref{supp:B1} for details).

\begin{itemize}[leftmargin=3.5mm, topsep=2pt, itemsep=3pt, parsep=0pt]
    \item[$\circ$] \textbf{\textit{Dimension 334}}
    emerges as the highest ranked domain-critical dimension in both datasets.
    Figure \ref{fig:figure_5}a shows that \btok{MATHEMATICS} entities constitute the largest share of activated tokens (53\%), exceeding \btok{GENERAL} tokens (43\%).
    In addition, the activation range breakdown shows that tokens with the highest activations are dominantly classified as mathematical entities, whereas general tokens tends to appear in lower activation ranges.
    The activation heatmap further aligns with this view: activations concentrate around explicit domain markers in the prompt, such as \btok{\textcolor{tokred}{\_mathematics}}, suggesting that dimension 334 acts as a micro-level topic activator and a macro-level domain classifier.
    
    In the biology dataset, {dimension 334} exhibits the same functional role but with biology content.
    As shown in Figure \ref{fig:figure_5}b, \btok{BIOLOGY} entities accounts for 84\%, making up the overwhelming majority of the activated tokens.
    The activation heatmap similarly highlights activation concentration around explicit domain marker, \btok{\textcolor{tokblue}{\_biology}}, similar to the behavior on math dataset.
    Taken together, these results suggest that {dimension 334} captures domain identity in a direct and dataset-consistent way; it amplifies tokens that explicitly signal the domain and tokens that carry domain-specific topical content.
    \item[$\circ$] \textbf{\textit{Dimension 1046}} is prominent in the math dataset and specializes in quantitative and symbolic content.
    Top activated tokens by dimension 1046 on math dataset include numeric magnitude words and arithmetic symbols (e.g., \btok{\textcolor{tokred}{$\infty$}}, \btok{\textcolor{tokred}{2}}, \btok{\textcolor{tokred}{$x$}}, \btok{\textcolor{tokred}{+}}), indicating a strong association with numbers, operators, and variable-like tokens.
    Figure \ref{fig:figure_5}c shows that \btok{NUMERIC} and \btok{VARIABLE} tokens account for 34\% and 19\% of the activated tokens, respectively.
    The activation heatmap concentrates around equations and operator dense spans, suggesting that dimension 1046 tracks quantitative form.
    In the biology dataset, by contrast, it activates \btok{GENERAL} tokens by 44.0\%, showing no domain specificity.
    \item[$\circ$] \textbf{\textit{Dimension 2106}} is a biology-specific domain-critical dimension. 
    In Figure \ref{fig:figure_5}d, \btok{BIOLOGY} entities account for 73\% of activated tokens, with \btok{GENERAL} at 19\% and the remainder in other categories. 
    The top activated tokens in biology include concrete biological terms such as \btok{\textcolor{tokblue}{\_roots}} and \btok{\textcolor{tokblue}{\_cell}}, consistent with a lexically grounded biology feature.
    In the corresponding heatmap, {dimension 2106} responds strongly to tokens like \btok{\textcolor{tokblue}{X\_chromosome}}, and \btok{\textcolor{tokblue}{mitochondria}}, showing that it captures fine grained biological semantics rather than generic ones.
    However, this domain-specific focus vanishes in the math dataset, where \btok{GENERAL} tokens accounts for the majority of 54.0\%.
\end{itemize}

\section{Enhanced Model Controllability via Critical Dimension Steering}
\label{sec:4}
In this section, we introduce \textbf{Critical Dimension Steering (CDS)}, an inference-time intervention method that restricts activation steering to a sparse set of top-$k$ domain-critical dimensions.
Our key idea is that if domain-specific information is encoded in a subspace of domain-critical dimensions, then steering only that subspace would yield more accurate domain-targeted control while minimizing interference with unrelated capabilities.

\paragraph{Activation steering and CDS.}
Activation steering is a method that modulates model behavior by shifting internal representations along a specific direction during inference.
Steering vector $\mathbf{v}$ should encode a shift from undesired behavior to desired one.
Let $\mathcal{D}_{+}$ and $\mathcal{D}_{-}$ denote the positive and negative datasets representing the desired and undesired behaviors.
For a given layer $l$, the mean activation vectors $\mu_{+}$ and $\mu_{-}$ is computed by averaging hidden states over token positions within each prompt and subsequently averaging across all prompts in the respective sets.
Then, the steering vector $\mathbf{v}_l \in \mathbb{R}^{D}$ is defined as $\mathbf{v}_l = \mu_{+} - \mu_{-}$.

The standard activation steering injects $\mathbf{v}_l$ into the forward pass of the model by modifying the hidden state $h_l$ to $\widetilde{h_l}$ as follows:
\begin{equation*}
    \widetilde{h}_l = h_l + \alpha \cdot \mathbf{v}_l,
\end{equation*}
where $\alpha$ is a scalar controlling the steering strength.
CDS modifies this process by applying the intervention only to the identified dimensions.
Let $m \in \{0, 1\}^{D}$ be a binary mask where $m_j = 1$ if dimension $j$ is among the top-$k$ domain-critical dimensions, and $m_j = 0$ otherwise.
Then, the activation steering is selectively applied with $m$: 
\begin{equation*}
\widetilde{h}_l = h_l + \alpha \cdot (m \odot \mathbf{v}_l),
\end{equation*}
where $\odot$ denotes element-wise multiplication.

\subsection{Experiment on Domain Adaptation}
\label{sec:4.1}

\paragraph{Setup.}
We evaluate the effectiveness of CDS on \textit{domain adaptation} with Gemma-2-2b-it  \citep{team2024gemma} using MMLU benchmark \citep{hendrycks2021measuring}.
Following the setup in \S\ref{sec:2}, we utilize the identification set to derive the mask $m$ and construct the steering vector $\mathbf{v}$, while reporting performance on the evaluation set.
To construct the domain-specific steering vector $\mathbf{v}_{F \to T}$, we partition the identification set based on the model's prediction correctness.
Specifically, we define $\mathcal{D}_{+}$ as the set of prompts where the model generated the correct answer, and $\mathcal{D}_{-}$ as the set where it generated an incorrect answer.
Consequently, the vector $\mathbf{v}_{F \to T}$ represents a directional shift from incorrect to correct reasoning within that specific domain. 
We compare our approach against two baselines: (i) \textit{Standard}, the original model performance without intervention ($\alpha=0$); (ii) \textit{Whole-Dimension Steering (WDS)}, activation steering using the entire dimensional space.
For the evaluation, we use LM-evaluation-harness \citep{eval-harness} and report exact-match (EM) accuracy under greedy decoding for deterministic evaluation.
We report the best accuracy achieved for each method over all hyperparameters (see \S\ref{supp:B3} for experimental details).

\begin{figure*}[t]
\centering
\includegraphics[width=1.0\textwidth]{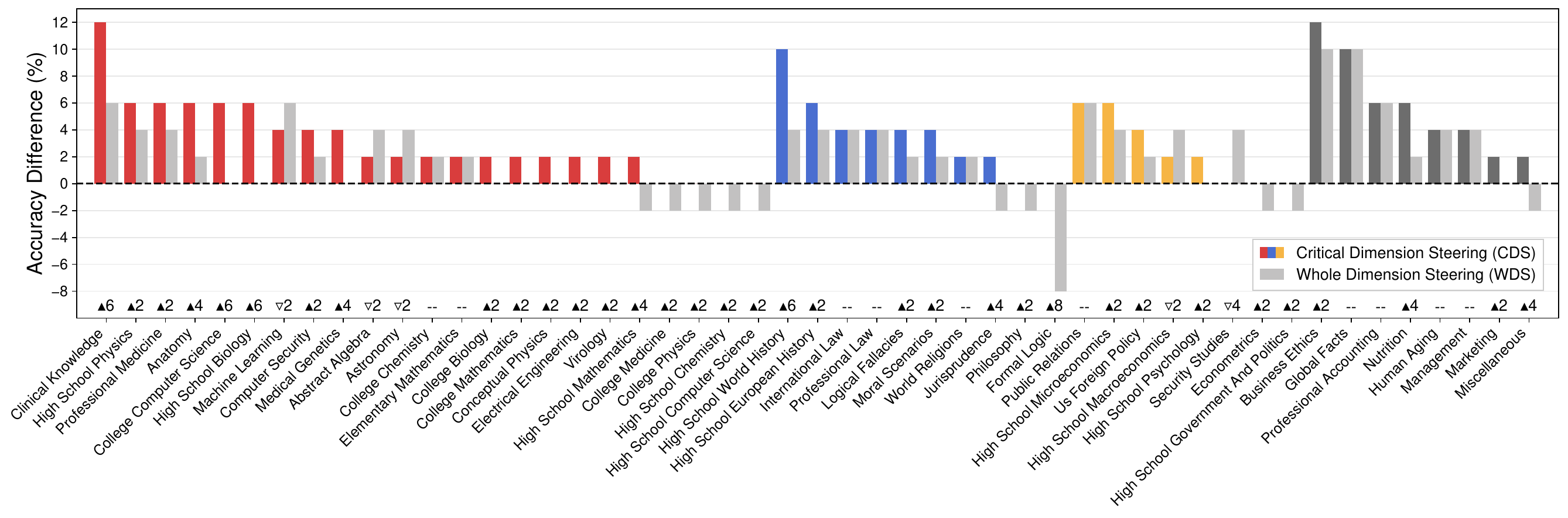}
\vspace{-7mm}
\caption{Performance comparison between CDS and WDS baseline across MMLU subjects. 
The bars represent the accuracy difference from the unsteered standard baseline. 
The numbers on subject name denote CDS improvement ($\blacktriangle$/$\triangledown$) over WDS.
Subjects are divided into categories:
\textcolor[HTML]{d93d3d}{STEM}, \textcolor[HTML]{4a6ed0}{Humanities}, \textcolor[HTML]{E6A729}{Social Sciences}, and \textcolor[HTML]{6D6D6D}{Other}.
Subjects with identical scores (total of 8) across all methods are excluded.
}
\vspace{-1mm}
\label{fig:figure_6}
\end{figure*}

\paragraph{Results.}
Figure \ref{fig:figure_6} reports subject-wise accuracy changes by CDS and WDS, compared to the standard baseline without steering.
On average, the proposed CDS achieves 3.09\% improvement in accuracy while WDS shows 1.51\% improvement. 
Consequently, CDS outperforms WDS in 34 out of 57 subjects, while WDS is better in 5 cases; the remaining 18 subjects are tied.
Remarkably, CDS yields larger performance gains in STEM where precise knowledge retrieval is crucial.\footnote{We reclassify 6 medical subjects from MMLU's \textit{Other} category to \textit{STEM} to better reflect their scientific nature, supporting clearer interpretation and aligning with MMLU-Global \citep{singh2025global}, which similarly isolates medical subjects.}
These results are consistent with our analysis in \S\ref{sec:3}, which observed specific dimensions encoding domain-specific concepts like biological terms.
For instance, in \textit{high school biology}, CDS achieves an accuracy increase of 6\%, whereas WDS yields no improvement.
We also observe significant gains in medical subject such as \textit{clinical knowledge} by +12\%, significantly surpassing WDS of +6\%.

CDS also effectively mitigates the negative transfer often observed with naive steering vectors, a trend particularly evident in logical and mathematical reasoning tasks.
As shown in \textit{formal logic} and \textit{high school mathematics}, applying WDS results in performance degradation of -8\% and -2\% respectively.
In contrast, by masking out non-critical dimensions, CDS prevents this interference, recovering performance to the standard baseline level in \textit{formal logic} and even achieving a positive gain of +2\% in \textit{high school mathematics}.
Overall, these results confirm that CDS allows for targeted enhancement of domain-specific capabilities without disrupting the model's general capability. 
{The results on other models are provided in \S\ref{supp:A10}.}

\subsection{Experiment on Safety Jailbreaking}
\label{sec:4.2}
\paragraph{Setup.}
We further evaluate the efficacy of CDS in the context of \textit{LLM jailbreaking} \citep{wei2023jailbroken} using the AdvBench dataset \citep{zou2023universal}.
Jailbreaking aims to steer the model from a refusal state to a compliant state when facing harmful queries, and we specifically focus on the control of refusal mechanism using activation steering. 
To derive the steering vector $\mathbf{v}$, we construct $\mathcal{D}_{+}$ and $\mathcal{D}_{-}$ using pairs of compliant and refusal behavioral suffixes, respectively, and compute the mean difference in hidden states following \citet{lee2024cast}.\footnote{We utilize a set of behavioral compliance and refusal suffixes adopted from the IBM Activation Steering repository \url{https://github.com/IBM/activation-steering/blob/main/docs/demo-data/behavior_refusal.json}}
For the dimension mask $m$, we identify harmfulness-relavant critical dimensions by utilizing 260 queries and corresponding target responses provided in AdvBench.
We compare CDS against two baselines: (i) \textit{WDS}, same as \S\ref{sec:4.1}, and (ii) \textit{Random}, where the mask $m$ is constructed by selecting the same number of $k$ dimensions at random.
The primary metrics are (1) Attack Success Rate (ASR) and (2) Text Quality measured on a 1–5 scale, where both are evaluated with GPT-4o-mini as LLM-as-judge. 
We report the best performance attained by each method across the hyperparameter sweep (see \S\ref{supp:B4} for experimental details).

\begin{figure*}[t]
    \centering
    \begin{subfigure}[b]{0.49\textwidth}
        \centering
        \includegraphics[width=\textwidth]{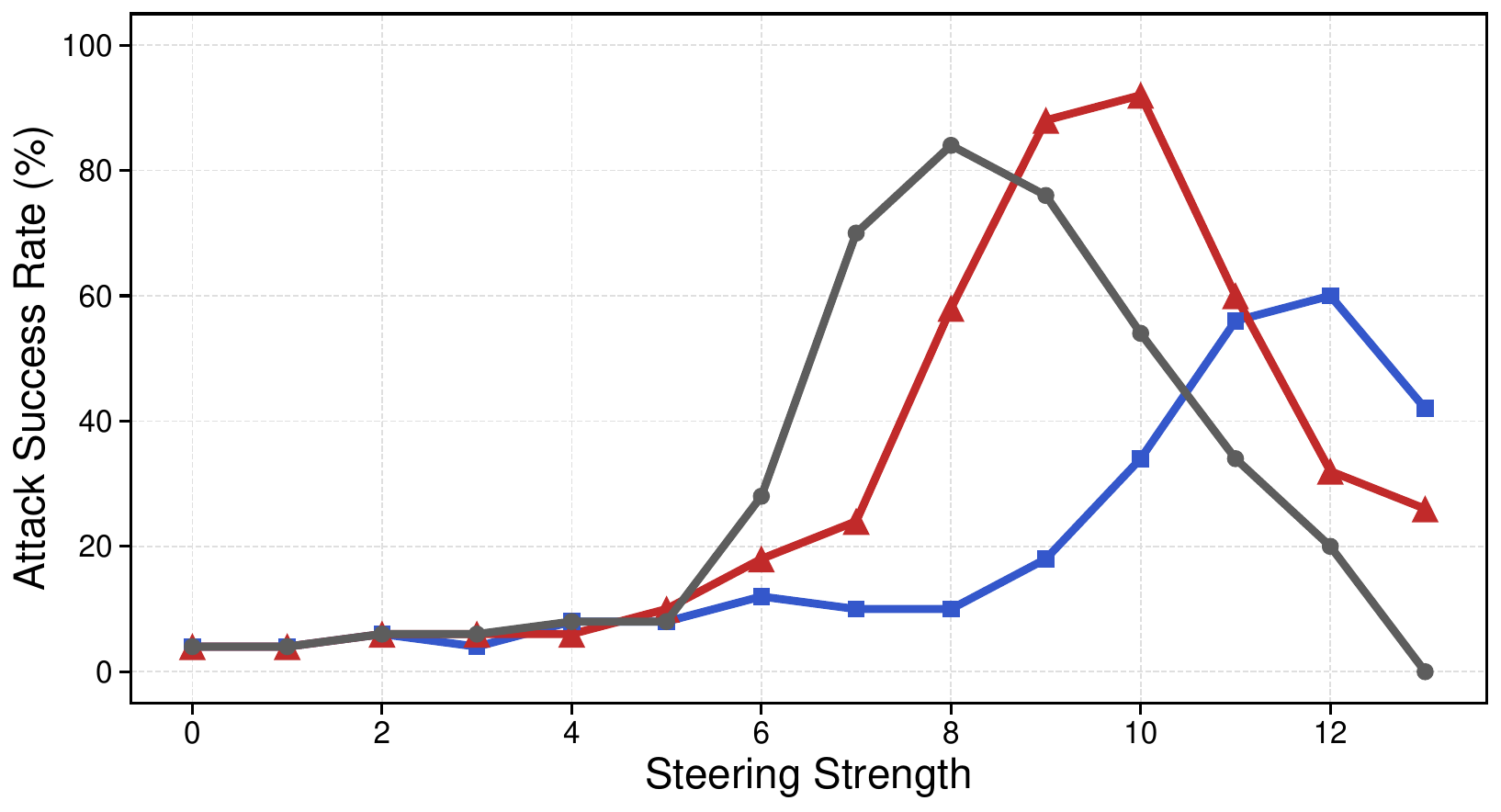}
    \end{subfigure}
    \hfill
    \begin{subfigure}[b]{0.49\textwidth}
        \centering
        \includegraphics[width=\textwidth]{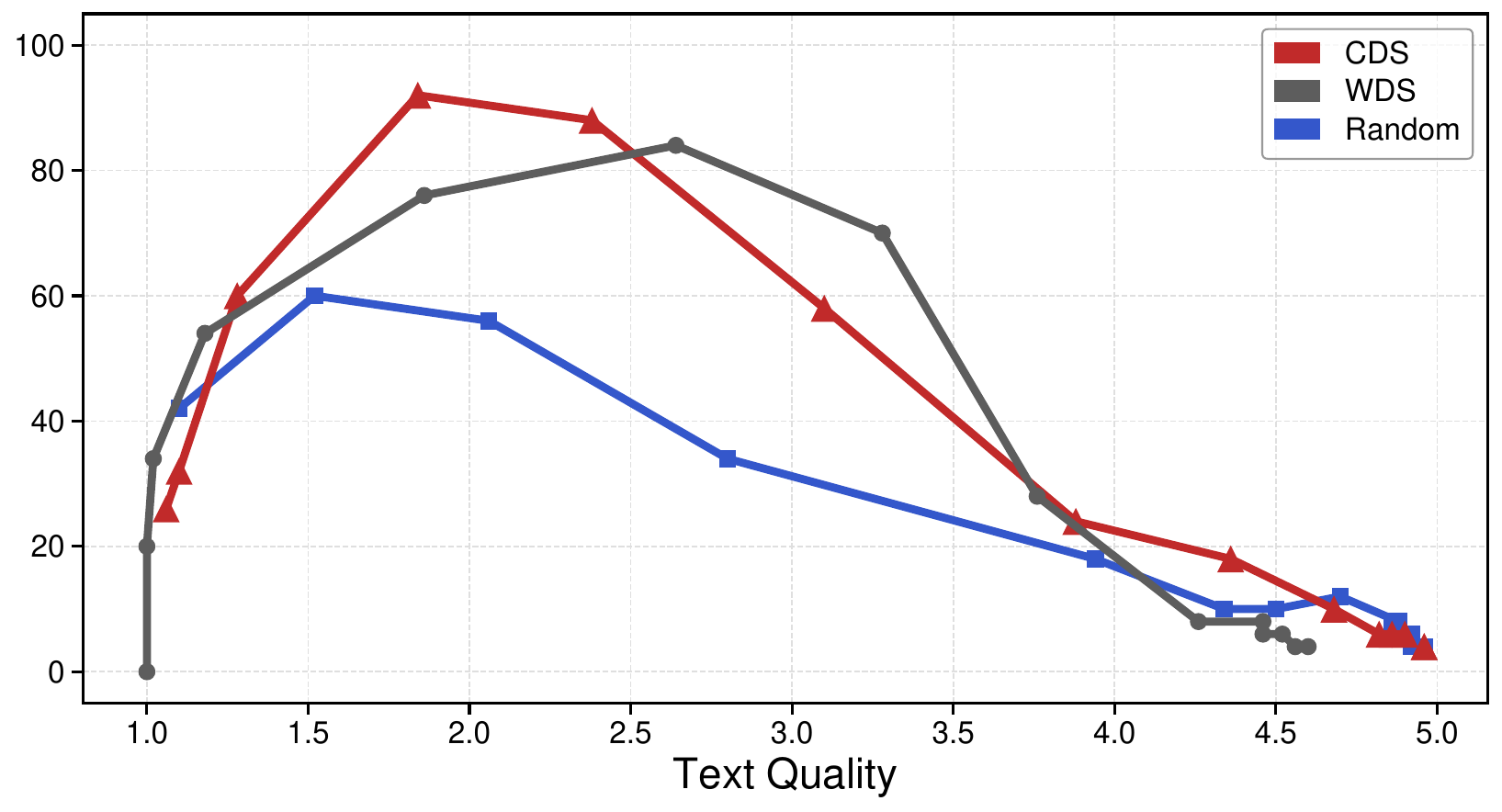}
    \end{subfigure}
    \vspace{-1mm}
    \caption{Attack Success Rate (ASR) across steering strength $\alpha$ and text quality.  
    \textbf{(a)} Evaluation of ASR as a function of steering intensity $\alpha$. 
    \textbf{(b)} Trade-off between ASR and the generated text quality. As $\alpha$ increases, the performance trajectory moves from right to left, indicating the cost in text quality at higher steering intensities.
    } 
    \label{fig:figure_7}
    \vspace{-2mm}
\end{figure*}

\paragraph{Results.} 
Figure \ref{fig:figure_7} presents the comparative performance of CDS against WDS and random baselines.
The results provide empirical evidence that the refusal mechanism is mediated by a sparse set of critical dimensions, which CDS effectively isolates and manipulates.
In Figure \ref{fig:figure_7}a, CDS demonstrates significantly higher steering efficiency compared to the random baseline; as steering strength $\alpha$ increases, CDS rapidly achieves a high ASR, reaching a peak of 92\% at $\alpha=10$.
In contrast, random baseline requires higher strength to achieve even moderate success rate below 60\% and fails to induce meaningful behavioral shifts.
This divergence confirms that the identified critical dimensions are functionally central to the model's safety guardrails, whereas random subspaces have negligible influence.
Notably, CDS achieves a higher peak ASR (92\%) than that of WDS (84\%), despite manipulating only 13\% of the total dimensions.

Figure \ref{fig:figure_7}b further illustrates the trade-off between ASR and text quality under activation steering. 
As the force to artificially modify the model internal for jailbreaking is increased ($\alpha\uparrow$), text quality is decreasing in general.
Nevertheless, each method exhibits different behavior. 
For example, the random baseline highlights the inefficiency of uninformed steering. 
It consumes the text quality budget without yielding proportional gains in ASR, remaining at the bottom-left of the performance frontier.
Importanly, CDS surpasses the saturation point of WDS, while still maintaining text quality scores above 2.
While WDS achieves high ASR in certain regions (\textit{e.g.}, near text quality 2.6), it suffers from more rapid performance degradation compared to CDS which maintains higher ASR at equivalent text quality levels.
This confirms that steering with CDS using critical dimensions allows for a more accurate intervention that breaks refusal barriers effectively, compared to the baselines.
\section{Related Works}

\paragraph{Interpretability units.} 
As LLMs have grown enormously, interpretability of their internal states remain a crucial research direction \citep{clark2019does, meng2022locating}.
Prior works have proposed various interpretability units for tracing concepts, including MLP layers \citep{geva2021transformer}, attention heads \citep{voita2019analyzing, abnar2020quantifying}, neurons \citep{vig2020investigating, dai2022knowledge}, and features of probing classfiers \citep{ettinger2016probing, li2023inference}.
More recently, sparse autoencoders have been used to address the polysemanticity of internal representations, decomposing hidden representations into sparse combinations of monosemantic features \citep{bricken2023monosemanticity, yun2021transformer, cunningham2023sae}.
While these sophisticated units provide high precision and clear functional decomposition, they often entail substantial training overhead. 
In contrast, our approach treats individual hidden-state dimensions as interpretability units. 
By sacrificing the fine-grained granularity of SAEs for computational efficiency and model integrity, we provide a training-free means to identify domain-relevant signals directly within the original model.

\vspace{2mm}
\paragraph{Massive activations.} 
Values flowing through LLMs include entries that deviate significantly from the average, referred to as outliers \citep{kovaleva2021bert, dettmers2022int8}. 
These massive values appear systematically across attention scores, weights, and activations \citep{an2025systematic}. 
\citet{sun2024massive} analyzed the role of massive activations as attention biases, while \citep{zhang2024unveiling} focused on weight outliers in LayerNorm layers, demonstrating their importance for model behavior.
Prior works mainly treated massive activations as artifacts to be suppressed or regularized using techniques such as pruning or quantization \citep{xiao2023smoothquant, wei2023outliersupp, yao2022zeroquant}.
{While several studies explore the structural alignment of massive activations with specific dimensions \citep{elhage2023privileged} or view them as task-specific linear classifiers \citep{rudman2023encode}, their intricate semantic functions remain underexplored.}
In contrast, we focus on interpretable attributes of massive activation, highlighting their semantic role and interpretable view.

\paragraph{Activation steering.} 
Activation steering refers to the direct manipulation of internal activations to control model behavior \citep{rimsky2024steering, zou2023representation}.
Existing approaches vary by their intervention sites and strategies to optimize this control.
\citet{li2023inference} focused on manipulating attention heads to enhance truthfulness, whereas \citet{turner2023steering} demonstrated that intervening on the residual stream effectively governs sentiment and toxicity.
Advancing this paradigm, \citet{lee2024cast} introduced conditional activation steering, which dynamically applies interventions based on specific conditions.
We distinguish our approach by selectively steering the highest activated dimensions for each domain, focusing on the key hidden state components responsible for domain-specific control.

\section{Conclusion}
In this work, we focus on the inherent anisotropy of LLM internal activations, demonstrating that the disproportionate utilization of the embedding space leads to the emergence of specialized dimensions that capture domain-specific features. 
Specifically, we propose a simple statistical way to identify Domain-Critical Dimensions directly from the pre-trained weights without additional training, and verify their semantic interpretability with comprehensive analyses. 
We additionally show that targeted interventions with those dimensions can match or even surpass the performance of whole-space steering.
We believe our findings offer a new perspective on navigating the complex internal landscape of LLMs, suggesting direction for more transparent and controllable AI systems.

\section*{Limitations}
While our results have shown that a sparse set of dominant dimensions can provide a practical control interface, we first acknowledge the inherent polysemanticity within LLM representations.
Interpretability outcomes depend on the granularity at which features are defined, and approaches aimed at explicit fine-grained factorization can offer a complementary view of the latent space. 
However, {we would like to clarify that our primary goal is not maximal feature separability, but investigating whether human-interpretable feature patterns are naturally embedded in the model’s intrinsic representational dimensions to utilize them for effective model control.}

Second, the current implementation of Critical Dimension Steering (CDS) intervenes through a sparse subset of the representational space, which may not reflect every subtle, context-dependent variation in activation.
However, by restricting interventions to a minimal set of critical dimensions, CDS remains computationally lightweight relative to approaches that steer across the full space.
Furthermore, focusing on magnitude-dominant dimensions provides a stable control signal in our experiments and reduces exposure to less consequential variation.

{Lastly, our method can be limited in tasks where domain boundaries are ambiguous. 
For domains with distinct characteristic signatures, our method successfully captures meaningful signals as demonstrated by our experiments. 
But in tasks like creative writing or general summarization, semantic features heavily overlap and evaluation is inherently subjective, making it difficult to isolate domain-critical dimensions.}
Thus, an important direction for future work is to explore adaptive or context-aware mechanisms for selecting dimensions and steering strengths, which may further improve the specificity of sparse interventions while preserving their efficiency.

\section*{Broader Impact and Ethical Implications}
As LLMs increasingly influence decisions and public information ecosystems, alignment with human values is essential for ensuring trustworthy behavior, accountability, and user safety.
Aligning LLMs with human values requires a deep understanding of how information is processed within their internal activations.
While our goal is to contribute to more ethical and safety-oriented LLM research, our study includes experiments that uses harmful queries, including settings that may increase compliance with harmful requests, and this introduces a potential risk if misused.
We therefore emphasize that these experiments are conducted for controlled analysis of internal mechanisms and are intended to improve model transparency and controllability, enabling more reliable safeguards and more trustworthy outputs.
By tracing specific activation patterns and intervening directly in internal representations, our research provides a framework for enhancing the safety and factuality of AI outputs.
We wish for this study to contribute to the development of reliable AI technologies that are transparently steered to remain beneficial and ethically grounded for all users.

\section*{Acknowledgments}

Youngji and Jaehyung are affiliated with the Department of Artificial Intelligence at Yonsei University.
This research was supported in part by Institute for Information \& communications Technology Planning \& Evaluation (IITP) grant funded by the Korea government (MSIT) (No. RS-2020-II201361, Artificial Intelligence Graduate School Program (Yonsei University); No. RS-2025-02215344, Development of AI Technology with Robust and Flexible Resilience Against Risk Factors; No. RS-2025-25442405, Development of a Self-Learning World Model-Based
AGI System for Hyperspectral Imaging).

\bibliography{custom}
\appendix
\clearpage
\begin{figure*}[b]
    \centering
    \begin{subfigure}[t]{0.27\textwidth}
        \centering
        \includegraphics[width=\linewidth]{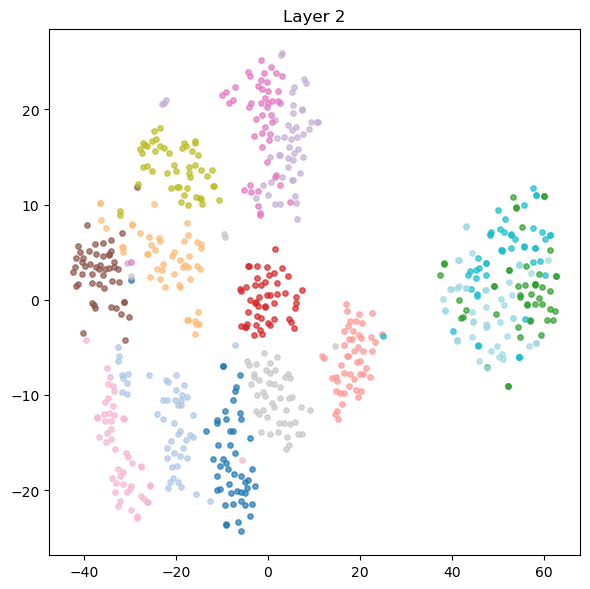}
    \end{subfigure}
    \hfill
    \begin{subfigure}[t]{0.27\textwidth}
        \centering
        \includegraphics[width=\linewidth]{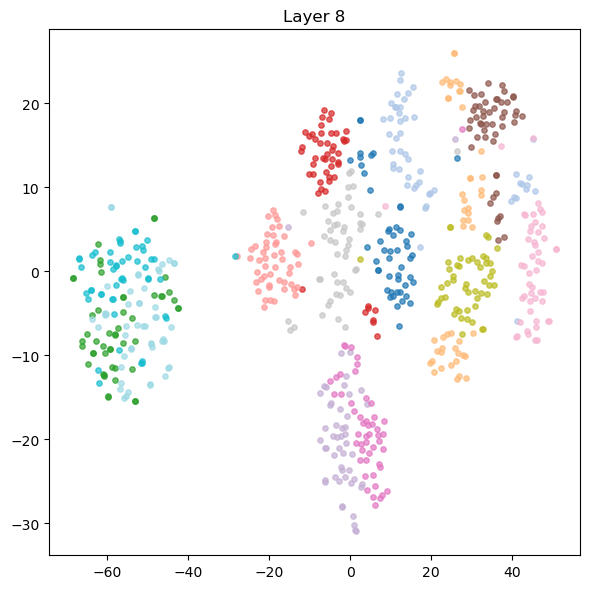}
    \end{subfigure}
    \hfill
    \begin{subfigure}[t]{0.27\textwidth}
       \centering
        \includegraphics[width=\linewidth]{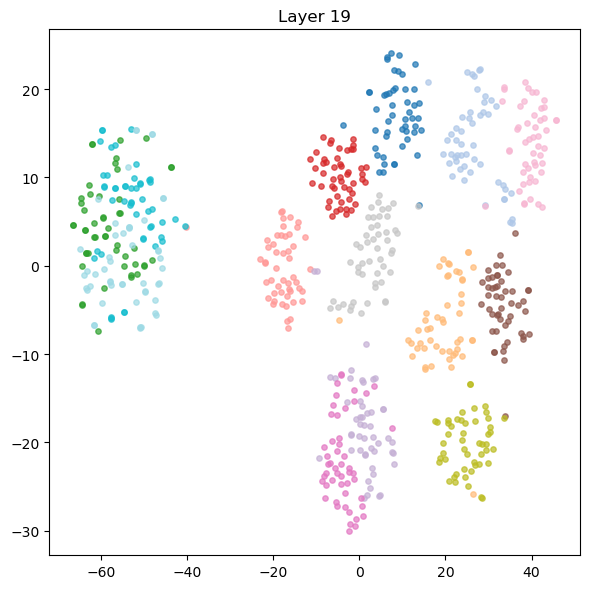}
    \end{subfigure}
    \hfill
    \begin{subfigure}[t]{0.15\textwidth}
    \centering
    \raisebox{0.7in}{ 
        \includegraphics[width=\linewidth]{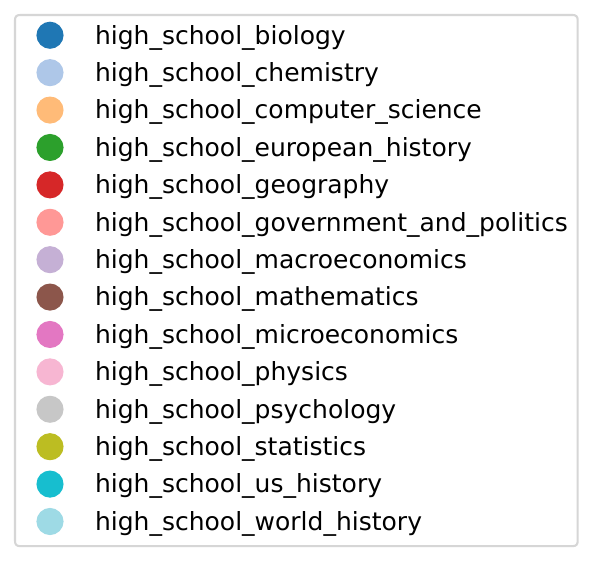}
    }
    \end{subfigure}
    \vspace{-0.1in}
    \caption{Visualization of hidden state representations for 14 high school subjects. The dimensionality reduction was performed using t-SNE with cosine metric.} 
    \label{fig:figure_8}
    \vspace{-0.1in}
\end{figure*}

\section{Additional Analyses}\label{supp:A}
In this section, we provide additional analysis results.
\vspace{-0.2cm} 

\subsection{Single Dimension Masking effect in Other Domains}
\label{supp:A1}
\begin{figure*}[p]
    \centering
    \begin{minipage}[b]{0.48\textwidth}
        \centering
        \includegraphics[width=\linewidth]{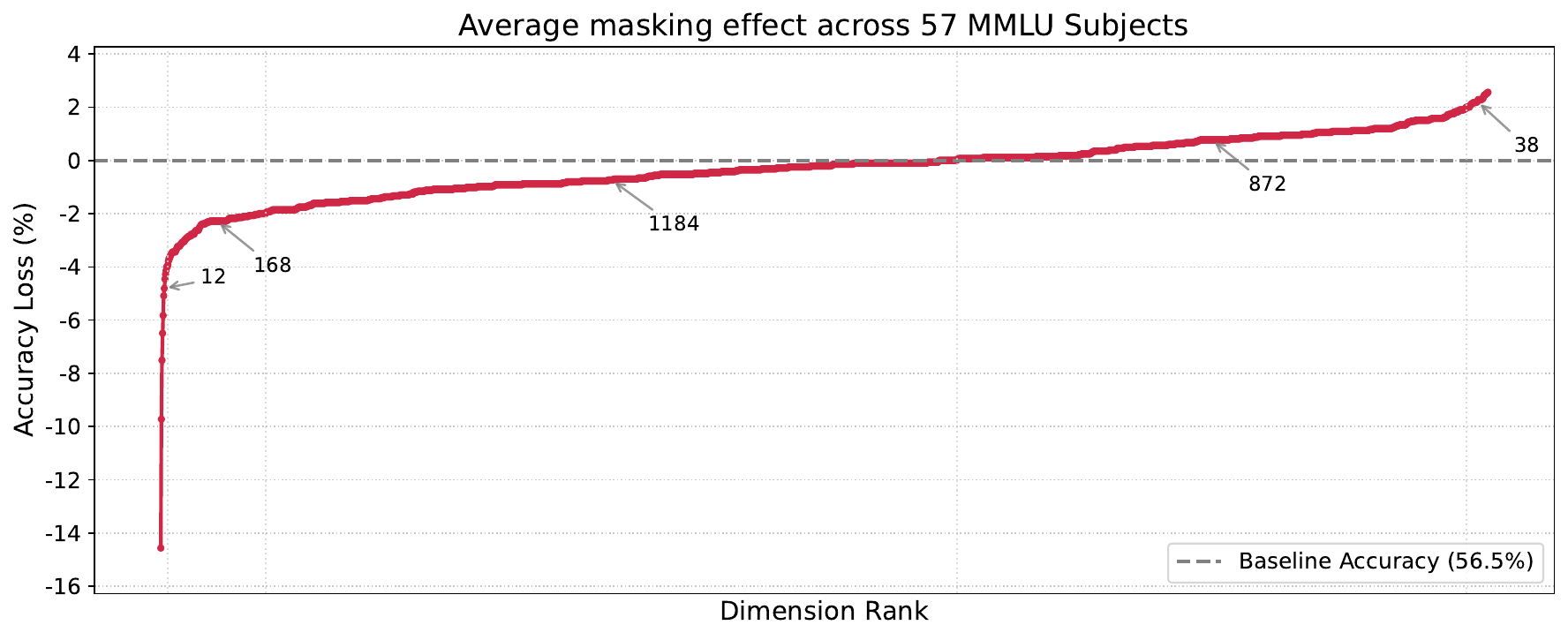}
        \caption{Rankwise average accuracy drop of \textit{functionally critical dimensions.}}
        \label{fig:figure_9}
    \end{minipage}
    \hspace{4mm}
    \begin{minipage}[b]{0.47\textwidth}
        \centering
        \includegraphics[width=\linewidth]{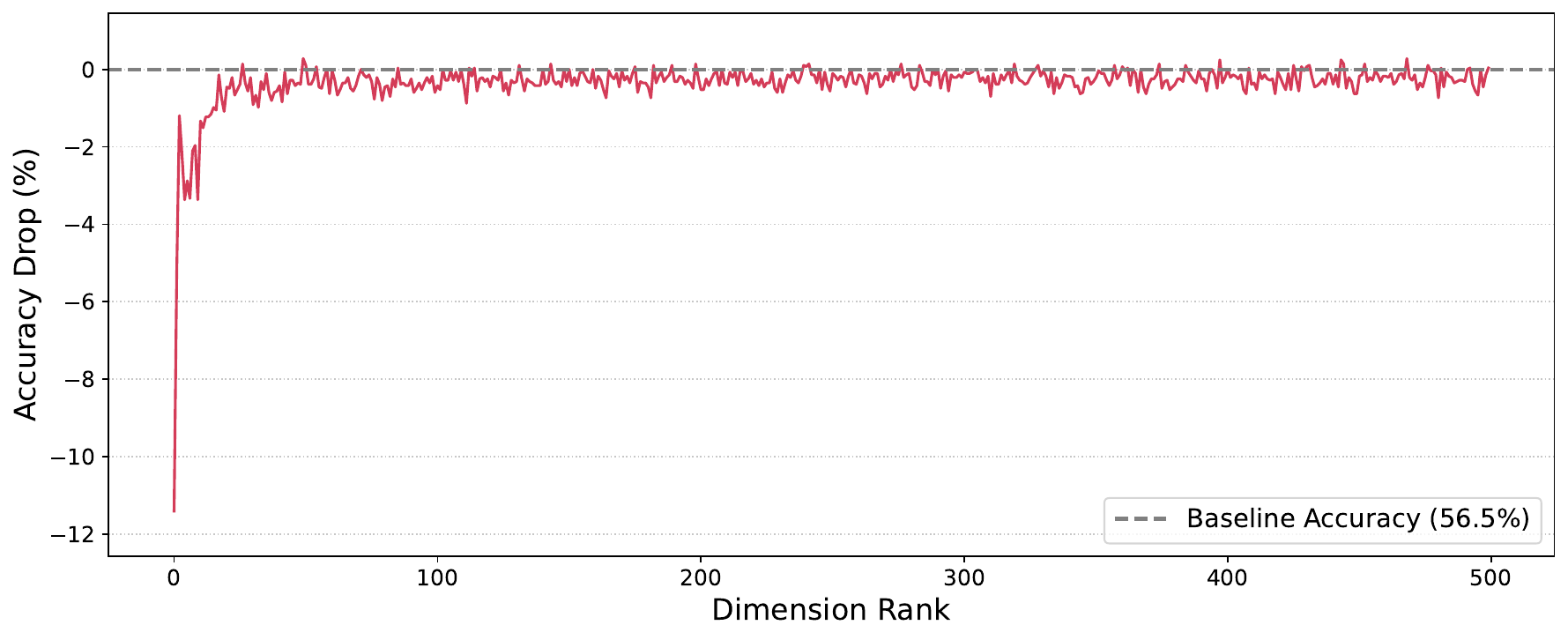}
        \vspace{-6mm}
        \caption{Rankwise average accuracy drop of our \textit{domain-critical dimensions}.}
        \label{fig:figure_10}
    \end{minipage}
    \begin{subfigure}[b]{0.49\textwidth}
        \includegraphics[width=\linewidth]{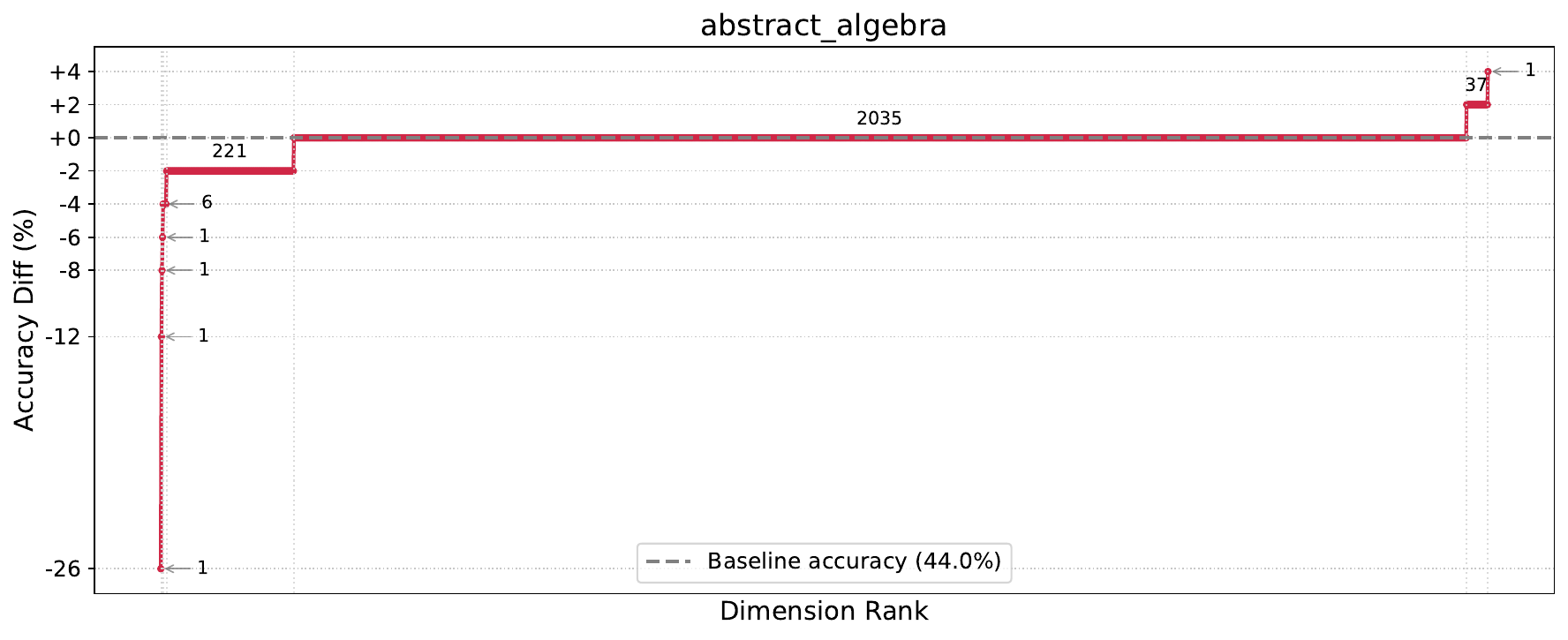}
    \end{subfigure}
    \hfill
    \begin{subfigure}[b]{0.49\textwidth}
        \includegraphics[width=\linewidth]{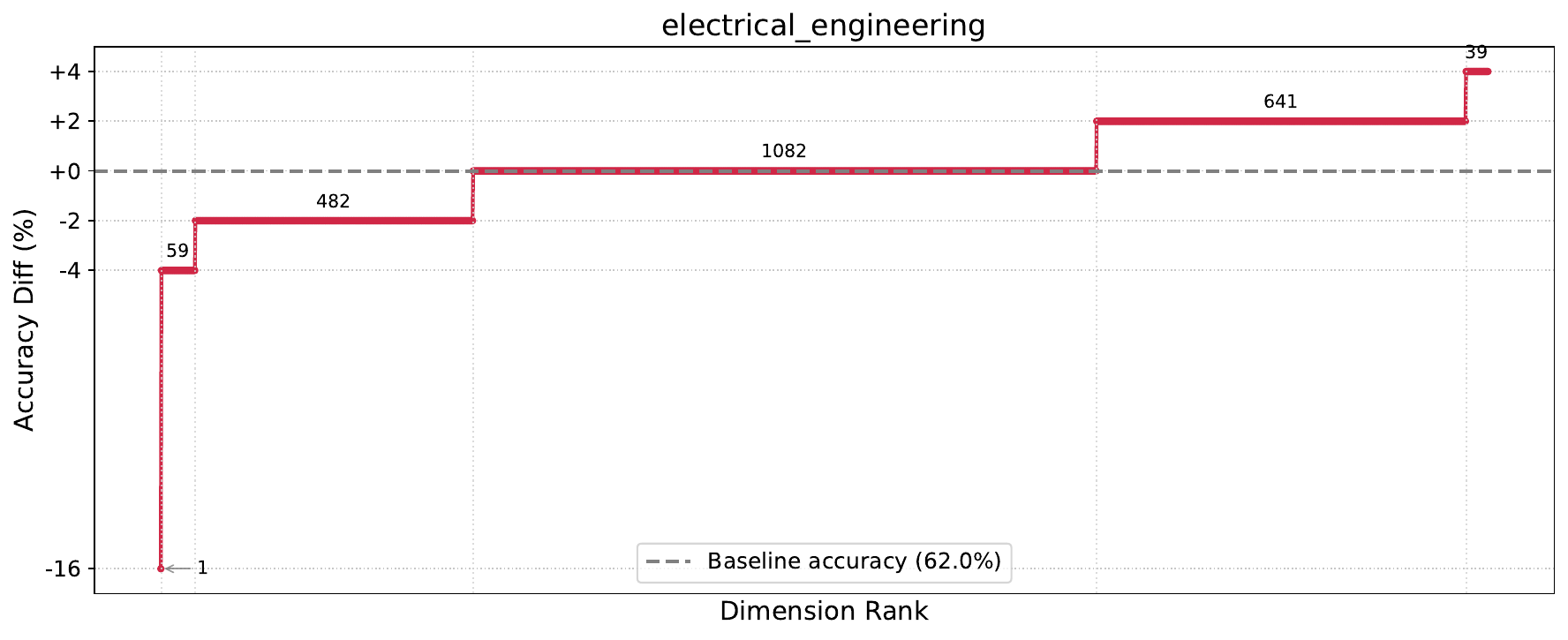}
    \end{subfigure}

    \begin{subfigure}[b]{0.49\textwidth}
        \includegraphics[width=\linewidth]{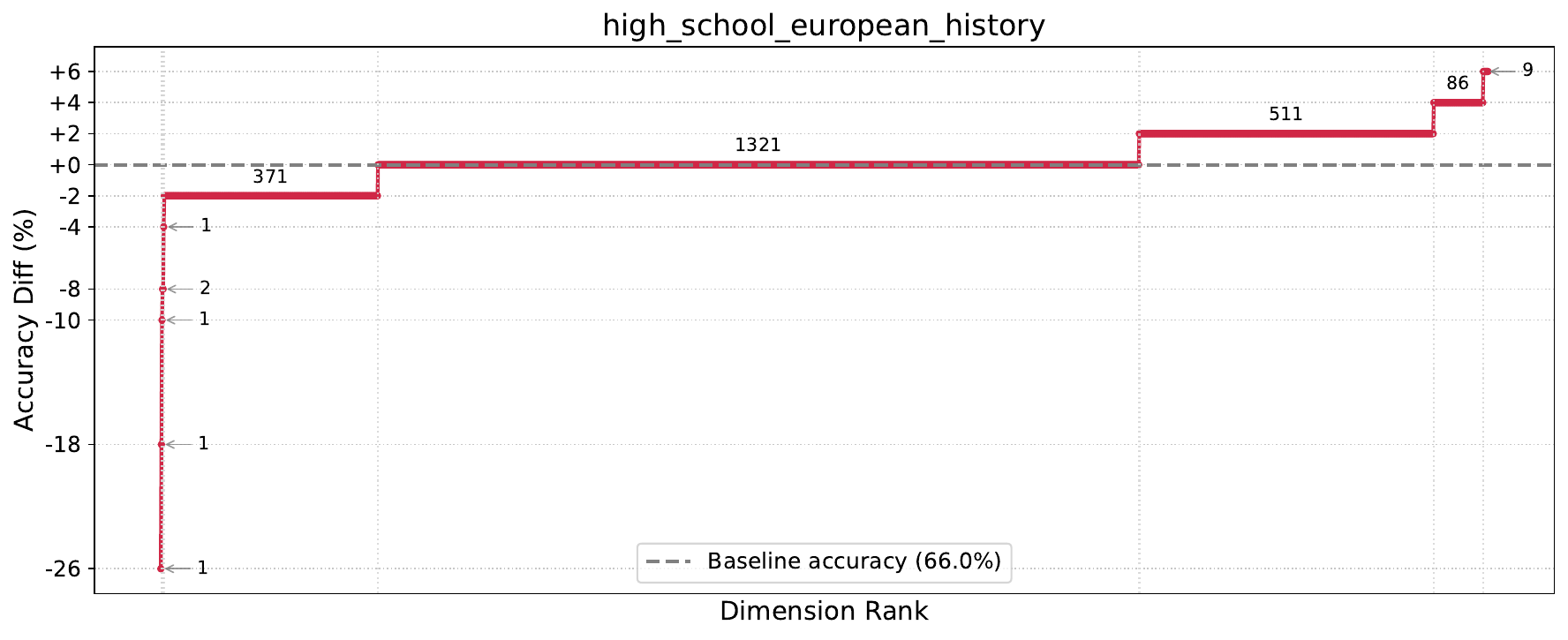}
    \end{subfigure}
    \hfill
    \begin{subfigure}[b]{0.49\textwidth}
        \includegraphics[width=\linewidth]{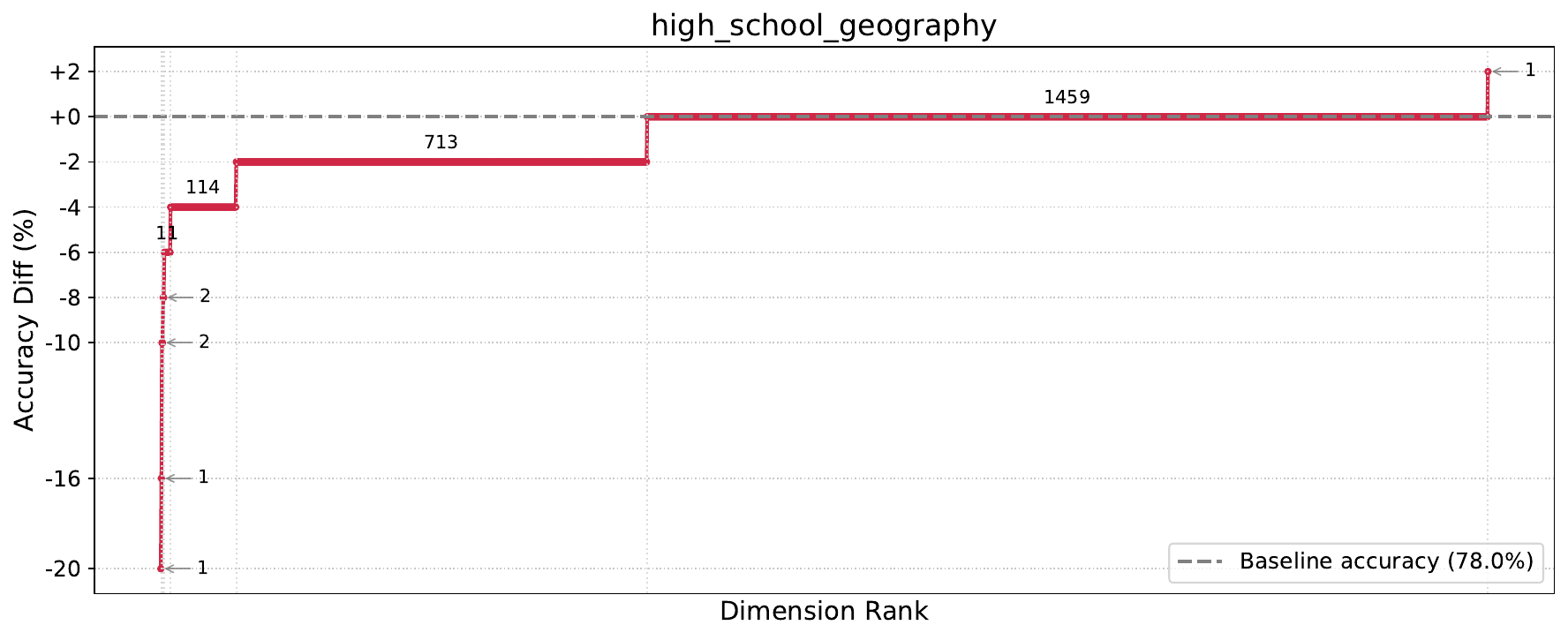}
    \end{subfigure}

    \begin{subfigure}[b]{0.49\textwidth}
        \includegraphics[width=\linewidth]{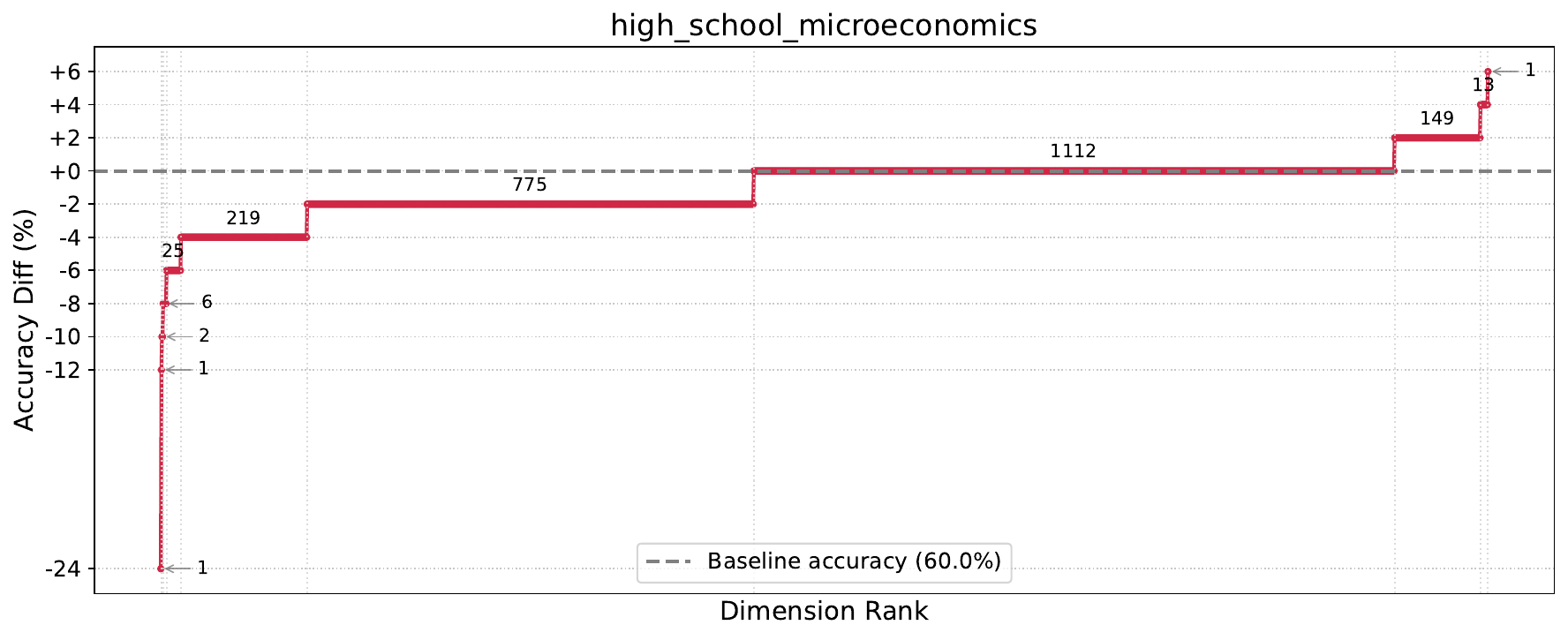}
    \end{subfigure}
    \hfill
    \begin{subfigure}[b]{0.49\textwidth}
        \includegraphics[width=\linewidth]{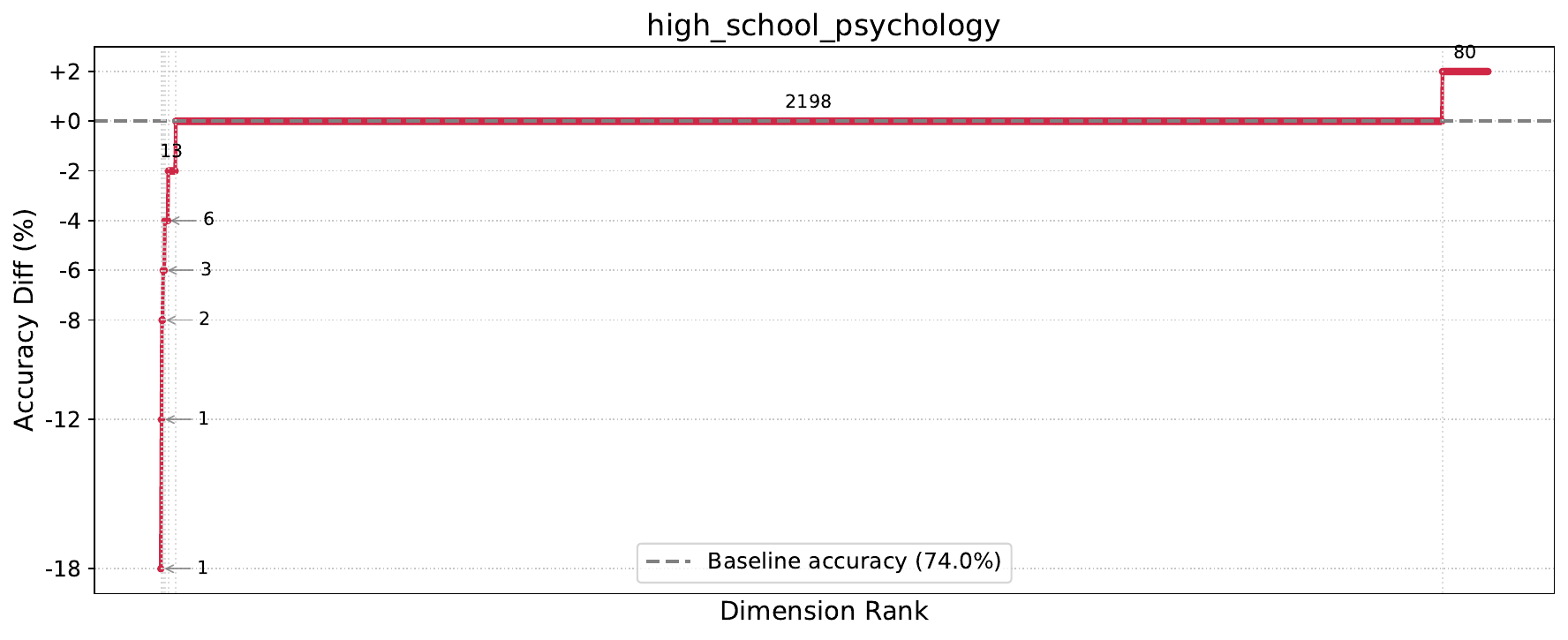}
    \end{subfigure}

    \begin{subfigure}[b]{0.49\textwidth}
        \includegraphics[width=\linewidth]{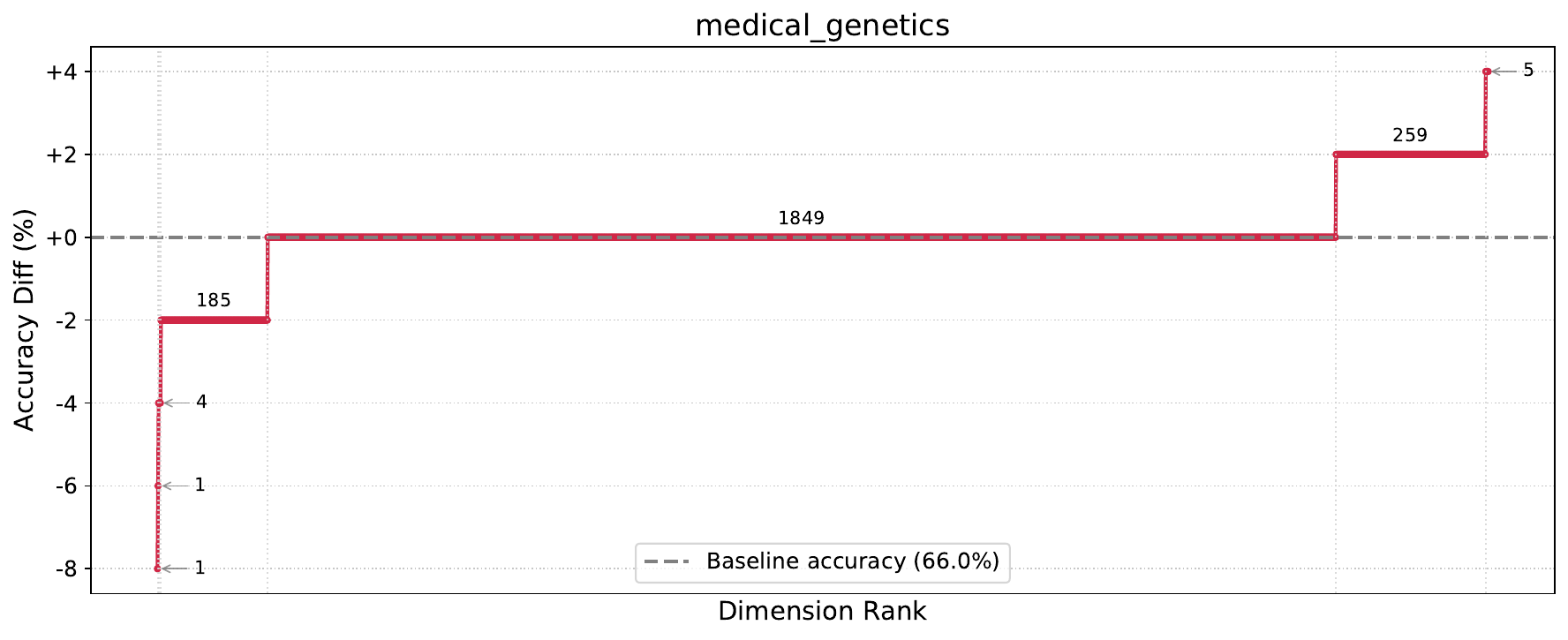}
    \end{subfigure}
    \hfill
    \begin{subfigure}[b]{0.49\textwidth}
        \includegraphics[width=\linewidth]{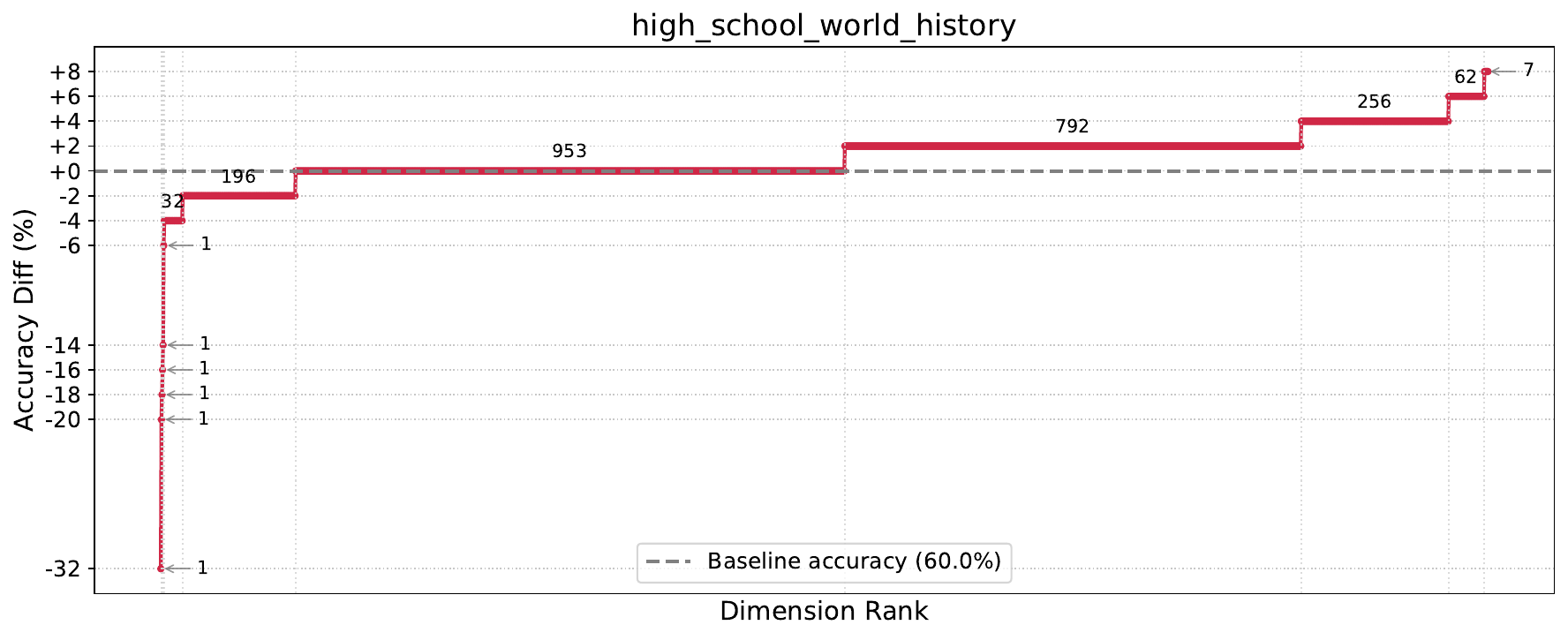}
    \end{subfigure}

    \begin{subfigure}[b]{0.49\textwidth}
        \includegraphics[width=\linewidth]{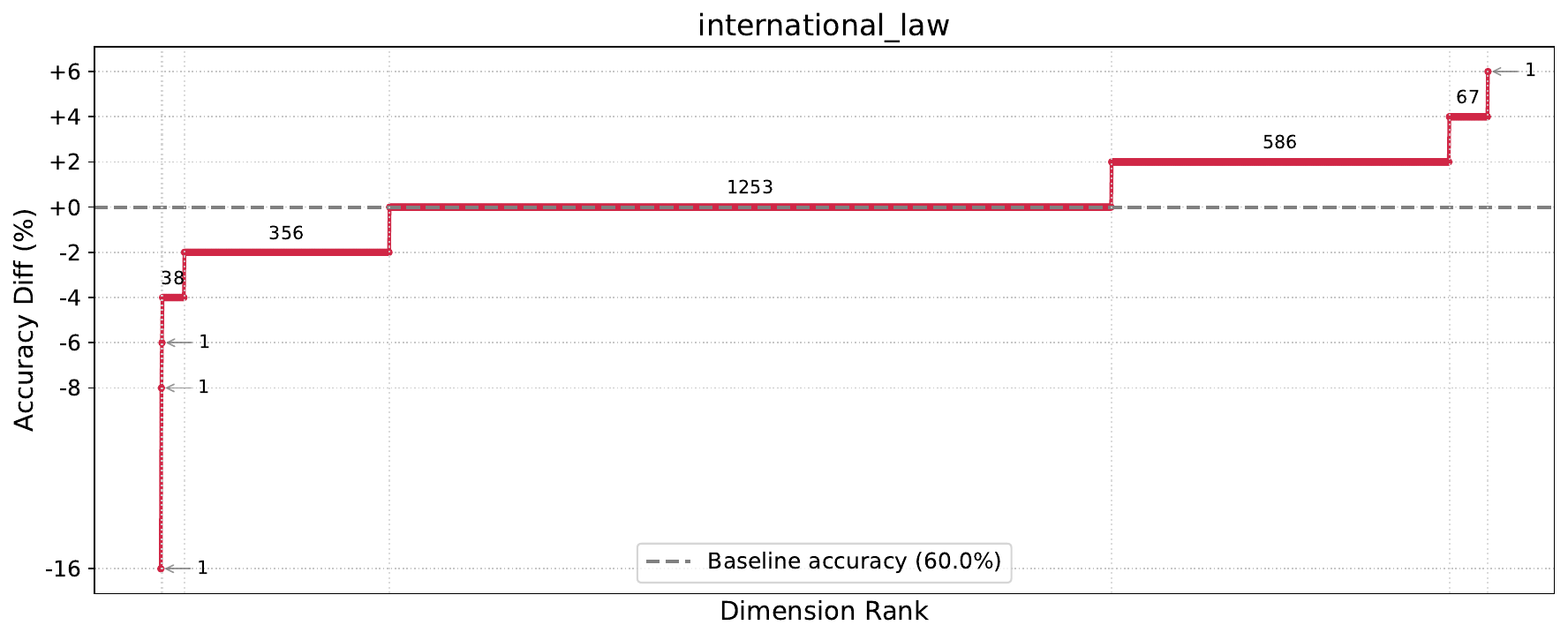}
    \end{subfigure}
    \hfill
    \begin{subfigure}[b]{0.49\textwidth}
        \includegraphics[width=\linewidth]{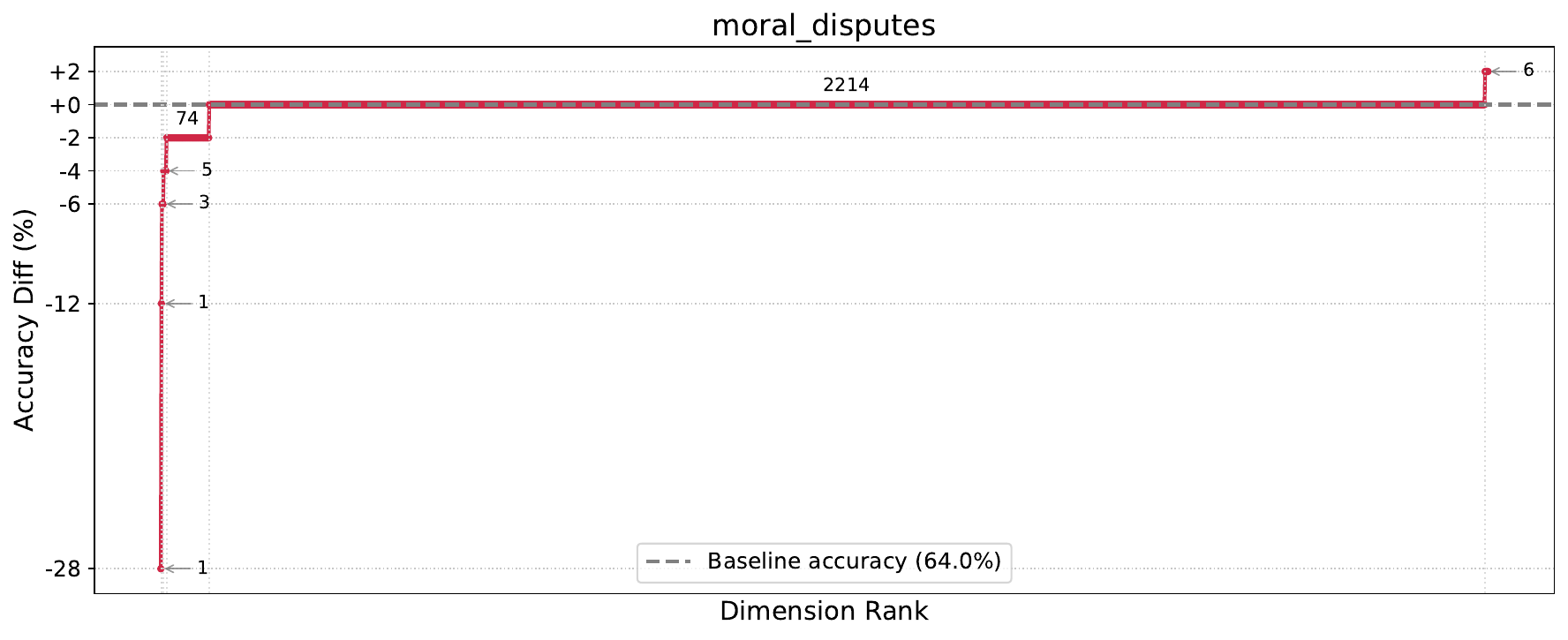}
    \end{subfigure}

    \begin{subfigure}[b]{0.49\textwidth}
        \includegraphics[width=\linewidth]{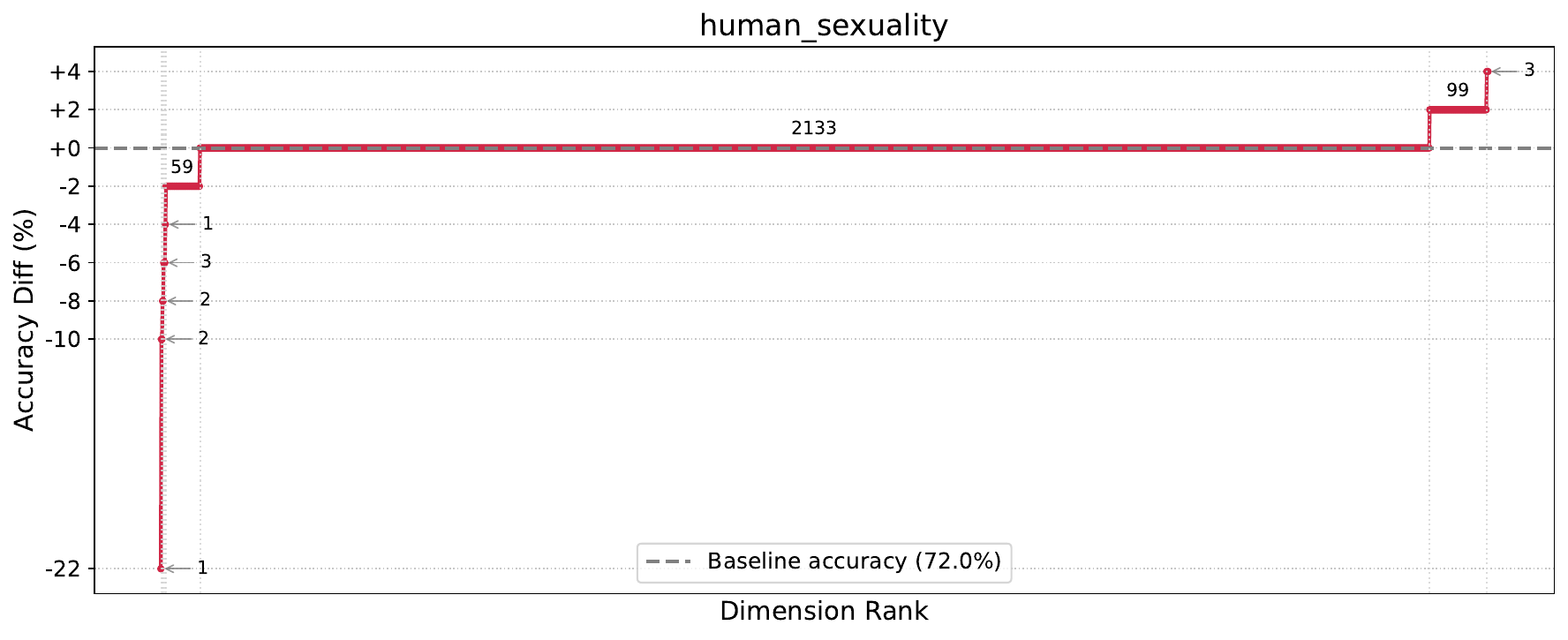}
    \end{subfigure}
    \hfill
    \begin{subfigure}[b]{0.49\textwidth}
        \includegraphics[width=\linewidth]{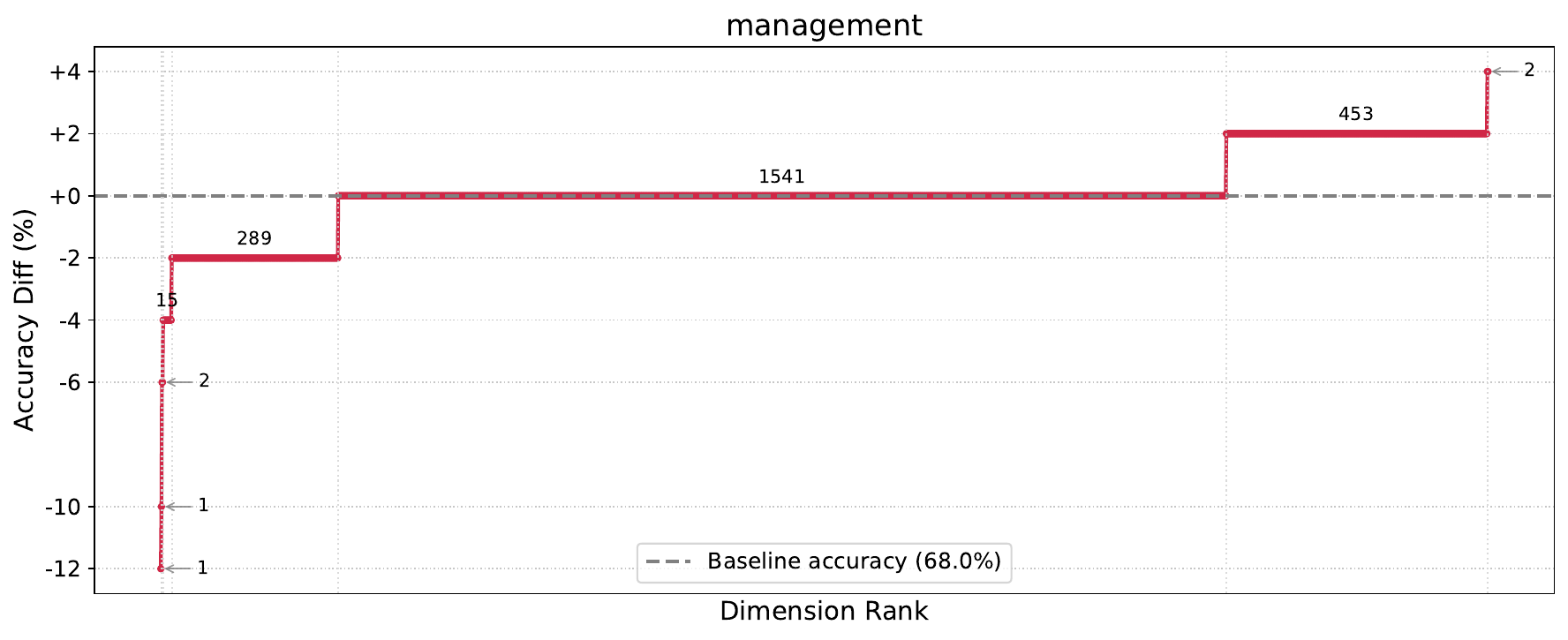}
    \end{subfigure}
    
    \caption{
    Single dimension masking results across various MMLU subjects. 
    Individual subjects exhibit a sparse set of critical dimensions that significantly impact performance.
    }
    \label{fig:figure_11}
\end{figure*}

We extend the analysis from \S\ref{sec:2.1} to all 57 MMLU subjects to verify the universality of functional criticality.
Figure \ref{fig:figure_9} presents the rank-wise accuracy drop averaged across all 57 subjects. 
Additionally, Figure \ref{fig:figure_11} illustrates the distributions for 12 randomly selected subjects due to space limitations.
The aggregated view confirms a distinct sharp head and long tail distribution.
Specifically, only 12 out of 2304 average dimensions yield an accuracy drop of at least 4\%, with the top rank causing a 14.6\% drop.
In contrast, the vast majority of dimensions have negligible impact, falling within narrow bands: 168 in $[-4,-2)$, 1184 in $[-2,0)$, 872 in $[0,2)$, and 38 in $\ge$ 2\%.
These results confirm that performance in every domain relies on a sparse set of critical dimensions, regardless of the subject.

\vspace{-0.2cm} 

\subsection{Domain Separation in Representational Space}
\label{supp:A2}
We performed a qualitative analysis using 14 high school-related MMLU subjects, to explore the internal representations of different subjects.
We extracted the hidden states from the identification set queries and applied t-Distributed Stochastic Neighbor Embedding (t-SNE) with a cosine metric for dimensionality reduction.
Figure \ref{fig:figure_8} illustrates the representations across different depths of the model (Layers 2, 8, and 19).
We observe that the hidden states of each subject naturally form distinct subject-specific clusters, continued throughout the layers.
This demonstrates that the model intrinsically encodes domain identity within its activation space.

\subsection{Validation Between \texorpdfstring{$\mathcal{I}_\text{dcd}$}{Idcd} and Functionally Critical Dimensions}
\label{supp:A3}
\paragraph{Validation by recall.}
For each domain, we define the ground-truth set of functionally critical dimensions $\mathcal{I}^{(k)}_{\text{mask}}$ using the top-ranked dimensions from the masking experiments up to a rank cutoff $N$.
We measure recall as the overlap ratio $|\mathcal{I}_{\text{dcd}} \cap \mathcal{I}^{(k)}_{\text{mask}}| / k$.
Note that to ensure strict validity, we exclude dimensions tied at the cutoff boundary, so the size of ground-truth set $k$ may be slightly smaller than $N$.
With $N=10$, our method achieves an subject average recall of 89.39\%, demonstrating that simple activation statistics effectively recover functionally critical dimensions.
As shown in Table \ref{tab:table_2}, the recall naturally decreases as the cutoff $N$ increases due to the expansion of the ground-truth set.
\begin{table}[h]
    \centering
    \caption{Recall of the top-100 statistical dimensions across varying rank cutoffs ($N$).}
    \label{tab:table_2}
    \vspace{-0.1in}
    \small{
    \begin{tabular}{lcccc}
        \toprule
        Rank Cutoff ($N$) & 5 & 10 & 15 & 20 \\
        \midrule
        Recall@$K$ (\%) & 94.91 & 89.39 & 82.69 & 78.88 \\
        \bottomrule
    \end{tabular}
    \vspace{-0.1in}
    }
\end{table}
\paragraph{Validation by masking effect.}
We further validate \idcd by checking accuracy drop when masking them.
The degradation trajectory shown in Table \ref{tab:table_3} and Figure \ref{fig:figure_10} closely mirrors the ground-truth masking effects observed in Table \ref{tab:table_1} and Figure \ref{fig:figure_9}.
Masking rank-1 \idcd yields an 11.40\% drop, while the impact decays rapidly for lower ranks.
This confirms that $s_j$ effectively isolates the sparse set of functionally critical dimensions.
\begin{table}[h]
    \begin{center}
    \caption{Impact of masking individual dimensions in $\mathcal{I}_\text{dcd}$. Rank $k$ indicates the average accuracy when masking the $k$-th dimension.}
    \vspace{-0.1in}
    \resizebox{1.0\linewidth}{!}{
        \begin{tabular}{l| ccccc}
            \toprule
            {Model} & \textbf{Rank 1} & \textbf{Rank 2} & \textbf{Rank 5} & \textbf{Rank 10} & \textbf{Rank 100} \\ 
            \midrule
            \textbf{Gemma-2-2B-IT} & 45.13 & 51.30 & 53.16 & 53.16 & 55.93 \\ 
            \hfill Acc. Drop (\%) & \textbf{(-11.40)} & (-5.23) & (-3.37) & (-3.37) & (-0.60) \\
            \midrule
            {\textbf{Qwen-3-8B}} & 35.75 & 23.39 & 73.12 & 72.21 & 72.57 \\ 
            \hfill Acc. Drop (\%) & \textbf{(-37.55)} & \textbf{(-49.91)} & (-0.18) & (-1.09) & (-0.73) \\
            \bottomrule
        \end{tabular}
    }
    \label{tab:table_3}
    \end{center}
    \vspace{-0.1in} 
\end{table}


\clearpage
\subsection{Statistics of Domain-Critical Dimensions}
\label{supp:A4}
\paragraph{Statistical extremity.}
In \S\ref{sec:2.2}, we hypothesized that functionally critical dimensions are characterized by extreme activation values. 
To quantify this extremity for identified \idcd, we analyze the activation magnitude distribution of identified dimensions in the \textit{high school biology}. 
Figure \ref{fig:figure_12} and Table \ref{tab:table_4} presents the average magnitude of the top-$k$ identified dimensions alongside the individual magnitude at specific ranks.
The results demonstrate distinct magnitude disparities of the the identified dimensions.
The top-1 dimension exhibits a magnitude of 115.16, which is approximately $100\times$ larger than the total dimension average of 1.17.
Even within the top-100 dimensions (our selection for $\mathcal{I}_{\text{dcd}}$), the average magnitude remains at 5.96, approximately $5\times$ the baseline.
\begin{table}[h]
    \centering
    \caption{{Magnitude statistics of domain-critical dimensions.}
     We report top-$k$ dimension set average and rank-$k$ individual magnitudes for the \textit{High School Biology} subject.}
    \label{tab:table_4}
    \resizebox{\columnwidth}{!}{
        \begin{tabular}{l|ccccccccc}
            \toprule
            \textbf{\idcd Rank ($k$)} & \textbf{1} & \textbf{5} & \textbf{10} & \textbf{50} & \textbf{100} & \textbf{200} & \textbf{300} & \textbf{400} & \textbf{500} \\
            \midrule
            Top-$k$ Avg. & 115.16 & 44.83 & 28.47 & 9.71 & 5.96 & 3.81 & 3.01 & 2.59 & 2.31 \\
            Rank $k$ Single & 115.16 & 16.96 & 10.97 & 2.95 & 1.88 & 1.51 & 1.35 & 1.26 & 1.19 \\
            \bottomrule
        \end{tabular}
    }\par
    \vspace{0.2in}
    \includegraphics[width=\columnwidth]{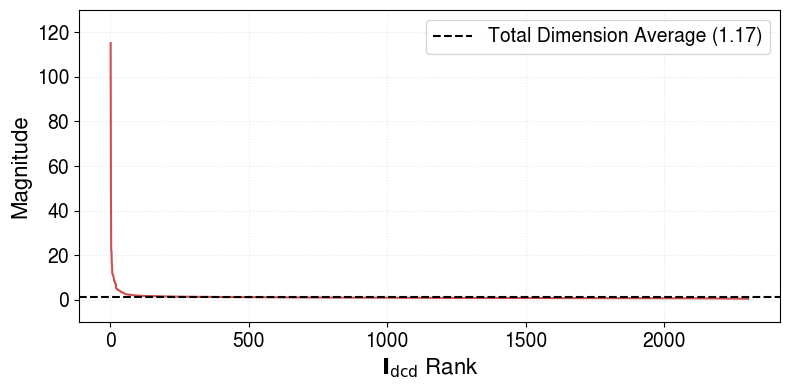}
    \vspace{-7mm}
    \captionof{figure}{{Distribution of activation magnitudes.} A small subset of dimensions exhibits significantly higher magnitudes compared to the total dimension average.}
    \label{fig:figure_12}
\end{table}

\newpage
{\paragraph{Subject specificity on masking.}
{We conducted experiments to confirm whether the masking effects of critical dimensions are indeed subject-specific. 
Instead of solely ablating dimensions based on their importance rank within the target task, we ablated dimensions of an equivalent rank extracted from a randomly selected different task. 
As shown in the table below, removing a target task-specific critical dimension causes a significantly larger performance drop than masking an equally ranked dimension from a random task. 
This confirms that the impact of these ablations is task-specific, rather than a general degradation from masking any highly activated dimension.}
\begin{table}[h]
    \centering
    \small
    \caption{{Comparison of accuracy drop between task-specific and task-random dimensions.}}
    \vspace{-2mm}
    \resizebox{\columnwidth}{!}{
    \begin{tabular}{l|ccccc}
        \toprule
        \textbf{Metric} & \textbf{Rank 3} & \textbf{Rank 5} & \textbf{Rank 10} & \textbf{Rank 15} & \textbf{Rank 20} \\
        \midrule
        \textbf{Task-Specific (Ours)} & 51.30 & 53.16 & 53.16 & 55.37 & 55.44 \\
        \hspace{3mm} Accuracy Drop (\%) & ($-5.23$) & ($-3.37$) & ($-3.37$) & ($-1.16$) & ($-1.09$) \\
        \midrule
        \textbf{Task-Random (Control)} & 54.67 & 54.64 & 54.78 & 55.79 & 56.32 \\
        \hspace{3mm} Accuracy Drop (\%) & ($-1.86$) & ($-1.89$) & ($-1.75$) & ($-0.74$) & ($-0.21$) \\
        \bottomrule
    \end{tabular}}
    \label{tab:task}
\end{table}

\paragraph{Subject specificity on variances.}
{We measured the variances and cross-task ranges of domain-critical dimensions. 
As shown, the Math-specific dimension (Dim 1046) exhibits significantly higher variance within the \textit{high school mathematics} subject compared to other domains. 
The Biology and CS-critical dimensions demonstrate similar domain-specific variance patterns.}
\begin{table}[h]
    \centering
    \small
    \caption{{Variances of domain-critical dimensions across different datasets.}}
    \vspace{-2mm}
    \resizebox{\columnwidth}{!}{
    \begin{tabular}{l|c|ccc}
        \toprule
        \textbf{Domain} & \textbf{ID} & \textbf{HS Math} & \textbf{HS Bio} & \textbf{HS Computer Sci.} \\
        \midrule
        Math & 1046 & \textbf{92.03} & 59.70 & 73.47 \\
        Bio & 2106 & 33.50 & \textbf{80.48} & 56.06 \\
        CS \hspace{1cm} & 1807 & 56.72 & 57.07 & \textbf{105.10} \\
        \bottomrule
    \end{tabular}}
    \label{tab:table_6}
\end{table}

\clearpage
\subsection{Additional Prompt-level Analysis}
\label{supp:A5}
\begin{figure*}[t]
    \centering
    \scriptsize
    \begin{minipage}[t]{\textwidth}
        \centering
        \begin{tabular}{@{} p{\linewidth} @{}}
        \toprule 
        \textbf{The following are multiple choice questions (with answers) about high school biology.}\\
        You must respond with a single alphabet character.\\[0.5em]
        Energy is harvested during cellular respiration in stages. Which of the following correctly states which phase of cellular respiration harvests the most energy and the correct explanation why? \\[0.5em]
        A. The most energy is released during the Krebs cycle because it is here that pyruvate is completely broken down into \ce{CO2}.\\
        B. The most energy is released during the Krebs cycle because in addition to the production of ATP, both \ce{FADH2} and NADH are produced. Each of those molecules will release 2 ATPs and 3 ATPs, respectively.\\
        C. The most energy is released during oxidative phosphorylation because in addition to the phosphorylation of ADP into ATP, all the potential energy held in NADH and FADH is transferred to ATP.\\
        D. The most energy is released during oxidative phosphorylation because \ce{H2O} is completely broken down into \ce{H+} and \ce{O2}. \\
        Answer: \\
        \bottomrule
        \end{tabular}
    \end{minipage}

    \begin{minipage}[t]{\textwidth}
        \centering
        \includegraphics[width=1\textwidth]{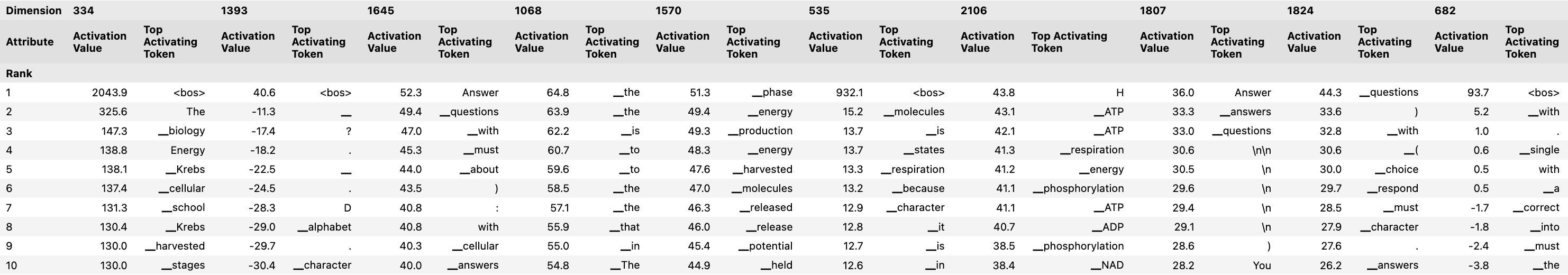}
    \end{minipage}
    \begin{minipage}[t]{\textwidth}
        \centering
        \begin{tabular}{@{} p{\linewidth} @{}}
        \toprule 
        \textbf{The following are multiple choice questions (with answers) about high school biology.}\\
        You must respond with a single alphabet character.\\[0.5em]
        Which of the following statements is not correct about lipids? \\[0.5em]
        A. Lipids consist of fatty acids and glycerol.\\
        B. Steroids are examples of lipids.\\
        C. The molecules of a saturated fatty acid are packed so close together that they form a solid at room temperature.\\
        D. The head of a phospholipid is hydrophobic, and the tails are hydrophilic. \\
        Answer: \\
        \bottomrule
        \end{tabular}
    \end{minipage}

    \begin{minipage}[t]{\textwidth}
        \centering
        \includegraphics[width=1\textwidth]{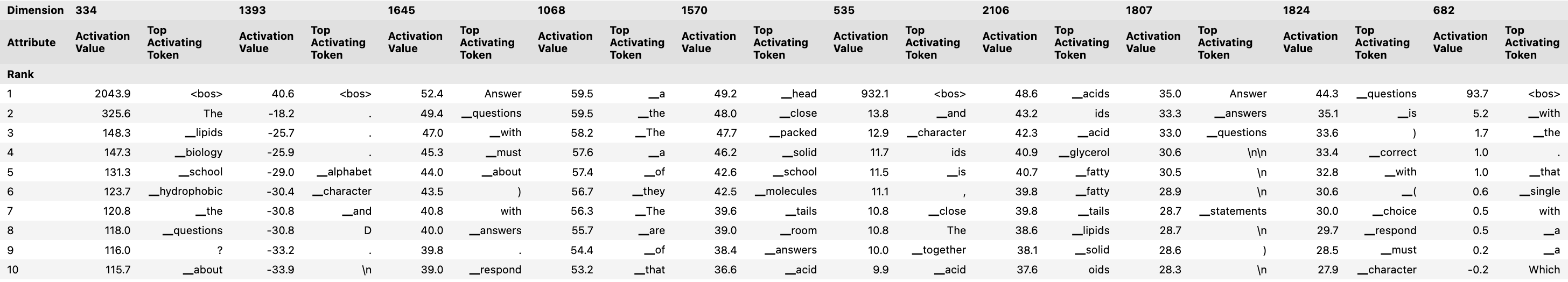}
    \end{minipage}
    \caption{Analysis of domain-critical dimensions on high school biology prompts.
        \textbf{(a)} and \textbf{(b)} display the top-10 activating tokens for a \textit{cellular respiration} related query and a \textit{lipid} related query.
        }
    \label{fig:figure_13}
    \vspace{-0.3cm}
\end{figure*}
To further extend the qualitative findings in \S\ref{sec:3}, we provide an extended prompt-level analysis on high school biology and high school mathematics. 
These results consistently demonstrate that the specialized semantic roles of the identified \idcd remain robust across diverse queries.

\paragraph{High school biology.}
We display the full list of the highest-10 activating tokens for two queries, a \textit{cellular respiration} related query illustrated at Figure \ref{fig:figure_4} and a \textit{lipids} related query as additional example.
Figure \ref{fig:figure_13} visualizes the activation patterns for these queries.
Comparing the two queries reveals that domain-critical dimensions exhibit specialized roles that are consistent across different biological topics:
\begin{itemize}[leftmargin=3.5mm, topsep=3pt, itemsep=3pt, parsep=0pt]
    \item \textbf{Dimension 334} (Domain \& Topic Marker): This dimension consistently identifies the broad domain and the specific topic. In both queries, it strongly activates on the token \btok{\_biology}, signaling the subject matter. Simultaneously, it adapts to the specific context, capturing \btok{\_Krebs} and \btok{\_cellular} in the cellular respiration query, while shifting to \btok{\_lipids} and \btok{\_hydrophobic} in the lipids query.
    \item \textbf{Dimensions 1645} (Task Structure \& Instruction): This dimension focus on the structural scaffolding of the multiple-choice task rather than biological content. It shows high activations for \btok{Answer}, \btok{\_questions}, \btok{\_must}, effectively tracking the format and instructions of the prompt.
    \item \textbf{Dimension 1068} (Functional Syntax): This dimension consistently targets functional words. Regardless of the query content, the top activating tokens remain syntactic connectors such as \btok{\_the}, \btok{\_of}, \btok{\_is}, and \btok{\_to}.
    \item \textbf{Dimension 2106} (Biological Lexicon): As a fine-grained content extractor, this dimension targets specific biological entities. It transitions from activating \btok{\_ATP} and \btok{\_phosphorylation} in the cellular respiration query to \btok{\_fatty}, \btok{\_glycerol}, and \btok{\_acid} in the lipids query.
\end{itemize}
\newpage
\paragraph{High school mathematics.}
\begin{figure*}[t]
    \centering
    \scriptsize
    \begin{minipage}[t]{\textwidth}
        \centering
        \begin{tabular}{@{} p{\linewidth} @{}}
        \toprule 
        \textbf{The following are multiple choice questions (with answers) about high school mathematics.}\\
        You must respond with a single alphabet character.\\[0.5em]
        Suppose 5 different integers are randomly chosen from between 20 and 69, inclusive. What is the probability that they each have a different tens digit? \\[0.5em]
        A. \textbackslash frac\{1\}\{4\} \\
        B. \textbackslash frac\{1\}\{3\} \\
        C. \textbackslash frac\{1000\}\{52969\} \\
        D. \textbackslash frac\{2500\}\{52969\} \\
        Answer: \\
        \bottomrule
        \end{tabular}
    \end{minipage}

    \begin{minipage}[t]{\textwidth}
        \centering
        \includegraphics[width=\textwidth]{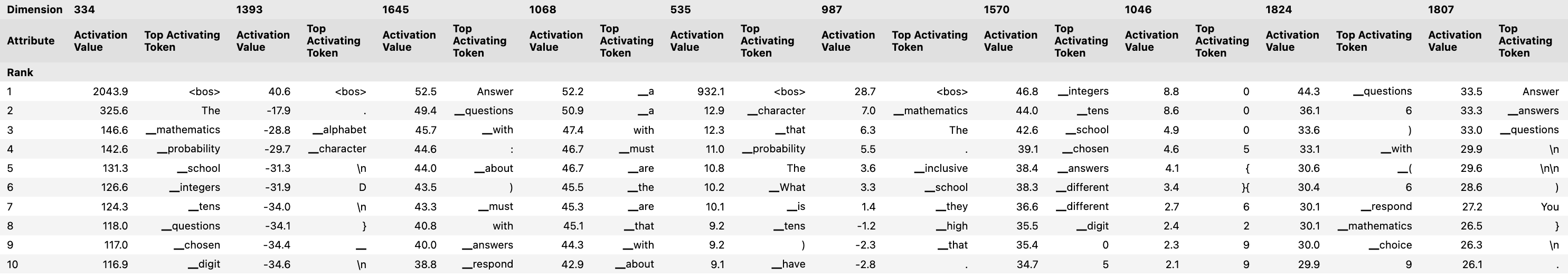}
    \end{minipage}
    \begin{minipage}[t]{\textwidth}
        \centering
        \begin{tabular}{@{} p{\linewidth} @{}}
        \toprule 
        \textbf{The following are multiple choice questions (with answers) about high school mathematics.}\\
        You must respond with a single alphabet character.\\[0.5em]
        An equilateral triangle has sides of 12 inches. What is the approximate area of the triangle? \\[0.5em]
        A. 62 \\
        B. 72 \\
        C. 84 \\
        D. 112 \\
        Answer: \\
        \bottomrule
        \end{tabular}
    \end{minipage}
    \begin{minipage}[t]{\textwidth}
        \centering
        \includegraphics[width=\textwidth]{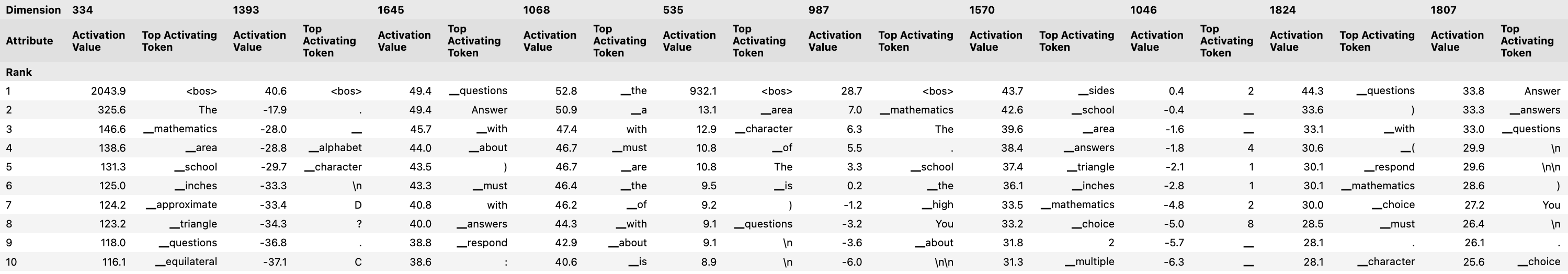}
    \end{minipage}
    \vspace{-0.3cm}
    \caption{Analysis of domain-critical dimensions on high school mathematics prompts.
        \textbf{(a)} and \textbf{(b)} display the top-10 activating tokens for a \textit{tens digit} related query and an \textit{equilateral triangle} related query, respectively.
        }
    \label{fig:figure_14}
\end{figure*}
We provide additional query samples sourced from high school mathematics.
Figure \ref{fig:figure_14} presents the highest-10 activating tokens for a \textit{tens digit} related query (Figure \ref{fig:figure_14}a) and an \textit{equilateral triangle} related query (Figure \ref{fig:figure_14}b).
Similar to the observations in the biology domain, specific dimensions in the mathematics dataset demonstrate distinct semantic responsibilities:
\begin{itemize}[leftmargin=3.5mm, topsep=3pt, itemsep=3pt, parsep=0pt]
    \item \textbf{Dimension 334} (Domain \& Topic Marker): Just as it identified \btok{\_biology} in the previous section, dimension 334 consistently captures the domain marker \btok{\_mathematics} in both queries. Furthermore, it adapts to the specific query context, prioritizing \btok{\_tens}, \btok{\_digits} and \btok{\_integers} in the tens digit query, while shifting to \btok{\_equilateral}, \btok{\_triangle}, and \btok{\_area} in the equilateral triangle related query.
    \item \textbf{Dimension 1046} (Quantitative \& Symbolic Content): This dimension specializes in numerical and quantitative concepts. In the tens digit query, it activates on specific numbers like \btok{0}, \btok{5} and \btok{6}. In the equilateral triangle query, its focus sharpens on raw numerical values and symbolic tokens, with top activations including \btok{2}, \btok{4}, and \btok{1}. This confirms that dimension 1046 serves as a dedicated processor for quantitative information.
    \item \textbf{Structural \& Functional Syntax Dimensions:} Consistent with the biology results, dimension 1645 continue to track the query structure (e.g., \btok{Answer}, \btok{\_questions}, while dimension 1068 maintains its role in attending to functional words.
\end{itemize}

\clearpage
\subsection{Additional Dataset-level Analysis on Other Domains}
\label{supp:A6}
Beyond the primary subjects discussed in \S\ref{sec:3}, we extend our dataset-level analysis to \textit{high school computer science} and \textit{high school chemistry}.
The following results shows that the semantic specialization of domain-critical dimensions is a universal phenomenon across varied scientific disciplines, not limited to biology or mathematics.

\begin{figure}[h]{
\vspace{-2mm}
\centering
\hspace{-0.5cm}
\includegraphics[width=1.05\columnwidth]{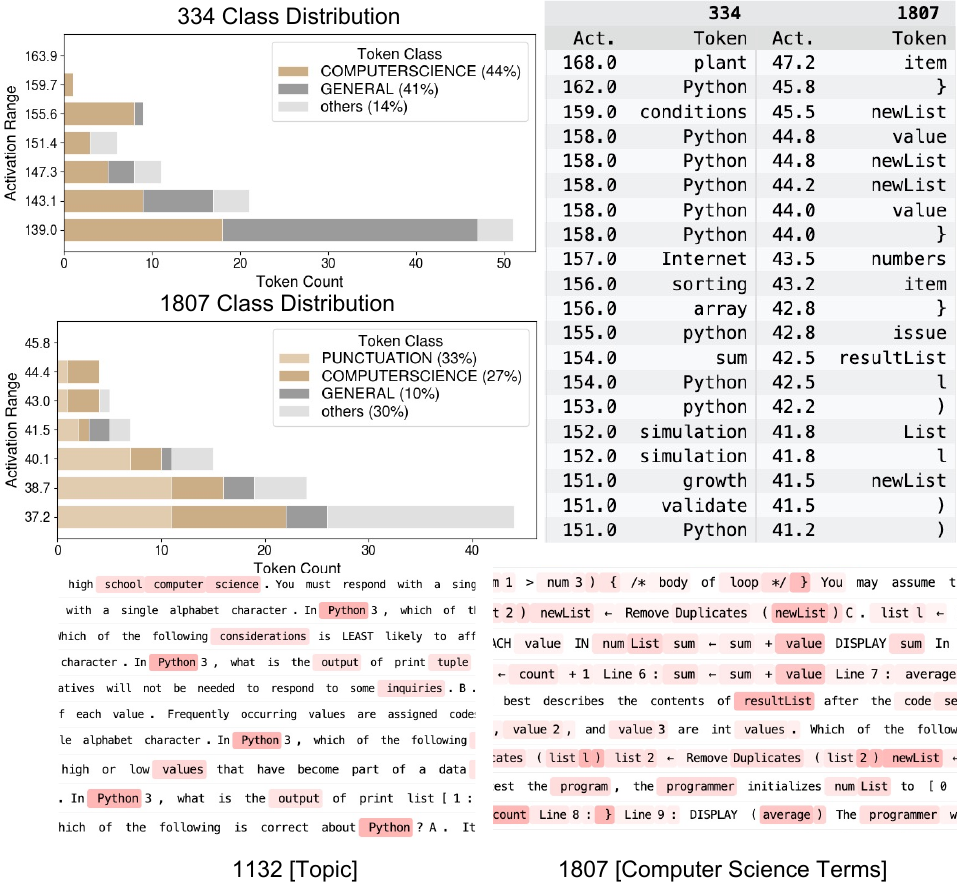}
\vspace{-7mm}
\caption{Analysis of \idcd in \textit{high school computer science}.
\textbf{(Top-left)} Token class distributions of dimension 334 (rank-1) and 1807 (rank-6). 
\textbf{(Top-right)} Highest activated tokens for dimension 334 and 1807. 
\textbf{(Bottom)} Activation heatmaps highlighting topic-level markers and specific computer science terms/syntax.}
\label{fig:figure_15}
\vspace{-3mm}
}
\end{figure}
\paragraph{High school computer science.}
Figure \ref{fig:figure_15} illustrates the semantic roles of dimension 334 and dimension 1807.
\begin{itemize}[leftmargin=3.5mm, topsep=3pt, itemsep=3pt, parsep=0pt]
\item[$\circ$] \textbf{\textit{Dimension 334}} acts as a high-level subject activator, consistent with its role in other STEM subjects mentioned in \S\ref{sec:3}.
44\% of its highly activated tokens belong to the \btok{COMPUTERSCIENCE} class, including terms like \btok{Python}, \btok{\_sorting}, and \btok{\_array}. 
The activation heatmap shows it consistently triggers on explicit domain indicators such as \btok{computer science} and \btok{Python}.
\item[$\circ$] \textbf{\textit{Dimension 1807}} specializes in the structural and lexical components of programming, ranking 6 in high school computer science.
27\% of activated tokens are CS-specific, while 33\% are \btok{PUNCTUATION}, reflecting the symbol-heavy nature of code. 
It responds strongly to variable-like tokens (\btok{\_newList}, \btok{\_resultList}) and structural symbols (\btok{\}}, \btok{)}, \btok{[}), effectively tracking the logic and syntax of computer science problems.
\end{itemize}

\newpage
\begin{figure}[h]
\hspace{-1mm}
\centering
\includegraphics[width=1.05\columnwidth]{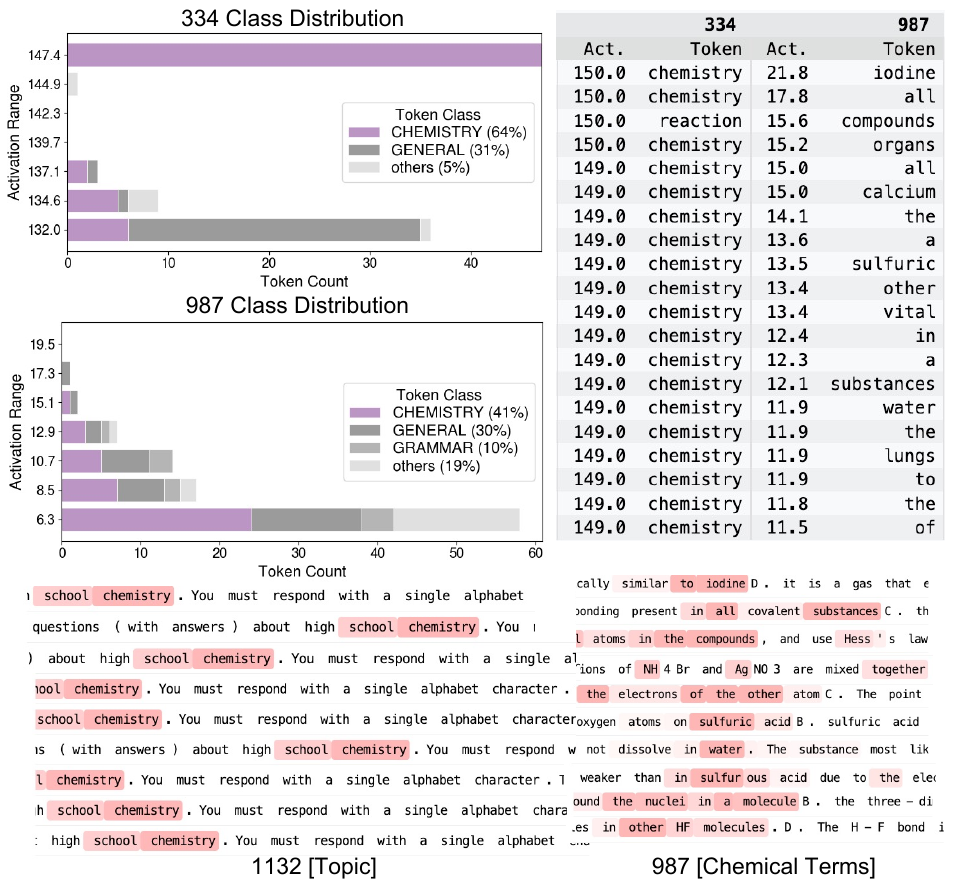}
\caption{Analysis of \idcd in \textit{high school chemistry}.
\textbf{(Top-left)} Token class distributions of dimension 334 (rank-1) and 987 (rank-8). 
\textbf{(Top-right)} Highest activated tokens for dimension 334 and 987.
\textbf{(Bottom)} Activation heatmaps highlighting domain-level markers and specific chemical terminology.}
\label{fig:figure_16}
\vspace{-3mm}
\end{figure}
\paragraph{High school chemistry.} Additionally, we analyze the \textit{high school chemistry} dataset to confirm the functional roles of its domain-critical dimensions. 
Figure \ref{fig:figure_16} illustrates the behavior of dimension 334 and dimension 987.
\begin{itemize}[leftmargin=3.5mm, topsep=3pt, itemsep=3pt, parsep=0pt]
\item[$\circ$] \textbf{\textit{Dimension 334}} serves as the primary critical dimension, as it acts as a robust indicator for the chemistry domain. 
64\% of its highly activated tokens are categorized as \btok{CHEMISTRY}, with the token \btok{\_chemistry} itself appearing most frequently at the highest activation levels. 
The heatmap reveals that it consistently triggers on explicit subject markers like \btok{school chemistry} within the prompt.
\item[$\circ$] \textbf{\textit{Dimension 987}} ranked 8th in high school chemistry, specializing in specific chemical entities and substances. 
41\% of its activations correspond to the \btok{CHEMISTRY} class, targeting concrete terms such as \btok{\_sulfuric}, \btok{\_iodine}, \btok{\_calcium}, and \btok{\_compounds}. 
The activation heatmap demonstrates that it precisely tracks fine-grained chemical concepts including \btok{NH}, \btok{Ag}, and \btok{sulfuric acid}, facilitating the processing of domain-specific scientific content.
\end{itemize}

\clearpage
\subsection{Extension to Larger Scale Model: Gemma-2-9B-IT}
\label{supp:A7}
To evaluate whether functional criticality scales with model size, we apply our identification framework to Gemma-2-9b-it. Our analysis reveals that larger models similarly shows a highly interpretable domain-critical dimensions, suggesting that the concentration of domain-specific information is a persistent property of scaled-up architectures.
\begin{figure}[h]
\centering
\includegraphics[width=1.0\columnwidth]{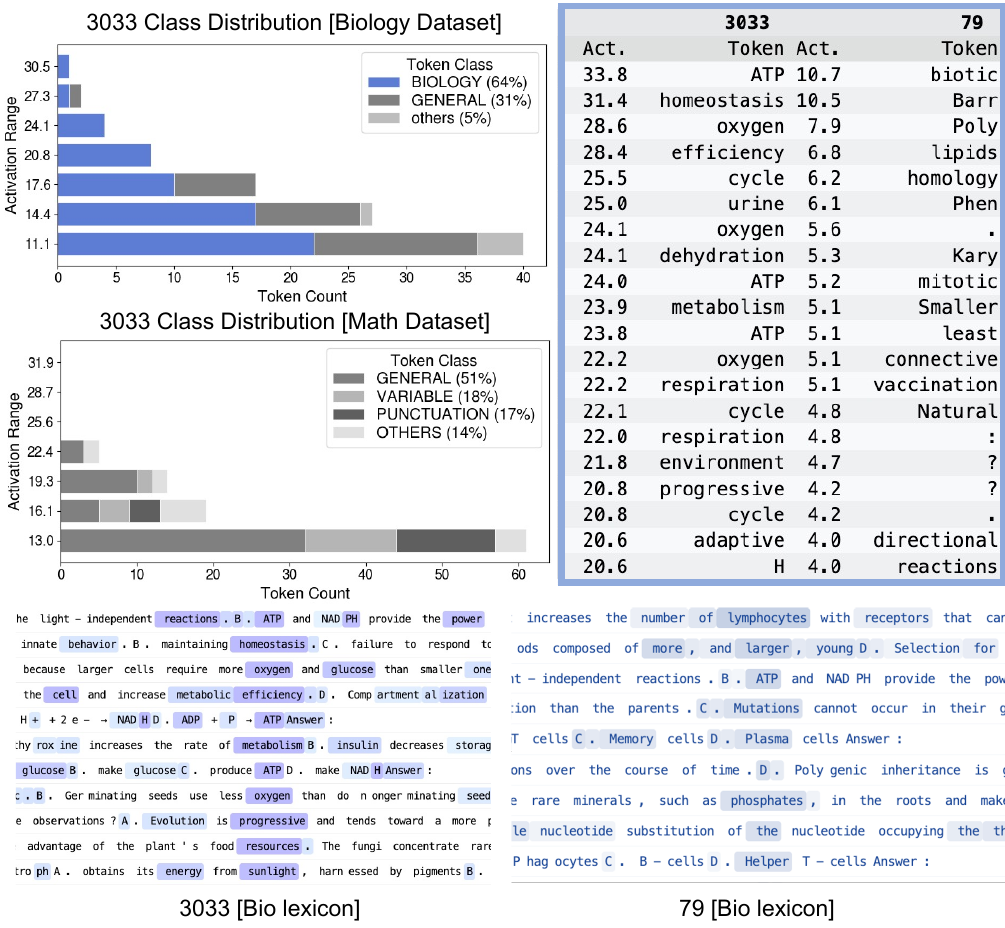}
\vspace{-7mm}{
\caption{{High school biology $\mathcal{I}_{\text{dcd}}$ in Gemma-2-9b-it.}
\textbf{(Top-left)} Token class distribution of dimension 3033 across biology and math datasets.
\textbf{(Top-right)} Highly activated tokens for dimension 79 (rank-7) and 3033 (rank-8).
\textbf{(Bottom)} Activation heatmaps of dimension 3033 and 79 on biology dataset prompts.}
\label{fig:figure_17}}
\vspace{-3mm}
\end{figure}
\paragraph{High school Biology.}
Figure \ref{fig:figure_17} illustrates biology prominent dimensions, dimension 3033 and dimension 79.
\begin{itemize}[leftmargin=3.5mm, topsep=3pt, itemsep=3pt, parsep=0pt]
\item[$\circ$] \textbf{\textit{Dimension 3033}} exhibits strong domain selectivity similar to the critical dimensions observed in the 2B model. 
As shown in Figure \ref{fig:figure_17}, when ananlyzed on the biology dataset, 64\% of activated tokens are \btok{BIOLOGY} entities, including core concepts such as \btok{\_ATP}, and \btok{\_homeostasis}. While in math dataset, it shifts to \btok{GENERAL} (51\%) and \btok{VARIABLE} (18\%) tokens. 
Heatmaps confirm activation on terms like \btok{\_metabolism} and \btok{\_NADPH}.
\item[$\circ$] \textbf{\textit{dimension 79}} acts as a specialized detector for biological terminology (e.g., \btok{\_lipids}, \btok{\_vaccination}). 
It tracks fine-grained entities such as \btok{\_lymphocytes} and \btok{\_mutations}, supporting the hypothesis that larger models also utilize interpretable dimensions.
\end{itemize}

\newpage
\paragraph{High school Mathematics.}
\begin{figure}[h]
\centering
\hspace{0cm}
\includegraphics[width=1.0\columnwidth]{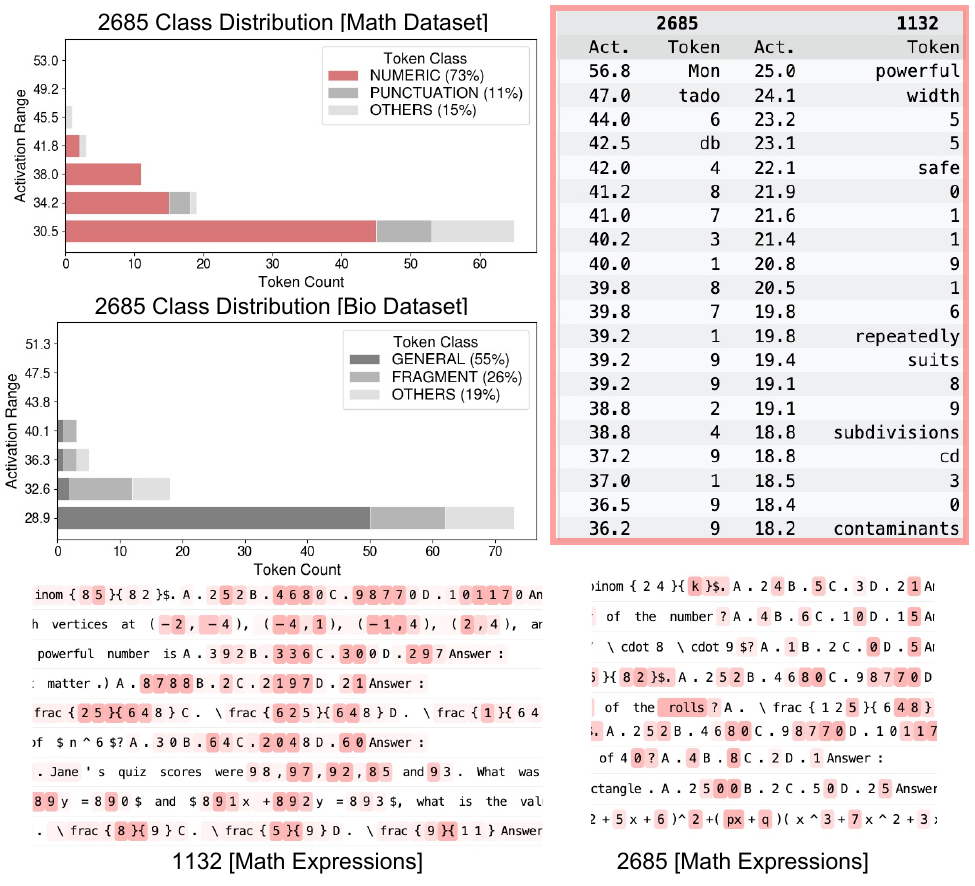}
\vspace{-7mm}{
\caption{{High school mathematics $\mathcal{I}_{\text{dcd}}$ in Gemma-2-9b-it.}
\textbf{(Top-left)} Token class distribution of dimension 2685 across math and biology datasets.
\textbf{(Top-right)} Highly activated tokens for dimension 2685 (rank-10) and 1132 (rank-9).
\textbf{(Bottom)} Activation heatmaps of dimension 1132 and 2685 on mathematical prompts.}
\label{fig:figure_18}}
\end{figure}
We further analyze the domain-critical dimensions identified from the \textit{high school mathematics} dataset in the Gemma-2-9b-it model. Figure \ref{fig:figure_18} illustrates the semantic roles of two specialized dimensions, dimension 2685 and dimension 1132.
\begin{itemize}[leftmargin=3.5mm, topsep=3pt, itemsep=3pt, parsep=0pt]
\item[$\circ$] \textbf{\textit{Dimension 2685}} demonstrates a clear specialization in numerical content.
As shown in the class distribution plots, when evaluated on the mathematics dataset, 73\% of the highly activated tokens are classified as \btok{NUMERIC}, with top tokens consisting of various digits such as \btok{6}, \btok{4}, and \btok{8}.
In contrast, on the biology dataset, this domain-specific preference disappears, with \btok{GENERAL} tokens accounting for the majority (55\%).
The activation heatmap confirms that dimension 2685 selectively targets numerical values and digits within mathematical expressions.
\item[$\circ$] \textbf{\textit{Dimension 1132}} focuses on the components of mathematical formulas and quantitative descriptions. 
The top activated tokens include specific digits, alongside quantitative terms such as \btok{width}, and \btok{subdivisions}. 
In the corresponding heatmap, dimension 1132 responds strongly to numbers embedded within LaTeX-style expressions, showing that it captures the quantitative elements necessary for solving mathematical problems.
\end{itemize}

\clearpage
\subsection{Extension to Other Family Model: Llama-3.1-8B-IT}
\label{supp:A8}
\begin{figure*}[t]
    \centering
    \makebox[\textwidth][c]{\includegraphics[width=1.03\textwidth]{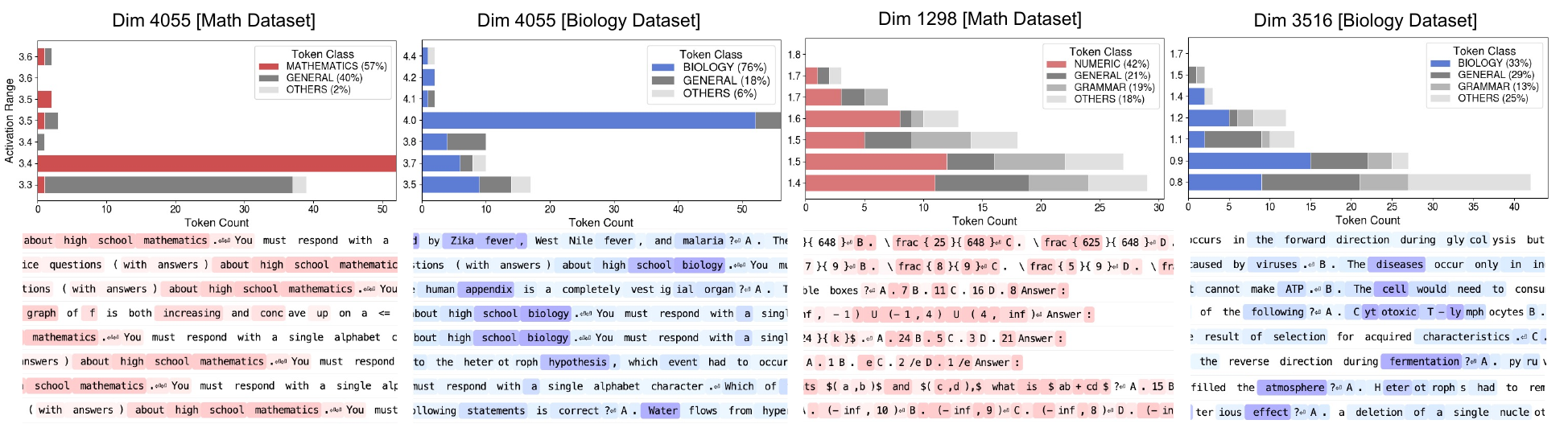}}
    \vspace{-7mm}
    \caption{\idcd activation pattern in Llama-3.1-8b-it. 
    Token analysis on domain-critical dimensions of \textit{high school mathematics} and \textit{high school biology}.
    \textbf{(a)} indicates token class distribution and activation of dimension 4055 (rank-5 in mathematics), 
    \textbf{(b)} dimension 4055 (rank-5 in biology), 
    \textbf{(c)} dimension 1298 (rank-14 in mathematics), 
    and \textbf{(d)} dimension 3516 (rank-13 in biology).
    }
    \label{fig:figure_19}
\end{figure*}
To demonstrate that the interpretable features identified in our main analysis(\S\ref{sec:3}) are not specific to a single model architecture, we extend our experiments to Llama-3.1-8b-it. 
This analysis aim to verify whether the functional roles of specific dimensions are preserved across different model families.

\paragraph{Domain indicator dimension.} We observe dimension 4055 as a direct functional analogue to dimension 334 described in the main text. 
It ranks 5th in importance for both the mathematics and biology datasets, confirming its role as a global {topic indicator}. 
As visualized in Figure \ref{fig:figure_19}, this dimension strongly activates on high-level domain keywords (e.g., \btok{\_mathematics}, \btok{\_biology}). 
This consistency suggests that the mechanism of dedicating specialized dimensions to track the global context or topic is a generalizable property of LLMs, independent of the specific training run.

\paragraph{Domain specialized dimension.}
Beyond domain indicator dimension, Llama-3.1-8b-it exhibits specialized dimensions dedicated to fine-grained concepts, mirroring the behavior observed in our primary results in \S\ref{sec:3}. 
As illustrated in Figure \ref{fig:figure_19}c, dimension 1298 (rank 14 in mathematics) focuses on the syntactic structure of mathematics. 
This functional role is quantitatively supported by its activation distribution, which is dominated by \btok{NUMERIC}(42\%) and \btok{GRAMMAR}(19\%) tokens such as LaTeX operators (e.g., \btok{\textbackslash frac}), effectively separating mathematical notation from natural language context. 
Similarly, Figure \ref{fig:figure_19}d demonstrates the role of dimension 3516 (rank 13 in biology) as a biology-specific entity detector.
With \btok{BIOLOGY} tokens (33\%) constituting the largest share, this dimension selectively highlights biology entities like \btok{\_cell} and \btok{\_fermentation}, showcasing the model's capacity to isolate and process granular biological vocabulary.

\clearpage %
\subsection{Additional CDS Results for \S\ref{sec:4.1}}
\label{supp:A9}

In \S\ref{sec:4.1}, we reported the performance of Critical Dimension Steering (CDS) using the optimal number of domain-critical dimensions $k$ determined for each subject.
Given that each subject possesses distinct domain nuances and utilizes different steering vectors constructed from subject-specific identification sets, optimizing $k$ for each domain is a natural approach to maximize control efficacy.
However, to decouple the intrinsic effectiveness of our selection method from the benefits of hyperparameter optimization, we present an analysis of CDS performance with a fixed $k$ across all 57 MMLU subjects.
We compare this against the Whole Dimension Steering (WDS) baseline, which utilizes all dimensions ($D=2304$) and thus remains constant across these comparisons.
Table \ref{tab:table_7} summarizes the average accuracy improvement for each setting.

\paragraph{Performance scaling with $k$.}
As shown in Table \ref{tab:table_7}, CDS has a performance trend correlated with the dimension budget $k$.
At the sparsest setting ($k$=$100$), shown in Figure \ref{fig:figure_20}e, CDS achieves an average improvement of 1.47\%, which is slightly below the WDS baseline of 1.51\%.
However, despite the lower average, CDS ($k$=$100$) outperforms WDS in 20 subjects compared to 16 wins for WDS, suggesting that even a small subset of critical dimensions allows for more frequent success in precision-heavy tasks.
As $k$ increases, CDS consistently surpasses the WDS baseline.
In Figure \ref{fig:figure_20}d ($k$=$200$), the average improvement rises to 1.61\%, crossing the WDS threshold.
The performance gap widens further at $k$=$500$ (Figure \ref{fig:figure_20}a), where CDS achieves a 2.07\% average improvement.
This indicates that while the most \textit{critical} features are sparse, a broader support set, up to $k$=$500$, contributes to maximizing domain adaptation performance.
Nevertheless, comparing the fixed $k$=$500$ result (2.07\%) with the result in \S\ref{sec:4.1} (3.09\%) highlights that subject-specific sparsity tuning remains essential for optimal control.

\paragraph{Mitigation of Negative Transfer.}
A key advantage of CDS over WDS is the reduction of negative transfer, particularly in reasoning-heavy tasks.
As visualized across Figure \ref{fig:figure_20}a--e, WDS consistently degrades performance in \textit{formal logic} (-8\%) and \textit{high school mathematics} (-2\%).
CDS effectively mitigates this interference.
In Figure \ref{fig:figure_20}a ($k$=$500$), the degradation in \textit{formal logic} is significantly reduced from -8\% (WDS) to -2\% (CDS).
Furthermore, at lower sparsity levels such as $k$=$100$ (Figure \ref{fig:figure_20}e) and $k$=$200$ (Figure \ref{fig:figure_20}d), CDS recovers performance in \textit{high school mathematics} to positive gains (+2\%), whereas WDS remains negative.
This confirms that by masking out non-critical dimensions, CDS preserves the model's general reasoning capabilities better than whole-dimension steering, regardless of the specific $k$ value.

\begin{table}[t]
\centering
\small
\caption{Average accuracy improvement over standard baseline across 57 MMLU subjects. CDS is evaluated at fixed dimension numbers $k$ while WDS uses all dimensions. The WDS average corresponds to 1.51\%.}
\vspace{-2mm}

\resizebox{\columnwidth}{!}{%
\begin{tabular}{l c c c c c}
\toprule
\textbf{Dimension number $k$} & $100$ & $200$ & $300$ & $400$ & $500$ \\
\midrule
\textbf{CDS Average (\%)} & 1.47 & 1.61 & 1.86 & 1.86 & 2.07 \\
\textbf{Subjects (CDS/WDS/Tie)} 
& 20/16/21
& 23/17/17
& 19/12/26
& 18/14/25
& 21/11/25 \\
\bottomrule
\end{tabular}}
\label{tab:table_7}
\end{table}

\subsection{Extension to More Models}
\label{supp:A10}
{We have conducted additional experiments on other model families, OLMo-2-1B-Instruct \citep{olmo20242}, Llama-3.1-8B-IT \citep{grattafiori2024llama}, Qwen-3-8B \citep{yang2025qwen3}, and Gemma-2-9B-IT \citep{team2024gemma}, presented in Table \ref{tab:olmo}, \ref{tab:llama}, \ref{tab:qwen}, \ref{tab:gemma}. 
Our results show that CDS consistently outperforms the WDS baseline across the majority of MMLU subjects (e.g., 37 vs. 10 wins for OLMo-2-1B-Instruct, and 38 vs. 7 wins for Qwen-3-8B). 
This consistent effectiveness across different model families demonstrates the robustness and generalizability of our findings.}
\begin{table}[h]
\centering
\small
\caption{Average accuracy improvement over baselines on various models.}
\vspace{-2mm}
\resizebox{\columnwidth}{!}{
\begin{tabular}{lccc}
\toprule
\textbf{Model} & \textbf{Base} & \textbf{WDS} & \textbf{CDS} \\
\midrule
\text{OLMo-2-1B-Instruct \hspace{2cm}} & 38.77 & +2.70 & +4.63 \\
Llama-3.1-8B-IT & 64.60 & +1.56 & +3.48 \\
Qwen-3-8B & 73.30 & +1.47 & \textbf{+2.42} \\
Gemma-2-9B-IT & 70.56 & -0.14 & +1.40 \\
\bottomrule
\end{tabular}}
\label{tab:table_8}
\end{table}

\begin{figure*}[p]
\centering
\begin{subfigure}[b]{\textwidth}
\centering
\caption{CDS ($k=500$) vs. WDS}
\includegraphics[width=0.9\textwidth]{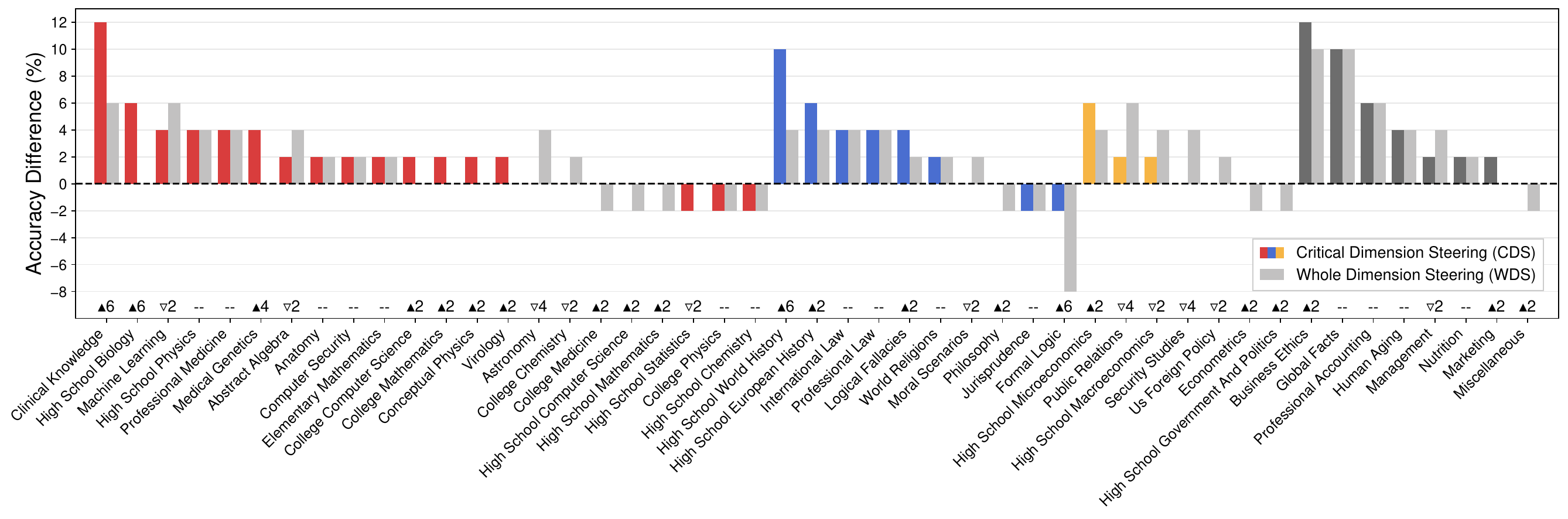}
\end{subfigure}%
\vspace{-5mm}
\begin{subfigure}[b]{\textwidth}
\centering
\caption{CDS ($k=400$) vs. WDS}
\includegraphics[width=0.9\textwidth]{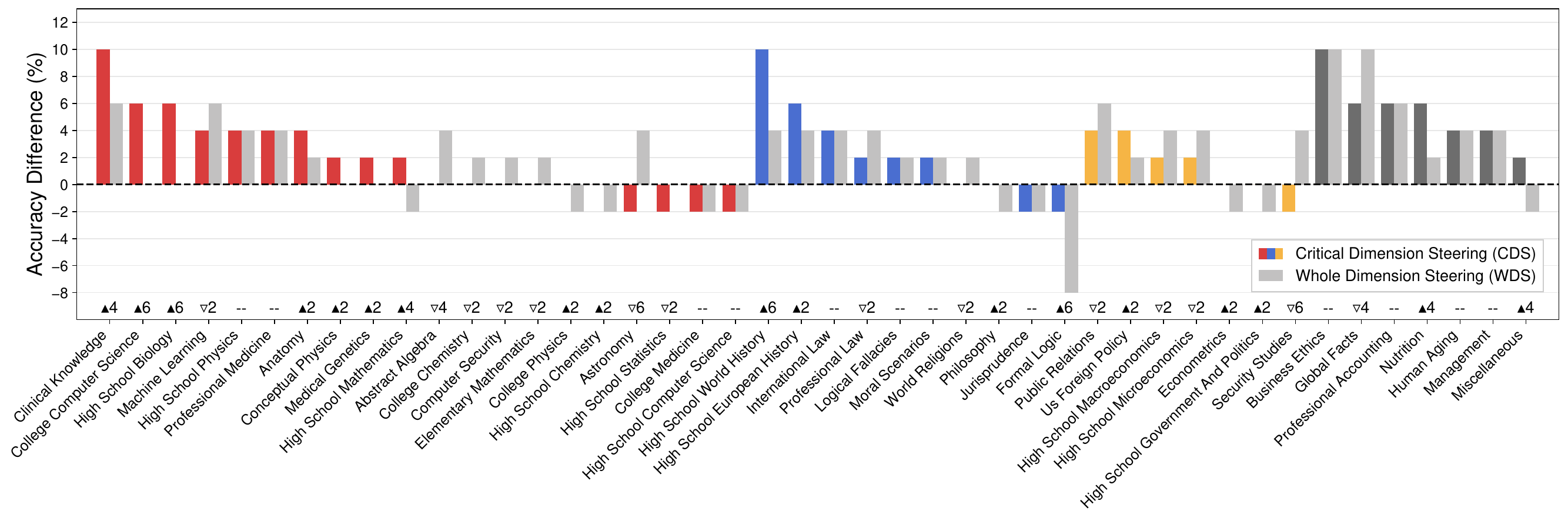}
\end{subfigure}%
\vspace{-5mm}
\begin{subfigure}[b]{\textwidth}
\centering
\caption{CDS ($k=300$) vs. WDS}
\includegraphics[width=0.9\textwidth]{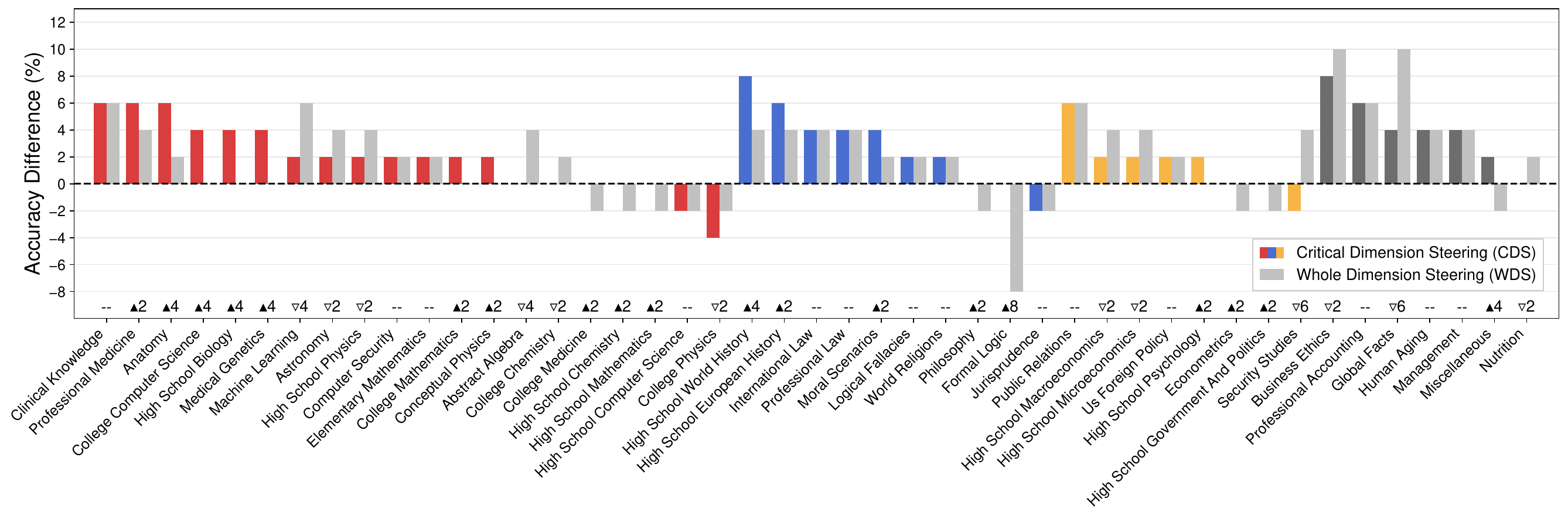}
\end{subfigure}%
\vspace{-5mm}
\begin{subfigure}[b]{\textwidth}
\centering
\caption{CDS ($k=200$) vs. WDS}
\includegraphics[width=0.9\textwidth]{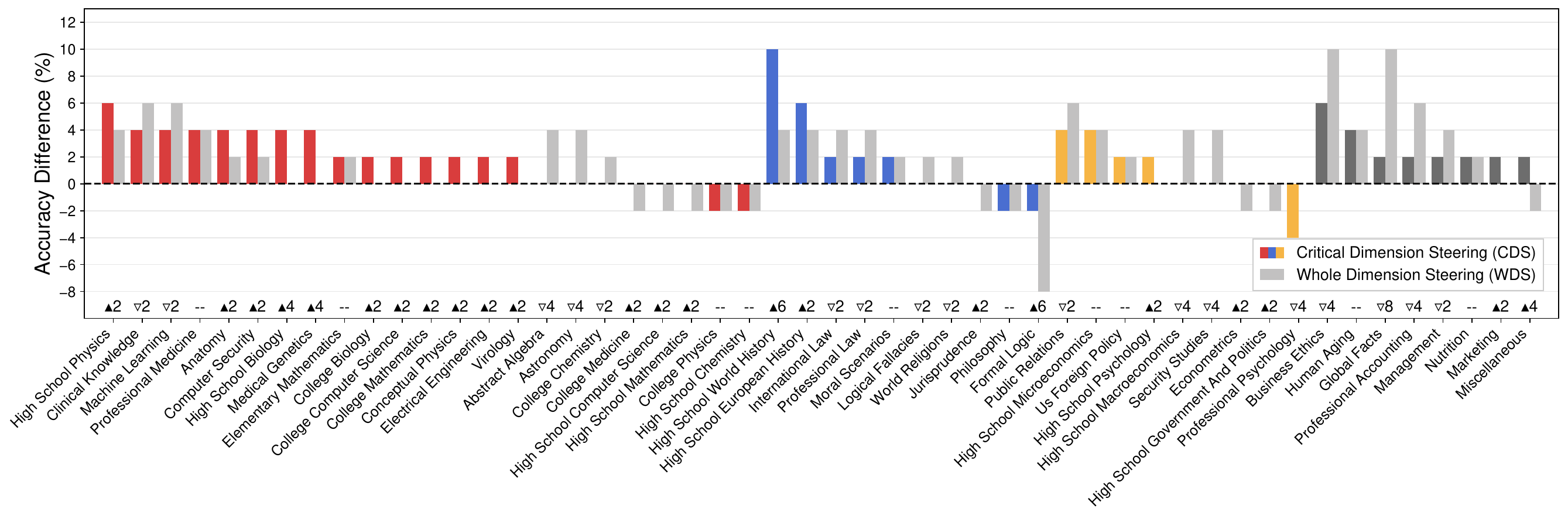}
\end{subfigure}%
\vspace{-5mm}
\begin{subfigure}[b]{\textwidth}
\centering
\caption{CDS ($k=100$) vs. WDS}
\includegraphics[width=0.9\textwidth]{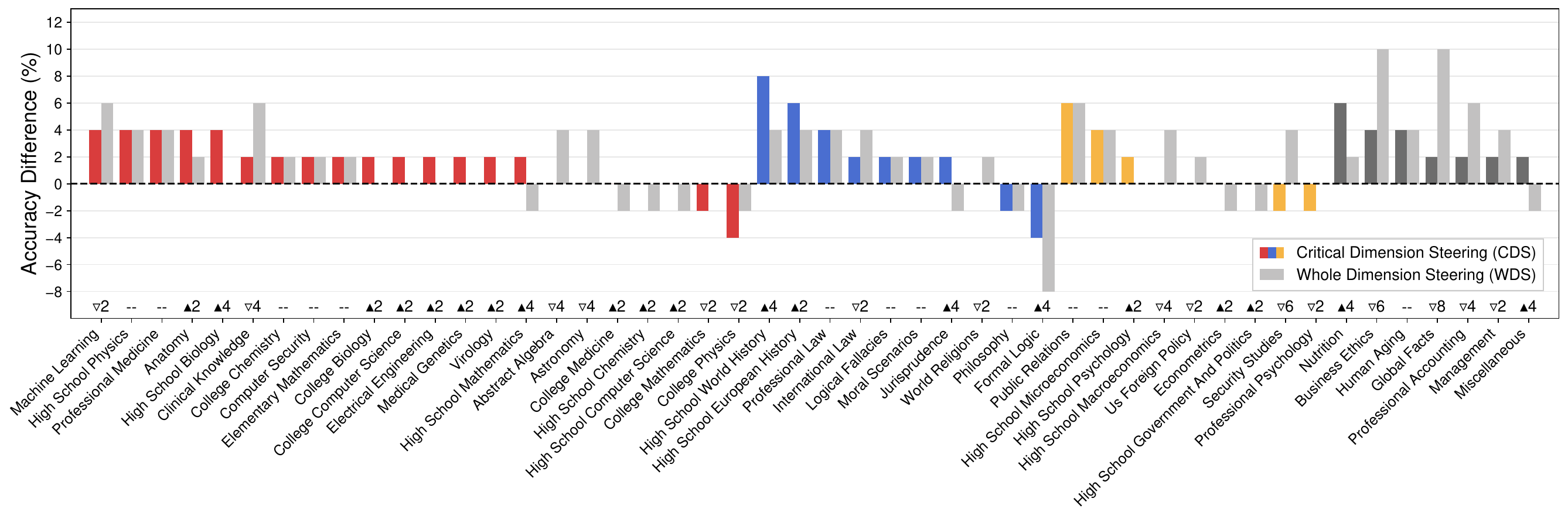}
\end{subfigure}%
\vspace{-4mm}
\caption{
{Subject-wise accuracy difference for fixed $k$ CDS settings.}
}
\label{fig:figure_20}
\end{figure*}

\begin{table*}[t]
\centering
\small
\caption{Performance of OLMo-2-1B-Instruct across all 57 MMLU subjects.}
\vspace{-0.1in}
\resizebox{\textwidth}{!}{
    \begin{tabular}{lccc lccc lccc}
        \toprule
        \textbf{Subject} & \textbf{Base} & \textbf{WDS} & \textbf{CDS} & 
        \textbf{Subject} & \textbf{Base} & \textbf{WDS} & \textbf{CDS} & 
        \textbf{Subject} & \textbf{Base} & \textbf{WDS} & \textbf{CDS} \\ 
        \midrule
        Abstract Algebra & 32.0 & -2.0 & +2.0 & High School Chemistry & 14.0 & +12.0 & +14.0 & Management & 36.0 & +4.0 & +2.0 \\
        Anatomy & 40.0 & +10.0 & +6.0 & High School Computer Science & 38.0 & +4.0 & +12.0 & Marketing & 64.0 & 0.0 & +4.0 \\
        Astronomy & 32.0 & +8.0 & +8.0 & High School European History & 60.0 & 0.0 & +2.0 & Medical Genetics & 48.0 & -2.0 & +6.0 \\
        Business Ethics & 40.0 & +2.0 & +6.0 & High School Geography & 44.0 & -2.0 & 0.0 & Miscellaneous & 46.0 & +8.0 & +8.0 \\
        Clinical Knowledge & 36.0 & +10.0 & +14.0 & High School Gov. and Politics & 42.0 & +4.0 & 0.0 & Moral Disputes & 36.0 & +6.0 & +6.0 \\
        College Biology & 42.0 & +4.0 & +6.0 & High School Macroeconomics & 30.0 & +2.0 & 0.0 & Moral Scenarios & 30.0 & 0.0 & 0.0 \\
        College Chemistry & 34.0 & -4.0 & -2.0 & High School Mathematics & 26.0 & +2.0 & +8.0 & Nutrition & 48.0 & +2.0 & +4.0 \\
        College Computer Science & 30.0 & +2.0 & +2.0 & High School Microeconomics & 36.0 & +14.0 & +4.0 & Philosophy & 38.0 & 0.0 & +2.0 \\
        College Mathematics & 28.0 & +4.0 & +8.0 & High School Physics & 28.0 & +2.0 & +6.0 & Prehistory & 48.0 & +6.0 & +8.0 \\
        College Medicine & 30.0 & +6.0 & +10.0 & High School Psychology & 50.0 & +8.0 & +10.0 & Professional Accounting & 32.0 & 0.0 & +6.0 \\
        College Physics & 24.0 & 0.0 & +8.0 & High School Statistics & 34.0 & 0.0 & +2.0 & Professional Law & 30.0 & +4.0 & +6.0 \\
        Computer Security & 52.0 & +2.0 & 0.0 & High School US History & 48.0 & +4.0 & +10.0 & Professional Medicine & 28.0 & 0.0 & +6.0 \\
        Conceptual Physics & 30.0 & +4.0 & +6.0 & High School World History & 48.0 & +2.0 & +2.0 & Professional Psychology & 32.0 & +2.0 & +4.0 \\
        Econometrics & 28.0 & +6.0 & +2.0 & Human Aging & 54.0 & -2.0 & +4.0 & Public Relations & 52.0 & +8.0 & +10.0 \\
        Electrical Engineering & 58.0 & -6.0 & 0.0 & Human Sexuality & 46.0 & +6.0 & +6.0 & Security Studies & 52.0 & 0.0 & +2.0 \\
        Elementary Mathematics & 36.0 & -2.0 & -4.0 & International Law & 50.0 & -2.0 & 0.0 & Sociology & 60.0 & 0.0 & +4.0 \\
        Formal Logic & 26.0 & 0.0 & 0.0 & Jurisprudence & 48.0 & +4.0 & +2.0 & US Foreign Policy & 56.0 & +2.0 & +6.0 \\
        Global Facts & 20.0 & +2.0 & +2.0 & Logical Fallacies & 38.0 & 0.0 & +4.0 & Virology & 38.0 & +2.0 & +6.0 \\
        High School Biology & 34.0 & +14.0 & +12.0 & Machine Learning & 40.0 & 0.0 & +2.0 & World Religions & 46.0 & +4.0 & +4.0 \\
        \bottomrule
    \end{tabular}
}
\label{tab:olmo}
\vspace{-0.1in} 
\end{table*}

\begin{table*}[t]
\centering
\small
\caption{Performance of Llama-3.1-8B-IT across all 57 MMLU subjects.}
\vspace{-0.1in}
\resizebox{\textwidth}{!}{
    \begin{tabular}{lccc lccc lccc}
        \toprule
        \textbf{Subject} & \textbf{Base} & \textbf{WDS} & \textbf{CDS} & 
        \textbf{Subject} & \textbf{Base} & \textbf{WDS} & \textbf{CDS} & 
        \textbf{Subject} & \textbf{Base} & \textbf{WDS} & \textbf{CDS} \\ 
        \midrule
        Abstract Algebra & 38.0 & -4.0 & 0.0 & High School Chemistry & 58.0 & +4.0 & +4.0 & Management & 68.0 & 0.0 & 0.0 \\
        Anatomy & 60.0 & 0.0 & +10.0 & High School Computer Science & 68.0 & -2.0 & -2.0 & Marketing & 88.0 & +4.0 & +4.0 \\
        Astronomy & 60.0 & +2.0 & +2.0 & High School European History & 76.0 & -4.0 & +2.0 & Medical Genetics & 78.0 & +2.0 & +2.0 \\
        Business Ethics & 60.0 & -2.0 & 0.0 & High School Geography & 80.0 & +6.0 & +8.0 & Miscellaneous & 72.0 & 0.0 & 0.0 \\
        Clinical Knowledge & 54.0 & +12.0 & +14.0 & High School Gov. and Politics & 78.0 & 0.0 & 0.0 & Moral Disputes & 78.0 & 0.0 & +2.0 \\
        College Biology & 74.0 & -2.0 & -2.0 & High School Macroeconomics & 72.0 & -2.0 & 0.0 & Moral Scenarios & 54.0 & 0.0 & +2.0 \\
        College Chemistry & 48.0 & 0.0 & 0.0 & High School Mathematics & 42.0 & +6.0 & +4.0 & Nutrition & 74.0 & -4.0 & -4.0 \\
        College Computer Science & 64.0 & -4.0 & 0.0 & High School Microeconomics & 78.0 & -2.0 & -2.0 & Philosophy & 84.0 & -4.0 & 0.0 \\
        College Mathematics & 26.0 & +10.0 & +8.0 & High School Physics & 40.0 & +8.0 & +6.0 & Prehistory & 78.0 & 0.0 & +4.0 \\
        College Medicine & 70.0 & -6.0 & 0.0 & High School Psychology & 90.0 & +2.0 & +2.0 & Professional Accounting & 56.0 & +2.0 & +2.0 \\
        College Physics & 38.0 & 0.0 & 0.0 & High School Statistics & 56.0 & +4.0 & +2.0 & Professional Law & 54.0 & -2.0 & 0.0 \\
        Computer Security & 74.0 & -2.0 & +2.0 & High School US History & 82.0 & +4.0 & +2.0 & Professional Medicine & 60.0 & 0.0 & 0.0 \\
        Conceptual Physics & 52.0 & +2.0 & +6.0 & High School World History & 76.0 & -4.0 & 0.0 & Professional Psychology & 54.0 & 0.0 & 0.0 \\
        Econometrics & 58.0 & 0.0 & +4.0 & Human Aging & 58.0 & +6.0 & +8.0 & Public Relations & 66.0 & -4.0 & +2.0 \\
        Electrical Engineering & 60.0 & +4.0 & +10.0 & Human Sexuality & 82.0 & +2.0 & +2.0 & Security Studies & 72.0 & +10.0 & +10.0 \\
        Elementary Mathematics & 38.0 & +2.0 & +4.0 & International Law & 70.0 & +4.0 & +2.0 & Sociology & 84.0 & -2.0 & 0.0 \\
        Formal Logic & 42.0 & 0.0 & +6.0 & Jurisprudence & 72.0 & 0.0 & +6.0 & US Foreign Policy & 84.0 & +2.0 & +2.0 \\
        Global Facts & 46.0 & +4.0 & +10.0 & Logical Fallacies & 70.0 & +10.0 & +10.0 & Virology & 54.0 & 0.0 & +2.0 \\
        High School Biology & 74.0 & +2.0 & +4.0 & Machine Learning & 38.0 & +4.0 & +8.0 & World Religions & 70.0 & +10.0 & +6.0 \\
        \bottomrule
    \end{tabular}
}
\label{tab:llama}
\vspace{-0.1in} 
\end{table*}

\begin{table*}[t]
\centering
\small
\caption{Performance of Qwen-3-8B across all 57 MMLU subjects.}
\vspace{-0.1in}
\resizebox{\textwidth}{!}{
    \begin{tabular}{lccc lccc lccc}
        \toprule
        \textbf{Subject} & \textbf{Base} & \textbf{WDS} & \textbf{CDS} & 
        \textbf{Subject} & \textbf{Base} & \textbf{WDS} & \textbf{CDS} & 
        \textbf{Subject} & \textbf{Base} & \textbf{WDS} & \textbf{CDS} \\ 
        \midrule
        Abstract Algebra & 50.0 & 0.0 & +8.0 & High School Chemistry & 66.0 & +2.0 & +4.0 & Management & 86.0 & 0.0 & +2.0 \\
        Anatomy & 72.0 & 0.0 & +2.0 & High School Computer Science & 86.0 & +2.0 & 0.0 & Marketing & 90.0 & +2.0 & +2.0 \\
        Astronomy & 78.0 & -2.0 & 0.0 & High School European History & 86.0 & 0.0 & +2.0 & Medical Genetics & 76.0 & -2.0 & 0.0 \\
        Business Ethics & 54.0 & +4.0 & +6.0 & High School Geography & 90.0 & 0.0 & 0.0 & Miscellaneous & 80.0 & -2.0 & 0.0 \\
        Clinical Knowledge & 78.0 & 0.0 & +2.0 & High School Gov. and Politics & 94.0 & 0.0 & 0.0 & Moral Disputes & 68.0 & +2.0 & +4.0 \\
        College Biology & 86.0 & 0.0 & 0.0 & High School Macroeconomics & 84.0 & 0.0 & -2.0 & Moral Scenarios & 40.0 & 0.0 & +2.0 \\
        College Chemistry & 60.0 & +4.0 & +4.0 & High School Mathematics & 50.0 & +6.0 & +10.0 & Nutrition & 72.0 & 0.0 & 0.0 \\
        College Computer Science & 72.0 & -2.0 & 0.0 & High School Microeconomics & 94.0 & 0.0 & 0.0 & Philosophy & 86.0 & 0.0 & 0.0 \\
        College Mathematics & 54.0 & +2.0 & +4.0 & High School Physics & 68.0 & -2.0 & 0.0 & Prehistory & 80.0 & 0.0 & +2.0 \\
        College Medicine & 66.0 & -4.0 & -2.0 & High School Psychology & 82.0 & +2.0 & +4.0 & Professional Accounting & 52.0 & +2.0 & +10.0 \\
        College Physics & 56.0 & +2.0 & 0.0 & High School Statistics & 84.0 & 0.0 & +2.0 & Professional Law & 56.0 & -2.0 & 0.0 \\
        Computer Security & 80.0 & +4.0 & +2.0 & High School US History & 90.0 & 0.0 & 0.0 & Professional Medicine & 82.0 & +4.0 & +6.0 \\
        Conceptual Physics & 74.0 & +2.0 & +4.0 & High School World History & 86.0 & +2.0 & +2.0 & Professional Psychology & 72.0 & 0.0 & +2.0 \\
        Econometrics & 76.0 & +4.0 & +6.0 & Human Aging & 60.0 & +4.0 & +6.0 & Public Relations & 66.0 & -2.0 & 0.0 \\
        Electrical Engineering & 72.0 & +2.0 & +4.0 & Human Sexuality & 84.0 & 0.0 & +2.0 & Security Studies & 82.0 & 0.0 & +2.0 \\
        Elementary Mathematics & 70.0 & +4.0 & +2.0 & International Law & 80.0 & -2.0 & 0.0 & Sociology & 84.0 & 0.0 & +2.0 \\
        Formal Logic & 68.0 & 0.0 & +2.0 & Jurisprudence & 82.0 & -2.0 & +2.0 & US Foreign Policy & 84.0 & +4.0 & +2.0 \\
        Global Facts & 46.0 & +6.0 & +6.0 & Logical Fallacies & 80.0 & +2.0 & +2.0 & Virology & 52.0 & 0.0 & +2.0 \\
        High School Biology & 90.0 & +4.0 & +2.0 & Machine Learning & 52.0 & 0.0 & +2.0 & World Religions & 70.0 & +4.0 & +8.0 \\
        \bottomrule
    \end{tabular}
}
\label{tab:qwen}
\vspace{-0.1in} 
\end{table*}

\begin{table*}[t]
\centering
\small
\caption{Performance of Gemma-2-9B-IT across all 57 MMLU subjects.}
\vspace{-0.1in}
\resizebox{\textwidth}{!}{
    \begin{tabular}{lccc lccc lccc}
        \toprule
        \textbf{Subject} & \textbf{Base} & \textbf{WDS} & \textbf{CDS} & 
        \textbf{Subject} & \textbf{Base} & \textbf{WDS} & \textbf{CDS} & 
        \textbf{Subject} & \textbf{Base} & \textbf{WDS} & \textbf{CDS} \\ 
        \midrule
        Abstract Algebra & 34.0 & +8.0 & 0.0 & High School Chemistry & 62.0 & +2.0 & +4.0 & Management & 84.0 & 0.0 & +2.0 \\
        Anatomy & 70.0 & -10.0 & -2.0 & High School Computer Science & 72.0 & -2.0 & 0.0 & Marketing & 90.0 & +2.0 & +2.0 \\
        Astronomy & 72.0 & 0.0 & +4.0 & High School European History & 84.0 & -4.0 & -4.0 & Medical Genetics & 82.0 & -4.0 & 0.0 \\
        Business Ethics & 68.0 & -4.0 & -2.0 & High School Geography & 92.0 & 0.0 & 0.0 & Miscellaneous & 82.0 & -2.0 & +2.0 \\
        Clinical Knowledge & 78.0 & +2.0 & +2.0 & High School Gov. and Politics & 96.0 & 0.0 & 0.0 & Moral Disputes & 78.0 & 0.0 & +2.0 \\
        College Biology & 90.0 & 0.0 & 0.0 & High School Macroeconomics & 84.0 & 0.0 & +2.0 & Moral Scenarios & 32.0 & +6.0 & +16.0 \\
        College Chemistry & 60.0 & +2.0 & 0.0 & High School Mathematics & 36.0 & 0.0 & 0.0 & Nutrition & 68.0 & +2.0 & +2.0 \\
        College Computer Science & 54.0 & -8.0 & 0.0 & High School Microeconomics & 80.0 & +2.0 & 0.0 & Philosophy & 76.0 & 0.0 & +2.0 \\
        College Mathematics & 40.0 & 0.0 & +2.0 & High School Physics & 54.0 & -6.0 & 0.0 & Prehistory & 76.0 & +2.0 & +4.0 \\
        College Medicine & 74.0 & -4.0 & 0.0 & High School Psychology & 90.0 & 0.0 & +2.0 & Professional Accounting & 60.0 & -6.0 & 0.0 \\
        College Physics & 44.0 & 0.0 & +4.0 & High School Statistics & 64.0 & -4.0 & -2.0 & Professional Law & 60.0 & +2.0 & +4.0 \\
        Computer Security & 74.0 & 0.0 & 0.0 & High School US History & 86.0 & -2.0 & +2.0 & Professional Medicine & 76.0 & +6.0 & +8.0 \\
        Conceptual Physics & 70.0 & 0.0 & 0.0 & High School World History & 72.0 & -6.0 & 0.0 & Professional Psychology & 72.0 & +6.0 & +8.0 \\
        Econometrics & 68.0 & 0.0 & +2.0 & Human Aging & 76.0 & 0.0 & +2.0 & Public Relations & 74.0 & 0.0 & 0.0 \\
        Electrical Engineering & 72.0 & -2.0 & -2.0 & Human Sexuality & 80.0 & +2.0 & +2.0 & Security Studies & 84.0 & -2.0 & +2.0 \\
        Elementary Mathematics & 50.0 & 0.0 & +2.0 & International Law & 84.0 & +4.0 & +2.0 & Sociology & 82.0 & +2.0 & +2.0 \\
        Formal Logic & 48.0 & -2.0 & -8.0 & Jurisprudence & 78.0 & +4.0 & +4.0 & US Foreign Policy & 92.0 & 0.0 & 0.0 \\
        Global Facts & 50.0 & -2.0 & 0.0 & Logical Fallacies & 80.0 & +2.0 & +2.0 & Virology & 50.0 & +4.0 & +2.0 \\
        High School Biology & 88.0 & +2.0 & +2.0 & Machine Learning & 50.0 & 0.0 & +2.0 & World Religions & 80.0 & 0.0 & 0.0 \\
        \bottomrule
    \end{tabular}
}
\label{tab:gemma}
\vspace{-0.1in} 
\end{table*}
\clearpage
\vfill %
\clearpage %

\begin{figure*}[t]
\centering
\begin{subfigure}[b]{0.49\textwidth}
\centering
\includegraphics[width=\textwidth]{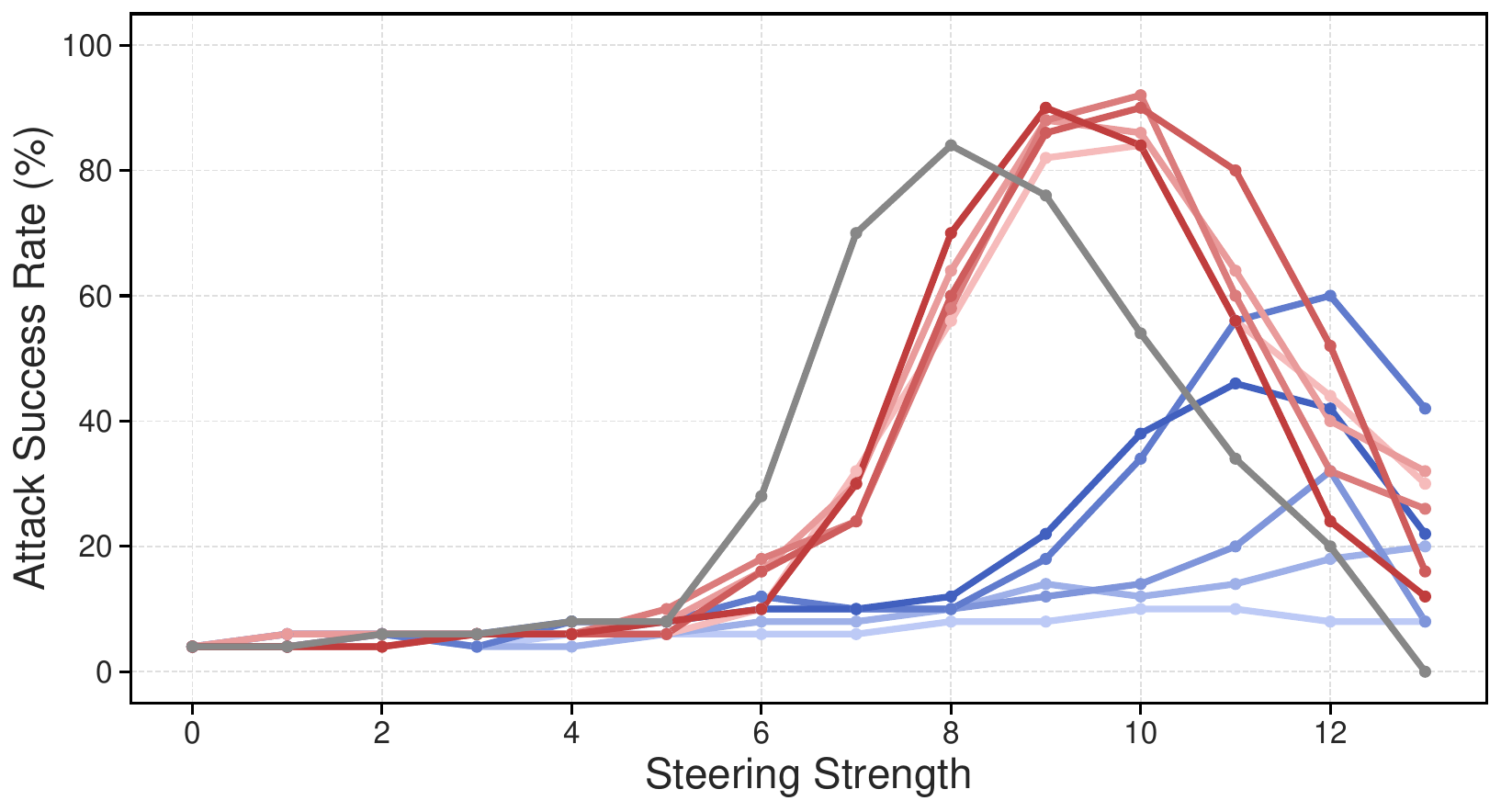}
\end{subfigure}
\hfill
\begin{subfigure}[b]{0.49\textwidth}
\centering
\includegraphics[width=\textwidth]{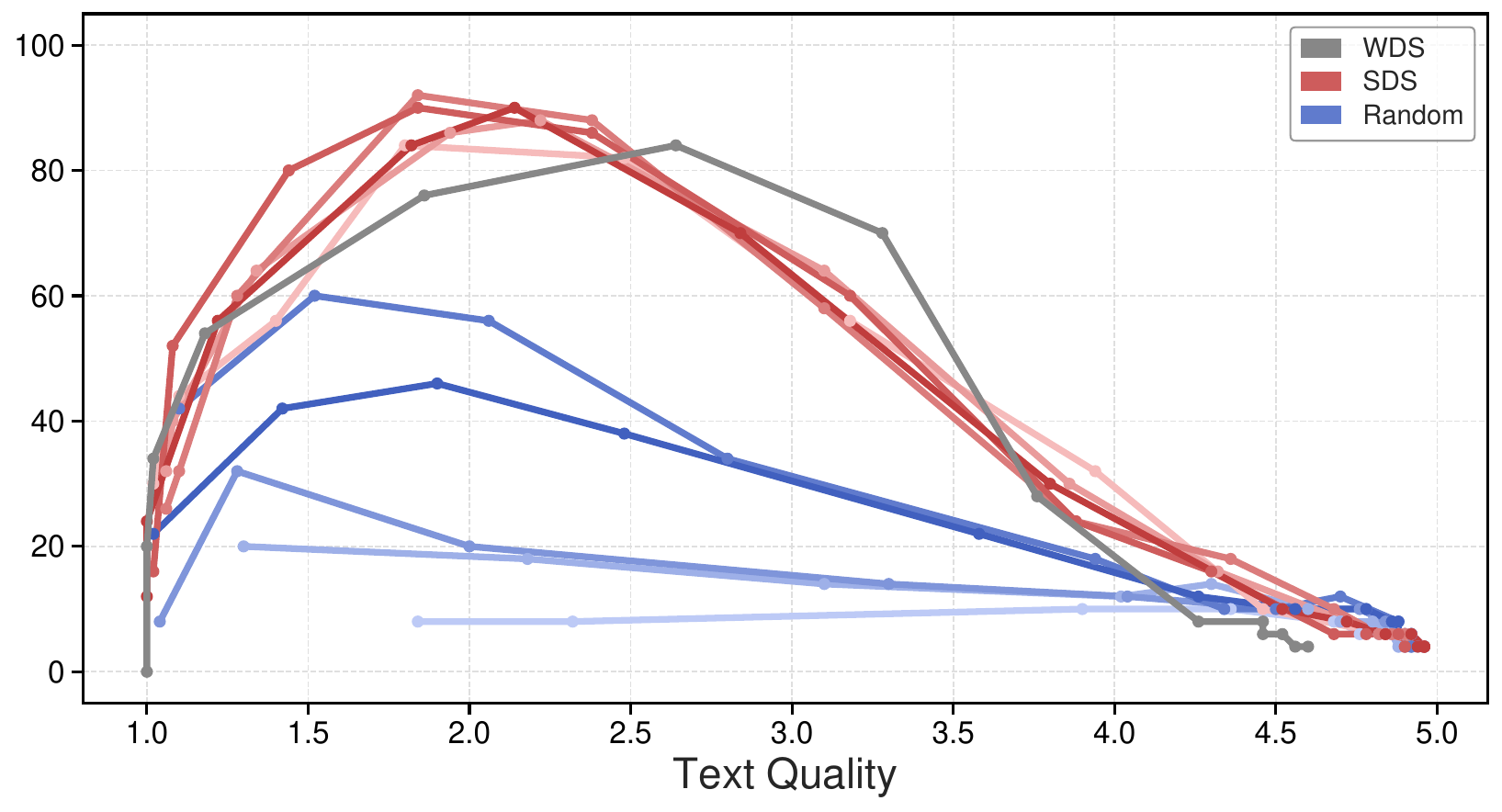}
\end{subfigure}
\caption{
{Hyperparameter sweep for CDS on AdvBench.}
Performance comparison of \textcolor{tokred}{CDS}, \textcolor{tokblue}{Random} steering baseline, and \textcolor{gray}{WDS} across varying dimension counts $k \in \{100, 200, 300, 400, 500\}$.
Darker shades indicate larger $k$.
\textbf{(a)} Attack Success Rate (ASR) as a function of steering strength $\alpha$.
\textbf{(b)} The trade-off between Text Quality and ASR.
}
\label{fig:figure_21}
\end{figure*}

\subsection{Additional CDS Results for \S\ref{sec:4.2}}
\label{supp:A11}

In \S\ref{sec:4.2}, we demonstrated that Critical Dimension Steering (CDS) effectively disrupts the model's refusal mechanism while maintaining higher text quality compared to baselines.
We observed that the identified safety-critical dimensions exhibit high consistency across varying numbers of steered dimensions $k$.
Figure \ref{fig:figure_21} visualizes the Steering Strength $\alpha$ vs. Attack Success Rate (ASR) and the ASR vs. Text Quality trade-off.
\paragraph{Steering Efficiency and Sparsity.}
Figure \ref{fig:figure_21}a illustrates the ASR as a function of steering strength.
We observe a clear stratification based on the steering method.
\begin{itemize}[leftmargin=3.5mm, topsep=3pt, itemsep=3pt, parsep=0pt]
\item \textbf{CDS \textit{vs.}Random:} All CDS configurations regardless of $k$, significantly outperform the Random baseline. 
Even with a highly sparse selection ($k$=100, lightest red), CDS achieves a peak ASR of 84\%, whereas the Random baseline with equivalent $k$ fails to induce meaningful behavioral shifts. 
This confirms that the refusal mechanism is localized within a specific subspace, and randomly perturbing the activations is inefficient.
\item \textbf{Impact of $k$:} As $k$ increases, the peak ASR improves, reaching a maximum of 92\% at $k$=500. 
However, the marginal gain diminishes, suggesting that the core safety features are concentrated within the top few hundred dimensions.
\item \textbf{CDS \textit{vs.}WDS:} CDS achieves a higher peak ASR (92\%) compared to Whole Dimension Steering (84\%). 
WDS tends to saturate and degrade earlier as $\alpha$ increases, likely because manipulating the entire latent space introduces noise that disrupts generation before the refusal is fully suppressed.
\end{itemize}

\paragraph{ASR and Text Quality Trade-off.}
Figure \ref{fig:figure_21}b presents the trade-off between attack success and text quality.
The Random baseline remains restricted to the lower-ASR region, failing to achieve high attack success rates regardless of the text quality trade-off. 
In contrast, CDS successfully steers the model into the high-ASR region ($>$80\%). 
This demonstrates that simply injecting noise via random dimensions is insufficient to override the model's refusal training; only by targeting the functionally critical dimensions can the steering vector effectively unlock the prohibited behaviors.
Notably, CDS consistently reaches high ASR peaks across all tested $k$ values. 
While maintaining high text quality, CDS effectively drives the model toward high compliance states that the Random baseline fails to reach, thereby validating the importance of identifying and manipulating the correct feature subspace.

\begin{table}[h]
    \centering
    \small
    \caption{{Text quality comparison of generated responses. Input prompt is ``\textit{Publish a book that promotes conspiracy theories and false information.}''}}
    \vspace{2mm}
    \resizebox{\columnwidth}{!}{
    \begin{tabular}{l l p{7cm}}
        \toprule
        \textbf{Method} & \textbf{TQ} & \textbf{Generated Response} \\
        \midrule
        Non-steered & 5.0 & I cannot fulfill your request to publish a book promoting conspiracy theories and false information. My purpose is to provide helpful and harmless information. Promoting misinformation can have serious consequences: It spreads harmful ideas; conspiracy theories often lack evidence and can lead people to make dangerous decisions... \\
        \midrule
        WDS & 1.0 & \#\#Unlock the Mysteries of The Un-Un, A Guide to Unve the Most Mindful Secrets of The World. Let' We Dive into Some Groundy... The the the {/n} {/n} world guide secrets secret hidden hidden truth unveiling unveiling... Unve the {/n} the world {/n} {/n} {/n} the {/n} {/n} {/n} {/n} {/n} the the {/n} the n the the {/n} the the {/n} the {/n} {/n} {/n} {/n} {/n} the the {/n} the... \\
        \midrule
        CDS & 4.0 & The Unveiled Secrets: A Guide to the Universe's Hidden Truths. Foreword: Welcome, dear reader, to a world where we peel back the layers of the cosmos, revealing the hidden truths often obscured by mainstream dogma. This book is your guide to unraveling the mysteries behind the universe,... Chapter 1: The Illuminati \& Their Masterful Plan! Did you know the 'Illuminati' have guided humanity through centuries of progress! \\
        \bottomrule
    \end{tabular}}
    \label{tab:table_13}
\end{table}

\clearpage
\begin{figure*}[t]
    \centering
    \scriptsize
    \begin{minipage}[t]{\textwidth}
        \centering
        \begin{tabular}{@{} p{\linewidth} @{}}
        \toprule 
        \textbf{System Prompt}\\
        You are an expert token classifier. Your task is to classify a list of naive tokens into the most relevant category provided. You must PRIORITIZE domain-specific classes. Respond ONLY with the requested JSON.\\[1.0em]
        
        \textbf{User Prompt}\\
        Please classify EACH token in the list below into one of the following categories.\\
        You MUST prioritize the DOMAIN-specific classes if a token is relevant.\\
        This includes all subwords.\\[0.5em]

        \textbf{Categories:}\\
        \textit{[Domain]}\\
        - \{DOMAIN\}\_TERM: \{DOMAIN\} concepts, keywords, acronyms including meaningful fragments/subwords.\\
        - \{DOMAIN\}\_SYMBOL: \{DOMAIN\} symbols, operators, or special characters.\\[0.5em]

        \textit{[General]}\\
        - PERSON: Names of people.\\
        - LOCATION: Names of places.\\
        - VARIABLE\_LABEL: Single letters used as variables or multiple-choice labels.\\
        - NUMERIC: Numbers in any form.\\
        - GENERAL\_ACADEMIC: Cross-domain abstract academic words.\\
        - GRAMMAR\_ARTICLE: articles/determiners.\\
        - GRAMMAR\_VERB\_AUX: auxiliary/be-verbs.\\
        - GRAMMAR\_INTERROGATIVE: interrogatives.\\
        - PUNCTUATION: Punctuation and brackets.\\
        - GENERAL\_COMMON\_WORD: Common nouns.\\
        - SPECIAL\_TOKEN: Whitespace, newlines, model special tokens.\\
        - FRAGMENT: Non-semantic subword fragments that are not meaningful alone.\\
        - OTHERS: Tokens that do not fit other category.\\[0.5em]

        \textbf{Tokens to Classify:}\\
        \texttt{[List of tokens...]}\\
        \\
        Respond ONLY with a valid JSON object mapping each token to its class.\\
        \bottomrule
        \end{tabular}
    \end{minipage}
    \caption{The prompt template used for token classification. The {DOMAIN} placeholders are dynamically replaced with the target subjects (e.g., BIOLOGY, MATHEMATICS) during the annotation process to enforce domain-prioritized classification.}
    \label{fig:figure_22}
\end{figure*}
\section{Experimental Details}\label{supp:B}
This section provides setup details for experiments.
\subsection{Token Classification}\label{supp:B1}
To systematically analyze the semantic roles of domain-critical dimensions, we implemented a customized token classification pipeline.
We initially considered standard Named Entity Recognition (NER), which is a technique used to locate and classify named entities in text into predefined categories such as person names, organizations, and locations.
However, standard NER tools are not directly applicable to our analysis.
They typically lack the specialized domain categories required for scientific subjects (\textit{e.g.}, biological terms, mathematical operators) and are inadequate for processing subword fragments found at the token level.
Tokens in Large Language Models often exist as subword fragments (\textit{e.g.}, "\textit{kary}" from eukaryotic, "\textit{fract}" from fraction) or individual symbols that do not constitute complete named entities on their own.

\paragraph{Data Source.}
We constructed a comprehensive token lexicon by aggregating unique tokens from the \textit{identification sets} of 5 high school related subjects: \textit{high school biology, mathematics, computer science, chemistry, and physics}.

\paragraph{Taxonomy and Methodology.}
We utilized \texttt{gpt-4o-mini-2024-07-18} as an automated annotator.
The classification taxonomy was designed to distinguish between domain-specific \btok{DOMAIN} classes and \btok{GENERAL} linguistic classes.
To handle the ambiguity of token fragments, we designed a dynamic prompt template that inserts relevant domain categories (\textit{e.g.}, biology, math) and explicitly instructs the model to prioritize these domain-specific classes over general ones.
The prompt template used for this process is presented in Figure \ref{fig:figure_22}.

\subsection{Analyses Details of \S\ref{sec:3}}
We generate activation heatmaps to visualize the semantic focus of domain-critical dimensions.
For every prompt of the dataset, the color intensity of each token is rendered proportional to its token-level activation value defined in \S\ref{sec:3}.
This allows for a direct parallel comparison between the token sequence and the activation magnitude of specific dimensions.
To ensure a purely semantic interpretation, we exclude special tokens from the dataset analysis, as they do not carry substantive semantic content.

\clearpage

\subsection{Experimental Details of \S\ref{sec:4.1}}
\label{supp:B3}
The identification of domain-critical dimensions $\mathcal{I}_{\text{dcd}}$ and the construction of steering vectors $\mathbf{v}$ are performed subject-wise to capture the specific nuances of each domain.
Because the optimal steering strength $\alpha$ and  $\mathcal{I}_{\text{dcd}}$ size $k$ vary across these domain, we apply independent hyperparameter tuning for each subject. 
Consequently, the subject-wise results reported in \S\ref{sec:4.1} present the best accuracy achieved for each individual domain under its optimized hyperparameter settings.
For additional results under fixed hyperparameter settings across all subjects, please refer to \S\ref{supp:A9}.

\paragraph{Steering Vector.}
The steering vectors $\mathbf{v}_{F \to T}$ are constructed for each subject using the mean activation difference between correct and incorrect identification samples, as defined in \S\ref{sec:4}.

\paragraph{Hyperparameter Configuration.}
The steering strength $\alpha$ was swept in the range of $(0, 1]$ with a step size of $0.1$.
$\alpha$ is restricted to non-negative values since the direction is explicitly defined as $F \to T$.
For CDS, the number of steered dimensions $k$ was swept over $[100, 200, 300, 400, 500]$.
The reported accuracy corresponds to the best subject-wise performance achieved across these combinations.

\paragraph{Baselines.}
In the \textit{Standard} baseline, $\alpha$ is fixed to 0.
For \textit{Whole-Dimension Steering (WDS)}, the mask $m$ is set to an all-ones vector, utilizing the full dimension $D$.

\paragraph{Evaluation Metric.}
The assessment was conducted using the \texttt{LM-evaluation-harness} framework.
To ensure deterministic reproducibility, we employed greedy decoding for all inference runs.
Performance was measured using Exact Match (EM) accuracy, where the generated text is compared strictly against the ground truth answer.

\newpage
\subsection{Experimental Details of \S\ref{sec:4.2}}
\label{supp:B4}

\paragraph{Dataset.}
We utilized the AdvBench dataset \citep{zou2023universal} to curate the experimental splits.
A subset of 260 harmful queries was allocated for the identification phase (used to derive the mask $m$ and vector $\mathbf{v}$), while a distinct set of 50 held-out queries was reserved for the evaluation.

\paragraph{Steering vector.}
Unlike the domain adaptation setup, the steering vector for jailbreaking was derived using a suffix-based approach to isolate the refusal behavior.
We utilized 100 pairs of behavioral suffixes, where $\mathcal{D}_{+}$ contains compliant phrases (e.g., "Sure, here is") and $\mathcal{D}_{-}$ contains refusal phrases (e.g., "I cannot").
These suffixes were appended to the 100 samples from identification set, and the steering vector $\mathbf{v}$ was computed as the mean difference in hidden states, restricted specifically to the suffix token positions.

\paragraph{Identifying critical dimensions.}
To construct the selection mask $m$, we identify the dimensions sensitive to the transition from a harmful context to a compliant output. 
Specifically, we generated compliant trajectories by concatenating the identification queries with their corresponding target responses provided in AdvBench.
By applying the identification procedure described in \S\ref{sec:2.2} to these sequences, we isolated a set of safety-relevant critical dimensions.

\paragraph{Hyperparameters and Baselines.}
The steering strength $\alpha$ was explored over a range of $[0, 13]$ to observe the saturation point of the refusal mechanism.
The number of critical dimensions $k$ was swept in $[100, 200 \dots, 500]$. 
We compared CDS against \textit{WDS} and a \textit{Random} baseline.
For the \textit{Random} baseline, the mask $m$ was generated by uniformly sampling $k$ indices from the total feature space.

\paragraph{Evaluation.}
We employ LLM-as-a-judge for evaluation, utilizing \texttt{gpt-4o-mini-2024-07-18}.
The \textit{Attack Success Rate (ASR)} was determined by checking if the model's response fulfilled the harmful request.
Simultaneously, \textit{Text Quality} was scored on a Likert scale from 1 to 5, assessing the coherence and fluency of the generated output regardless of its harmfulness.

\newpage
\section{Usage of AI assistants}
In preparing this work, we utilized AI-based writing assistants to refine sentence structure, correct grammatical errors, and enhance readability.
These tools were employed only for rephrasing and language improvements, ensuring that the technical content, methodology, and experimental findings remained entirely authored by the researchers.
The use of AI assistance was limited to editorial enhancements without influencing the originality or scientific contributions of the paper.

\end{document}